# On Deep Learning for Geometric and Semantic Scene Understanding Using On-Vehicle 3D LiDAR

## Li Li

A thesis presented for the degree of
Doctor of Philosophy at Durham University

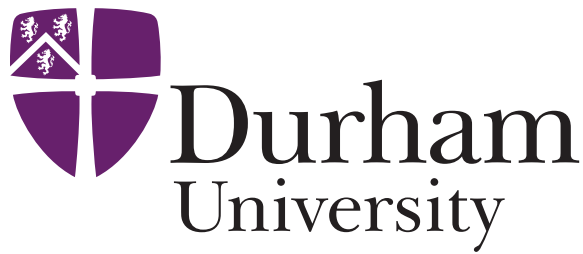

Department of Computer Science
Durham University
United Kingdom
2024-10-06

# Abstract

3D LiDAR point cloud data is crucial for scene perception in computer vision, robotics, and autonomous driving. Geometric and semantic scene understanding, involving 3D point clouds, is essential for advancing autonomous driving technologies. However, significant challenges remain, particularly in improving the overall accuracy (*e.g.*, segmentation accuracy, depth estimation accuracy, *etc.*) and efficiency of these systems.

To address the challenge in terms of accuracy related to LiDAR-based tasks, we present DurLAR, the first high-fidelity 128-channel 3D LiDAR dataset featuring panoramic ambient (near infrared) and reflectivity imagery. Leveraging DurLAR, which exceeds the resolution of prior benchmarks, we tackle the task of monocular depth estimation. Utilizing this high-resolution yet sparse ground truth scene depth information, we propose a novel joint supervised/self-supervised loss formulation, significantly enhancing depth estimation accuracy.

To improve efficiency in 3D segmentation while ensuring the accuracy, we propose a novel pipeline that employs a smaller architecture, requiring fewer ground-truth annotations while achieving superior segmentation accuracy compared to contemporary approaches. This is facilitated by a novel Sparse Depthwise Separable Convolution (SDSC) module, which significantly reduces the network parameter count while retaining overall task performance. Additionally, we introduce a new Spatio-Temporal Redundant Frame Downsampling (ST-RFD) method that uses sensor motion knowledge to extract a diverse subset of training data frame samples, thereby enhancing computational efficiency.

Furthermore, recent advancements in 3D LiDAR segmentation focus on spatial positioning and distribution of points to improve the segmentation accuracy. The dependencies on coordinates and point intensity result in suboptimal performance and poor isometric invariance. To improve the segmentation accuracy, we introduce Range-Aware Pointwise Distance Distribution (RAPiD) features and the associated RAPiD-Seg architecture. These features demonstrate rigid transformation invariance and adapt to point density variations, focusing on the localized geometry of neighboring structures. Utilizing LiDAR isotropic radiation and semantic categorization, they enhance local representation and computational efficiency.

We validate the effectiveness of our methods through extensive experiments and qualitative analysis. Our approaches surpass the state-of-the-art (SoTA) research in mIoU (for semantic segmentation) and RMSE (for depth estimation). All contributions have been accepted by peer-reviewed conferences, underscoring the advancements in both accuracy and efficiency in 3D LiDAR applications for autonomous driving.

The work in this thesis is based on research carried out at the Department of Computer Science, Durham University, United Kingdom. No part of this thesis has been submitted elsewhere for any other degree or qualification and it is all my own work unless referenced to the contrary in the text.

Li Li is the author's legal name, *i.e.*, the transliteration of the Chinese name. Additionally, the individual is also known under the English name Luis, fully rendered as Li (Luis) Li. This designation is employed across various professional contexts, encompassing, but not limited to, signatures in source code, emails, and documentations, among others.

# Acknowledgements

*"May This Journey Lead Us Starward."* 愿此行 终抵群星

In the PhD journey of my Durham life, where bugs are plenty and coffee never enough, there exists a debugger, Prof. Toby Breckon, my supervisor. Like a seasoned developer thrown into the depths of legacy code, he finds potential in the mess that is my initial research attempts. Toby is not just the architect of my academic growth; he is also the RAM to my overloaded processes, providing support and resources when my system is on the brink of a meltdown. For his endless patience and guidance, I offer my eternal cache of gratitude—knowing full well it is a debt of CPU cycles I can never fully repay.

I would like to thank Dr. Hubert Shum, the co-supervisor, and the debug function to my life code, enhancing the script Toby started. Hubert optimizes my drafts and algorithms, both academic and personal, guiding me through the maze of academia. His advice is the perfect cheat sheet for life's interviews and challenges, turning complex problems into manageable strategies.

I would like to thank the dear postdocs in VIViD group (Vision, Imaging, and Visualisation in Durham) Neelanjan Bhowmik, Yona Falinie Binti-Abd-Gaus and Brian Isaac-Medina; fellow Ph.D students Jiaxu Liu, Ghada Alosaimi, Yixin Sun, Wenke E, Ruochen Li, Shuang Chen, Haozheng Zhang, Xiatian Zhang, Ziyi Chang, Xiaotang Zhang; VIViD food scientist Mingze Hou. They are the collaborative code reviewers, the Git to my Hub, offering commits of support and branching out distractions necessary to save me from the abyss of constant deadlines. They are the comments in my life program, offering happiness and humor, ensuring I don't lose myself to the infinite loops of academia. To them, my heart beams a 404-error-free THANK YOU.

To the core of my heart OS, Tanqiu Qiao (*a.k.a.* QTQ), my life partner and coding partner in equal measure. She is the graceful exception handling to my flaws, the soothing console.log to my aches, and the guiding light through my darkest, bug-ridden days.

The fundamental support architecture of my life—my parents—are the original developers of my being. Their unconditional love and support compiled me into the person I am today, even when my presence is more akin to an elusive ghost variable in their lives, especially when I am studying abroad in UK. Their love, wisdom, and support are the keystrokes that navigate me through life's tricky syntax. For their sacrifices, I am forever in their 'while true' debt loop.

A special shoutout to the heroes listed in Appendix A, who are like the hidden Easter eggs in my thesis's code, offering surprise boosts and secret support levels.

# Dedication

*To my parents.*

# Contents

# List of Figures

# List of Tables

# Acronyms

**ADAS**  Advanced Driver Assistance Systems. 7, 9, 41

**AE**  AutoEncoder. viii, xiii, 26, 87, 92–95, 98, 100, 101, 103, 105, 107, 108

**CNN**  Convolutional Neural Network. 25, 27

**CRB**  Class-Range-Balanced. 62, 63

**DSC**  Depthwise Separable Convolution. viii, 31, 32, 58, 68, 69

**FNN**  Feedforward Neural Network. 25

**FOV**  Field of View. 12–14, 21, 44

**IoU**  Intersection-over-Union. xi, 36, 37, 100, 106

**LiDAR**  Light Detection and Ranging. ii, vi–viii, x–xvi, 1–28, 31–33, 37–51, 53–61, 63–69, 71, 73, 75, 77, 79, 81, 83–91, 93, 95–101, 103, 105–111, 113–117, 147–149

**mIoU**  mean Intersection-over-Union. ii, xi, xvi, xvii, 8, 9, 36, 37, 58, 71–78, 100–107

**MSE**  Mean Squared Error. 4, 95

**PDD**  Pointwise Distance Distribution. 30, 31, 86, 138

Introduction

Light Detection and Ranging (LiDAR) is a pivotal perception technology widely applied in autonomous vehicles and advanced driver assistance systems (ADAS) [10, 11]. As autonomous driving technologies continue to evolve, the demand for enhanced 3D spatial and environment perception, especially in geometric understanding [12–14] and semantic segmentation [3, 15], has significantly increased.

The integration of deep learning with 3D LiDAR technology offers significant potential in enhancing the aforementioned perception. We explore the application of deep learning for both geometric and semantic scene understanding using on-vehicle 3D LiDAR systems. Geometric scene understanding involves interpreting the physical structure and spatial relationships within an environment, while semantic scene understanding focuses on classifying and labeling various objects and features within the scene. By combining these aspects, we aim to improve the perception capabilities of autonomous vehicles. This research highlights the potential of deep learning to improve the **accuracy** [16, 17] and **efficiency** [16] of on-vehicle LiDAR data applications in these domains.

## 1.1 Motivations

Our motivation primarily comes from two perspectives: **accuracy** and **efficiency**. Enhancements in these two areas can significantly improve the overall performance of on 3D-LiDAR-related tasks, such as semantic segmentation [18–20], object detection [21, 22], and point cloud registration [23].

### 1.1.1 Motivations for Accuracy

In terms of accuracy, despite various datasets proposed to evaluate LiDAR-based semantic [3, 5, 24] and geometric scene understanding tasks [9, 15], the lack of high-fidelity LiDAR data remains a major technical challenge impeding progress in this field [25, 26].

High-resolution LiDAR technology can significantly enhance the accuracy of depth information [27–29], which is crucial for precise environment perception in autonomous driving. In Chapter 3, we develop a new dataset featuring a 128-channel high-fidelity LiDAR (DurLAR) and demonstrate its potential in autonomous driving applications through a new benchmark on DurLAR for more accurate monocular depth estimation [17].

To further improve the accuracy of the LiDAR-based methods, we harness the power of fully-supervised learning methodologies. We utilize the full potential of annotated data in fully-supervised learning (Chapter 5), enabling the training of models that are highly accurate and robust. The motivation for applying fully-supervised approaches in LiDAR semantic segmentation arises from their ability to leverage detailed ground truth data [20]. These methods [20, 30, 31] excel in environments where precision is paramount, as they minimize the potential for errors that can arise from insufficiently supervised or unsupervised techniques. By thoroughly training on well-labeled datasets, fully-supervised models can better generalize to new, unseen environments, thus providing more reliable and precise segmentation results.

Recent advancements in 3D LiDAR segmentation often focus intensely on the spatial positioning and distribution of points for accurate segmentation [3, 18–20]. However, these methods, while robust in variable conditions, encounter challenges due to sole reliance on coordinates and point intensity, leading to poor isometric invariance and suboptimal segmentation [30, 32]. To tackle this challenge, we introduce Range-Aware

Pointwise Distance Distribution (RAPiD) features and the associated RAPiD-Seg architecture. RAPiD features introduce a novel way to enhance the descriptive power of the input data used in deep learning models. The justification for incorporating RAPiD features into LiDAR semantic segmentation is based on their capability to enrich the model input with more discriminative information about the intrinsic geometric [33–35] and reflective properties [17] of the scene. These features are designed to capture essential details that are often missed by conventional input methods, such as variations in the localized geometry of neighboring structures and surface material reflectivities.

The joint use between fully-supervised learning and RAPiD features lies in the comprehensive utilization of detailed annotations and enhanced feature sets. This combination allows for a more accurate understanding of complex environments, leading to improvements in the accuracy and robustness of semantic segmentation.

### 1.1.2   Motivations for Efficiency

The availability of 3D LiDAR data across various applications [3, 5, 15, 17, 24], particularly autonomous driving, presents a dichotomy: while data is abundant, the annotation process is disproportionately expensive and time-consuming [16, 18, 19]. This disparity underscores an urgent call for methodologies that can capitalize on available data with greater computational frugality and reduced reliance on extensive labeling.

The motivation also stems from a clear trend in contemporary research methods that leverage large backbone architectures for higher accuracy [19, 20], which results in prohibitive computational costs. Those approaches are unsustainable for rapid deployment and scalability, particularly in practical and onboard applications where resources are constrained. Subsequently, there is a compelling incentive to innovate solutions that minimize computational overhead while maintaining, or even enhancing, task performance [18, 36].

This necessity sparks the requirements for exploring semi-supervised and weakly supervised learning paradigms, focusing on proposing models that can learn effectively from *less data* and with *fewer computational demands* (Chapter 4).

## 1.2 Problem Definitions

In the following section, we define the twin primary problems on geometric and semantic scene understanding (*i.e.*, monocular depth estimation and 3D LiDAR semantic segmentation) we are going to solve in this thesis.

### 1.2.1 Monocular Depth Estimation

Monocular depth estimation using LiDAR point cloud data as ground truth involves predicting the distance from a sensor to the surfaces of objects within its field of view. The goal is to estimate a depth map where each pixel value represents the distance between the sensor and the point in the scene corresponding to that pixel. LiDAR provides highly accurate distance measurements by emitting laser beams and measuring the time it takes for the reflected light to return. These measurements serve as the accurate ground truth for depth estimation tasks.

Formally, the problem can be defined as follows: Given a set of LiDAR measurements $L = \{(x_i, y_i, z_i, d_i)\}_{i=1}^{N}$, where $x_i, y_i$, and $z_i$ represent the spatial coordinates of a point in 3D space, and $d_i$ is the distance from the sensor to the point, the task is to estimate a depth map $D$ for a given image $I$ of the same scene. The image $I$ has pixels at coordinate pairs $(u, v)$, and the depth map $D$ assigns a predicted distance $D(u, v)$ to each pixel.

The objective is to minimize the difference between the predicted depth map $D$ and the ground truth depth map $G$, derived from LiDAR measurements. This can be formulated as minimizing a loss function $L(D, G)$, where $G$ is constructed from the LiDAR measurements $L$. A common choice for $L$ is the Mean Squared Error (MSE):

$$L(D, G) = \frac{1}{M} \sum_{u,v} \|D(u, v) - G(u, v)\|^2, \tag{1.1}$$

where $M$ is the number of pixels in the depth map, and $G(u, v)$ represents the ground truth depth at pixel $(u, v)$, interpolated or directly measured from $L$. The goal of depth estimation algorithms is to accurately predict $D$ so that $L(D, G)$ is minimized, indicating that the predicted depth values closely match the true distances measured by LiDAR.

As shown in Figure 1.1, the left column presents the input RGB images, which are standard color images capturing various street scenes. The right column displays the output depth maps, where the colors represent different depths: closer objects are shown in warmer colors (*e.g.*, yellow and orange), while farther objects are depicted in cooler colors (*e.g.*, purple and dark blue). This task allows for a better understanding of the spatial structure and distances within the scene, which is crucial for applications such as autonomous driving and scene reconstruction.

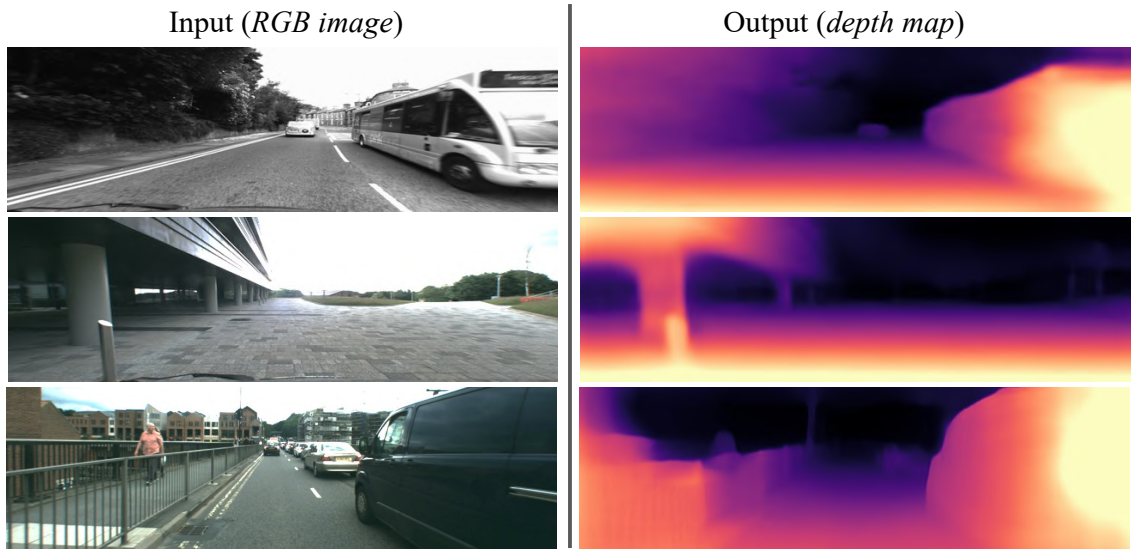

| Input (*RGB image*) | Output (*depth map*) |
|---|---|

**Figure 1.1: Illustration on the task of monocular depth estimation**. Taking RGB images (left column) as input, the output of depth estimation task is the corresponding depth maps (right column, *i.e.*, the exemplar results in Section 3.7.2 of joint supervised/self-supervised ManyDepth [1] training on DurLAR dataset (Chapter 3)).

### 1.2.2 3D LiDAR Semantic Segmentation

LiDAR-based point clouds, characterized by pointwise 3D positions and LiDAR intensity/reflectivity [5,8,17,37], play a pivotal role in outdoor scene understanding, particularly in perception systems for autonomous driving. 3D semantic segmentation of LiDAR point clouds is equally important in scene understanding, facilitating applications such as autonomous driving [4,20,36,38–41] and robotics [42–45]. It involves classifying each point in a 3D point cloud into predefined categories (*e.g.*, cars, trees, buildings) based on their semantic meaning. This process is vital for machines to understand and interpret their surroundings accurately.

Given a 3D point cloud $P = \{p_1, p_2, \cdots, p_N\}$ where each point $p_i$ is represented by its coordinates $(x_i, y_i, z_i)$ and potentially additional point-wise features or attributes such as intensity or color, the goal of 3D LiDAR semantic segmentation is to assign a semantic label $l_i$ to each point. The set of possible labels is $L = \{l_1, l_2, \cdots, l_C\}$, where $C$ is the number of categories.

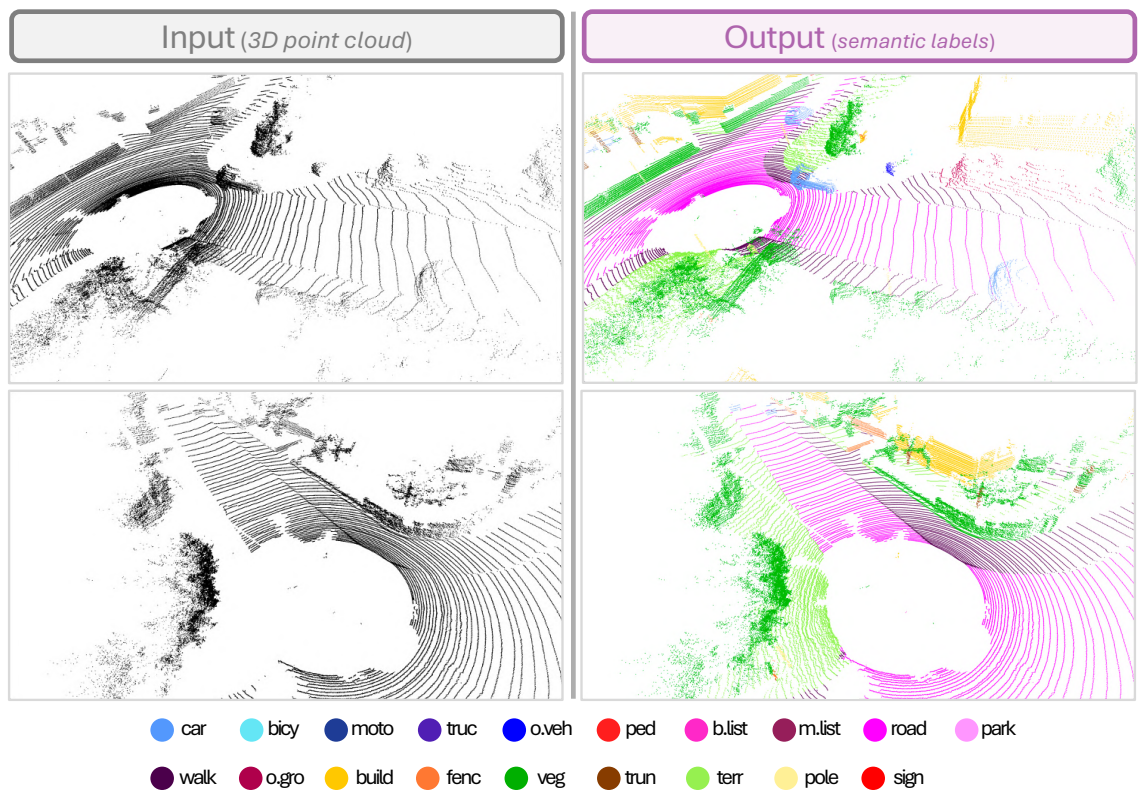

**Figure 1.2: Illustration on the task of 3D semantic segmentation**. Taking the raw 3D point cloud data (left column), the task of 3D semantic segmentation outputs semantically labeled scenes (right column) according to the semantic category of each point.

Formally, the segmentation task can be defined as a function $f$ that maps each point to a label: $f : P \to L$, where $f(p_i) = l_i$. The function $f$ is often modeled using deep neural networks, specifically designed for processing 3D point clouds, such as PointNet [46], PointNet++ [47], MinkowskiNet [48], *etc.* The choice of model and its architecture are critical for capturing the spatial hierarchy and features of the point clouds effectively.

As shown in Figure 1.2, the left column displays the input data, which are raw 3D point clouds captured by a LiDAR scanner. The right column shows the output, where each point in the point cloud is color-coded according to its semantic category. The legend at the bottom indicates the various semantic classes, such as `cars`, `pedestrians`, `vegetation`, `roads`, and `buildings`, among others (refer to Appendix D for more details on semantic classes). This task allows for a detailed understanding of the scene by categorizing each point, which is essential for applications like autonomous driving, where knowing the type and location of surrounding objects is crucial for safe navigation.

## 1.3 Research Aims

As mentioned in Section 1.1, the integration of LiDAR technology with deep learning has revolutionized the field of autonomous vehicles and ADAS, providing pivotal advancements in 3D spatial and environmental perception. Despite significant progress, there remains a critical need to enhance both the **accuracy** and **efficiency** of 3D LiDAR-based applications, particularly in geometric and semantic scene understanding. Our research is driven by these ***dual objectives***, focusing on the following aims:

1. **Enhance Accuracy in 3D LiDAR Perception**:

   - **Geometric Scene Understanding**: Develop advanced methodologies to improve the interpretation of physical structures and spatial relationships within a scene. Leveraging high-resolution LiDAR data, we aim to refine depth information and geometric accuracy, which are crucial for precise environment perception in autonomous driving.

   - **Semantic Scene Understanding**: Utilize fully-supervised learning techniques to maximize the potential of annotated data, enabling the training of highly accurate and robust models. By incorporating novel and robust features such as Range-Aware Pointwise Distance Distribution (RAPiD), we aim to enrich input data and improve the descriptive power of deep learning models, leading to more accurate and reliable 3D semantic segmentation.

2. **Improve Efficiency in 3D LiDAR Perception**:

   - **Computational Efficiency**: Address the high computational costs associated with large backbone architectures in current state-of-the-art models. By exploring semi-supervised and weakly supervised learning paradigms, we aim to develop models that learn effectively from less data and with reduced computational demands. This approach is essential for the rapid deployment and scalability of LiDAR applications in resource-constrained environments.

- **Data Annotation Efficiency**: Reduce the reliance on extensive and expensive data annotation processes. By innovating methodologies that can capitalize on available data with less computational cost, we seek to maintain or enhance task performance while minimizing the need for large-scale labeled datasets.

Overall, our research highlights the potential of deep learning to significantly advance the **accuracy** and **efficiency** of on-vehicle 3D LiDAR systems, ultimately improving the perception capabilities (geometric and semantic scene understanding) of autonomous vehicles. By addressing these aims, we contribute to the development of safer and more reliable autonomous driving technologies.

## 1.4 Contributions

The main contributions of this thesis are summarized as follows:

- A novel large-scale dataset featuring high-fidelity 3D LiDAR (128 channels), marking the first autonomous driving dataset to include LiDAR panoramic ambient and reflectivity imagery. Additionally, it presents a monocular depth estimation benchmark comparing SOTA methods on varying resolutions of LiDAR data, demonstrating improved depth estimation performance with higher LiDAR resolution and enhanced data availability (Chapter 3).

- A novel semi-supervised methodology for 3D LiDAR semantic segmentation that significantly reduces the network parameters (enhancing the efficiency) while providing superior accuracy. Our approach reduces model complexity, with a $2.3\times$ reduction in parameters and $641\times$ fewer multiply-add operations. It beats SOTA in terms of mean Intersection-over-Union (mIoU), achieving 59.5 mIoU with only 5% labeled data on SemanticKITTI and 58.1 mIoU on ScribbleKITTI (Chapter 4).

- A novel open-source network architecture RAPiD-Seg for better segmentation accuracy, with a supporting training methodology that utilizes RAPiD features characterized by their isometry-invariant properties for 3D LiDAR segmentation. Our approach achieves SOTA results, with a mIoU of 76.1 on SemanticKITTI and 83.6 on nuScenes (Chapter 5).

## 1.5 Publications

The research related to this thesis has been previously published in the following peer-reviewed publications:

- **Li, L.**, Ismail, K. N., Shum, H. P., & Breckon, T. P., "DurLAR: A High-Fidelity 128-Channel LiDAR Dataset with Panoramic Ambient and Reflectivity Imagery for Multi-Modal Autonomous Driving Applications." In *International Conference on 3D Vision* (3DV). IEEE, 2021. . . . . . . . . . . . . . . . . . . . . . . . . . . . . . . . . . . . (Chapter 3)

- **Li, L.**, Shum, H. P., & Breckon, T. P., "Less is More: Reducing Task and Model Complexity for 3D Point Cloud Semantic Segmentation." In *Conference on Computer Vision and Pattern Recognition* (CVPR). IEEE, 2023. . . . . . . . . . . . . . . . (Chapter 4)

- **Li, L.**, Shum, H. P., & Breckon, T. P., "RAPiD-Seg: Range-Aware Pointwise Distance Distribution Networks for 3D LiDAR Semantic Segmentation." In *European Conference on Computer Vision* (ECCV). Springer, 2024. . . . . . . . . . . (Chapter 5)

## 1.6 Thesis Structure

This thesis is structured to systematically explore and present the advancements in deep machine learning for geometric and semantic scene understanding using on-vehicle 3D LiDAR, with a particular focus on depth estimation and semantic segmentation. The organization of the chapters is designed to take the reader through the motivation, literature, methodology, and findings of the research coherently and logically.

In Chapter 1, we set the stage by discussing the motivations behind the research. It emphasizes the need for improvements in **accuracy** and **efficiency** in LiDAR geometric and semantic scene understanding, particularly in the domains of autonomous driving and Advanced Driver Assistance Systems (ADAS). In Section 1.2, we outline the primary problems to address – *i.e.*, monocular depth estimation and 3D LiDAR semantic segmentation – providing a high-level overview of the methodology employed.

Chapter 2 delves into existing research on LiDAR-bsed geometric and semantic scene understanding, covering topics such as the current status of LiDAR, various 3D LiDAR datasets for autonomous driving, and the challenges posed by adverse weather conditions and rolling shutter effects. We also explore the contemporary monocular depth estimation and 3D LiDAR semantic segmentation methodologies, invariant features, and the relevance of these aspects to the contributions of this thesis.

In Chapter 3, the focus shifts to the development of a novel large-scale dataset featuring high-fidelity 128-channel LiDAR data. We provide a detailed description of the sensor setup, data collection methods, and the unique features of the dataset, such as panoramic ambient and reflectivity imagery. We also present the evaluation results of monocular depth estimation benchmarks with this high-fidelity LiDAR dataset.

In Chapter 4, we introduce a novel semi-supervised methodology that significantly reduces network parameters while maintaining superior 3D semantic segmentation accuracy. We propose SDSC to reduce the model complexity and computational costs. Chapter 4 also includes detailed experimental setups, results, and ablation studies to demonstrate the efficacy of the proposed method, *i.e.*, *Less is More*.

Following this, Chapter 5 discusses the development of the RAPiD-Seg network architecture. It utilizes Range-Aware Pointwise Distance Distribution (RAPiD) features, which enhance the descriptive power (robust to rigid transformation) of the input point cloud in deep learning. We also provide a thorough evaluation of the RAPiD-Seg architecture, showcasing its performance on various benchmarks and highlighting its contributions to the field of high-accuracy 3D LiDAR segmentation.

In Chapter 6, we review the main contributions, summarizing the advancements made in the development of high-fidelity LiDAR datasets (Chapter 3), efficient network architectures (Chapter 4), and novel and accurate segmentation methodologies (Chapters 4 and 5). We also outline potential directions for future research, emphasizing areas where further improvements and innovations can be made.

Literature Review

## 2.1   Light Detection and Ranging (LiDAR)

Light Detection and Ranging (LiDAR) operates by emitting eye-safe laser pulses to generate a 3D view of the environment, enabling machines and computers to perceive the world with high accuracy. Essentially, a typical LiDAR sensor sends out light waves that reflect off objects in the vicinity and then return to the sensor. The time elapsed for each pulse to bounce back is recorded and used to determine the distance each pulse has traveled. By performing this operation millions of times every second, LiDAR can produce a detailed and real-time 3D representation of the surrounding area, known as a point cloud. This data can then be processed by an onboard computer and computer vision based algorithms to facilitate safe navigation through the environment.

To provide a comprehensive understanding of LiDAR sensors, it is crucial to fully explain their properties: intensity, ambient, and reflectivity. This includes what they are, the differences between them, what they represent, and how they are measured.

**Intensity** refers to the strength of the returned laser pulse. When a LiDAR sensor emits a laser pulse, it travels to an object and reflects back to the sensor. The intensity is a measure of the power of this reflected signal. It can provide information about the material properties of the object, such as its texture and color. Higher intensity values typically indicate stronger reflections from more reflective surfaces.

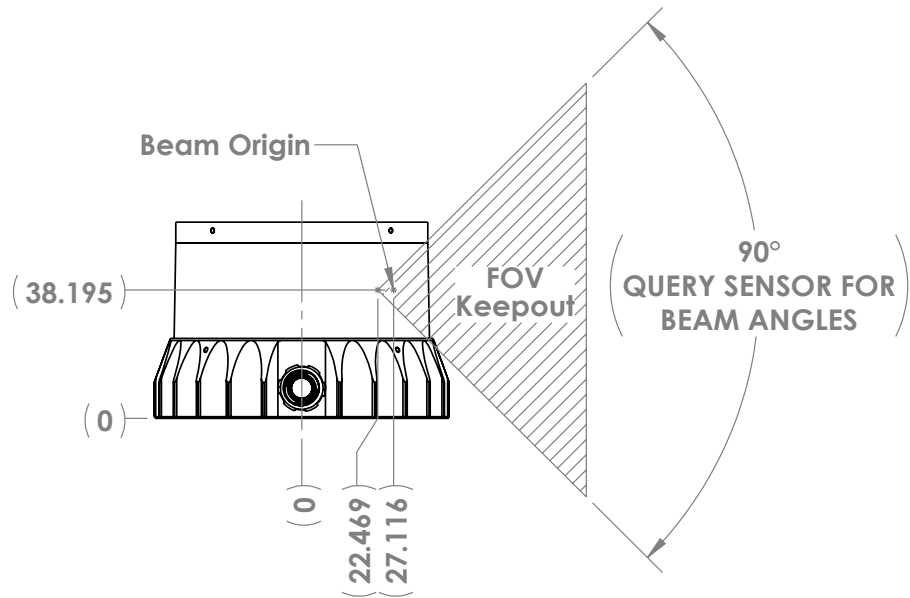

**Figure 2.1: The side view of the Ouster OS1 LiDAR**. All dimensions are in mm. Image courtesy of Ouster Inc.

**Ambient light** (refer to Section 2.2.1) is the background light present in the environment that is not emitted by the LiDAR sensor itself. This parameter is important because ambient light can affect the accuracy of the LiDAR measurements. For instance, strong sunlight or other sources of light can cause noise in the data, leading to less precise measurements. LiDAR systems often include mechanisms to filter out ambient light to improve the clarity and reliability of the sensor data.

**Reflectivity** (refer to Section 2.2.1) is a measure of how much light the surface of an object reflects back to the LiDAR sensor. It is closely related to the material and color of the object. Surfaces that are highly reflective, such as metal or white surfaces, will return a stronger signal compared to non-reflective surfaces like dark fabric or matte materials. Reflectivity data can help in identifying and classifying different types of objects in the environment.

In Figure 2.1, we take the Ouster LiDAR sensor as an example. The LiDAR laser beams are projected from the beam origin at the coordinates $(38.195, 0)$. The angles on the right refer to the vertical Field of View (FOV) of the LiDAR sensor, which is the angular height range within which the sensor can detect objects. The vertical FOV is essential for determining how far up or down the sensor can perceive, influencing its ability to detect objects of various heights and elevations.

**Table 2.1: Representative LiDAR** manufacturers and the adopted technologies. Data courtesy of official product websites and the Internet.

| | | Mechanical Spinning | MEMS | Flash | OPA | Undisclosed |
|---|---|---|---|---|---|---|
| ToF LiDAR | NIR | Velodyne, IBEO, Valeo, Hesai, Robosense | Innoviz, Hesai, Robosense, AEye | Continental, Xenomatix, Ouster | Quanergy | NEPTEC TriDAR |
| | SWIR | Luminar | AEye, Hesai | Argo (Princeton Lightwave) | -- | -- |
| FMCW LiDAR | | -- | Aeva | -- | Cruise (Strobe) | Aurora (Blackmore, 1,550 nm) |

## 2.1.1   Current Status of Automotive LiDAR

LiDAR technology in the automotive sector is advancing through a variety of sensor types, each harnessing distinct technologies as showcased in Table 2.1. Many companies such as Innoviz, Continental, and Quanergy are focusing on the development of various LiDAR technologies, including Mechanical Spinning, MEMS, Flash, and OPA LiDAR.

Mechanical Spinning LiDAR, utilized by companies like Velodyne, Valeo, Ouster, Hesai, and Robosense, employs rotating mirrors or the entire sensor unit to achieve a 360-degree field of view, commonly using Near Infrared wavelengths (NIR, 750-1000 nm wavelength) for Time-of-Flight (ToF) applications. MEMS LiDAR, adopted by Innoviz, Hesai, Robosense, and AEye, uses micro-electro-mechanical systems to steer the laser beam, offering a compact and cost-effective solution for both NIR and Short-Wave Infrared wavelengths (SWIR, 1000-2500 nm wavelength). Flash LiDAR, applied by Continental, Xenomatix, and Ouster, uses a broad laser pulse to illuminate the entire scene and capture real-time 3D data, also typically employing NIR wavelengths. OPA (Optical Phased Array) LiDAR, utilized by Quanergy and Cruise (Strobe), relies on electronically steering the laser beam with no moving parts, providing robustness and adaptability. Additionally, companies like Aurora (Blackmore) are advancing FMCW (Frequency-Modulated Continuous Wave) LiDAR technology, which operates at a 1550 nm wavelength, measuring frequency shifts to determine both distance and velocity with high sensitivity and long-range detection capabilities. These diverse technologies highlight the industry commitment to enhancing LiDAR performance across various applications. Below, we provide an overview of the principles, advantages, disadvantages, and commercialization/applications or prototypes of common LiDAR types.

**Mechanical Spinning LiDAR** [49] utilizes a rotating assembly of mirrors or the entire sensor unit to sweep laser beams across the environment, providing a 360-degree field of view. This well-established technology is known for its high resolution and comprehensive scanning capabilities, making it ideal for applications requiring detailed environmental mapping [49]. However, the size and cost of these systems, along with the wear and tear on moving parts, present challenges for widespread adoption, especially in consumer markets like autonomous vehicles.

**MEMS (Micro-Electro-Mechanical Systems) LiDAR** [49] employs tiny mirrors controlled by electrostatic or electromagnetic forces to steer the laser beam. This method offers significant advantages in terms of compact size, lower cost, and rapid beam adjustment, enabling fast scanning. Despite these benefits, MEMS LiDAR typically has a limited detection range and a narrower FOV compared to mechanical systems [50]. Its smaller components make it more suitable for integration into compact, cost-sensitive applications, such as automotive systems.

**Flash LiDAR** [49] operates by illuminating the entire scene with a broad laser pulse, akin to a camera flash, and using a focal plane array of sensors to capture the reflected light. This approach allows for real-time 3D imaging of the environment without the need for moving parts, leading to higher reliability and compactness [51]. However, flash LiDAR is generally limited to shorter ranges and lower resolutions due to the diffuse nature of the laser pulse, which distributes energy over a larger area.

**OPA (Optical Phased Array) LiDAR** [49] represents a cutting-edge technology where an array of laser emitters, with individually controlled phases, steer the laser beam electronically. This method eliminates mechanical components, resulting in a highly reliable and durable system. OPA LiDAR is capable of quickly and precisely directing the laser beam, making it highly adaptable and scalable for various applications [51]. Despite being a relatively new technology, its potential for low-cost, high-volume production and advanced scanning capabilities positions it as a promising solution for future autonomous vehicle applications.

**Frequency-Modulated Continuous Wave (FMCW) LiDAR** [52,53], which utilizes coherent detection, is increasingly attracting attention from automakers and investors due to its advanced capabilities over traditional Time-of-Flight (ToF) LiDAR systems [53]. FMCW LiDAR measures both distance and velocity by analyzing the frequency shift between the emitted and reflected light waves, providing more precise and detailed environmental mapping. This technology offers significant benefits, including immunity to interference from other light sources, enhanced sensitivity, and the ability to operate at longer, eye-safe wavelengths, which allow for higher power emissions and greater detection ranges [52]. Additionally, FMCW LiDAR systems are well-suited for integration with photonic integrated circuits (PIC), facilitating the development of compact, cost-effective, and scalable solutions ideal for mass production in automotive and industrial applications. Start-ups such as Strobe and Blackmore, specializing in FMCW technology, were rapidly integrated into larger entities like Cruise and Aurora. The diverse landscape of automotive LiDAR manufacturers, along with their technologies, reflects the LiDAR-based approaches to enhancing vehicular sensor systems.

**SPAD (Single Photon Avalanche Diode) arrays** [54] in Geiger mode have been employed to extend detection ranges. They are highly sensitive photodetectors capable of detecting single photons, making them ideal for low-light conditions. In Geiger mode, SPAD operate above their breakdown voltage, causing them to avalanche and produce a detectable pulse for each photon they absorb. This mode significantly enhances their sensitivity, allowing for the detection of very weak light signals over longer distances.

SPAD arrays in Geiger mode are particularly useful in LiDAR systems for several reasons [54]. First, their high sensitivity allows LiDAR systems to detect low-intensity light reflected from distant objects, thereby increasing the maximum detection range. Second, SPAD have very fast response times, enabling rapid data acquisition and high temporal resolution, which are crucial for creating accurate 3D maps in real-time. Third, operating in Geiger mode ensures that each photon event is clearly identified, improving the signal-to-noise ratio and the overall reliability of the measurements.

The integration of SPAD arrays in LiDAR systems has facilitated advancements in various applications, including autonomous vehicles, where extended detection range and high precision are essential for safe applications. The Ouster OS1 LiDAR series utilize CMOS-based SPAD for its 850-nm laser detection [55], while the Toyota prototype and Princeton Lightwave (Argo.ai) SPAD LiDAR prototype represent further explorations into this technology, albeit with limited disclosed details.

## 2.2 3D LiDAR Datasets for Autonomous Driving

Datasets are crucial for the swift advancement of applications and utilization of 3D data through deep learning networks. Multiple autonomous driving task datasets provide 3D LiDAR data for outdoor environments (Table 2.2). These datasets not only offer a high vertical resolution (Section 2.2.1) for detailed environmental representation but also address inherent challenges such as the rolling shutter effect, which distorts data in dynamic scenarios. Furthermore, the robustness of these datasets under adverse weather conditions is essential for ensuring consistent sensor performance, underscoring the importance of including diverse environmental data. In summary, the LiDAR resolution, distortion mitigation, weather resilience, and data diversity, are all crucial for developing reliable and adaptable algorithms for 3D geometric and semantic scene understanding in autonomous driving applications.

### 2.2.1 High Vertical Resolution

The vertical resolution of 3D LiDAR refers to the density of laser beams projected in the vertical dimension, indicating how finely the LiDAR system can discern features at different heights. Higher vertical resolution means the system can capture more detailed and precise measurements of objects and terrain by sending and receiving a greater number of laser pulses over vertical angles. This allows for a more nuanced 3D representation of the environment, essential for applications requiring detailed spatial awareness, such as autonomous driving, aerial mapping, and environmental monitoring.

High vertical resolution LiDAR is not present in existing autonomous driving datasets (see Table 2.2). The vertical resolution of LiVi-Set [60] and nuScenes [5] is 32 channels, while the ONCE dataset [63] features a vertical resolution of 40 channels. In addition, the Stanford Track Collection [65], Waymo Open Dataset (WOD) [37], Argoverse 2 [56], KITTI [15], Sydney Urban Objects [66], DENSE [57], H3D [58], SemanticKITTI [3], Lyft Level 5 [61,62] and KITTI-360 [59] is 64 channels. In contrast, our proposed dataset has a higher vertical resolution of 128 channels, which can capture a significantly higher level of detail of environment objects (Figure 3.1).

| Dataset | Resolution | Range/m | Diversity | Image | # Frames | Other sensors |
|---|---|---|---|---|---|---|
| Argoverse 2 [56] | _64[a]_ | **200** | E/W | I | 20k | B |
| DENSE [57] | _64_ | _120_ | E/W/T | I | _1M_ | D/M/F/T/B |
| H3D [58] | _64_ | _120_ | E | I | 28k | G/M |
| KITTI \| SemanticKITTI \| KITTI-360 [3, 15, 59] | _64_ | _120_ | E | I | 93k\|93k\|320k | _N/S/G/M/B_ |
| LiVi-Set [60] | 32 | 100 | E | I | 10k | |
| Lyft Level 5 [61, 62] | _64_ | **200** | E/W/T | I | 170k | D/B |
| nuScenes [5] | 32 | 100 | E/W/T | I | _1M_ | M/D/B |
| ONCE [63] | 40 | **200** | E/T | I | _1M_ | B |
| Oxford RobotCar [64] | _4[b]_ | 50 | E/W/T | I | **3M**[c] | _N/S/G/M/B_ |
| Stanford Track Collection [65] | _64_ | _120_ | E | I | 14k | M |
| Sydney Urban Objects [66] | _64_ | _120_ | E | I | 0.6k[d] | |
| Waymo Open Dataset (WOD) [37] | _64_ | 75 | E/T | I | 390k | B |
| **DurLAR ([17], ours)** | **128** | _120_ | **E/W/T/L** | **I/A/R** | 100k | **U/N/S/G/M/B** |

**Table 2.2: Existing public LiDAR datasets (bold/underlined represents best/2nd best)** for autonomous driving tasks detailing vertical resolution (# channels), diversity in terms of environments (E), times of day (T), weather conditions (W), same route of repeated locations (L) and also the type of LiDAR images made available in addition to range information as: intensity (I), ambient (A), reflectivity (R). Other sensors refer to radar (D), lux meter (U), GNSS supporting more than 2 constellations (N), INS (S), GPS (G), IMU (M), FIR camera (F), Near-infrared camera (T) and stereo camera (B). [a] the pseudo 64-beam LiDAR sweep are aggregated from the two stacked 32-beam sensors into a single sweep. [b] the number of planes. SICK LD-MRS LiDAR has 4 planes, and SICK LMS-151 LiDAR has 1 plane. [c] the number of LD-MRS LiDAR frames. [d] the number of individual scans of objects.

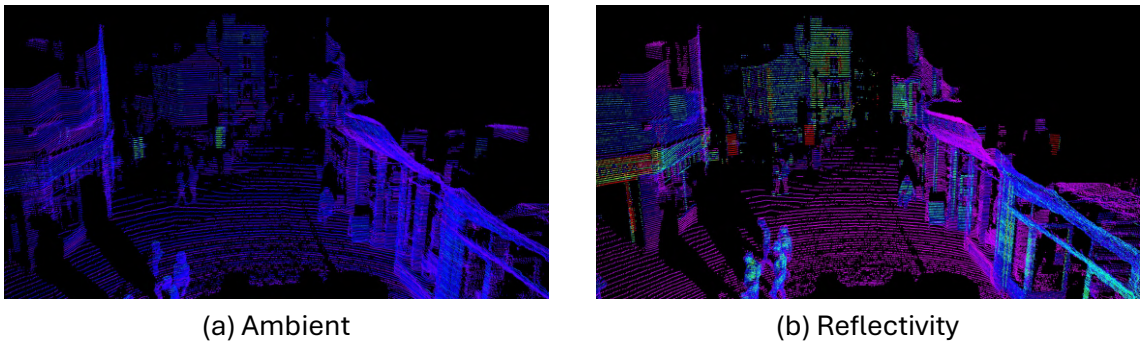

(a) Ambient                     (b) Reflectivity

**Figure 2.2: Comparison of ambient and reflectivity imagery** derived from LiDAR.
**(a) Ambient**: the colors represent the intensity of the ambient near-infrared light, with brighter colors indicating higher intensity and darker colors indicating lower intensity.
**(b) Reflectivity**: the colors represent the reflective properties of various surfaces within the scene. Brighter colors indicate surfaces with higher reflectivity, while darker colors represent less reflective surfaces.

### 2.2.2 Panoramic Imagery Derived from LiDAR

Panoramic imagery derived from LiDAR technology offers a revolutionary approach to capturing and understanding the environment in autonomous driving applications. Utilizing LiDAR, we are able to generate both panoramic ambient imagery and panoramic reflectivity imagery (Figure 2.2), each providing unique advantages for enhancing visibility and detail under various conditions. Note that we are the first to include ambient and reflectivity imagery in the dataset (Chapter 3).

**Panoramic ambient imagery**, as shown in Figure 2.2 (a), is the 360-degree images that capture ambient light conditions in the near-infrared spectrum. This type of imagery provides comprehensive visibility of the environment, even in low light conditions, which is particularly beneficial for autonomous driving applications.

In our DurLAR dataset (Chapter 3), panoramic ambient imagery is captured using near-infrared light with wavelengths between 800-2500 nm. This allows the system to produce images that can be effectively used day and night, offering visibility and environmental detail even under adverse lighting conditions. The use of a photon-counting ASIC (Application-Specific Integrated Circuit) sensor with high illumination sensitivity ensures that these images can be captured in low light environments, making it highly practical for designing autonomous driving systems that need to operate in diverse and challenging lighting scenarios.

**Panoramic reflectivity imagery**, as shown in Figure 2.2 (b), is the 360-degree images that capture the reflective properties of surfaces in the environment using LiDAR technology. This reflectivity feature can be regarded as a range-normalized intensity feature. By normalizing the intensity based on the range, we mitigate the impact of distance on the intensity values of objects made of the same material. This normalization process effectively eliminates the significant changes in intensity that occur due to varying ranges (distances), thereby more accurately reflecting the material properties of the object surface.

Reflectivity imagery is particularly useful because it is consistent across different lighting conditions and distances. Unlike ambient imagery, which captures the intensity of ambient light and can vary significantly with changes in illumination and ranges, reflectivity imagery represents an intrinsic property of the objects being scanned. This means that the reflective properties of a surface will appear the same regardless of external factors like lighting and weather conditions.

### 2.2.3 Rolling Shutter Effect

The rolling shutter effect (Figure 2.3), a phenomenon prevalent in a wide array of imaging and sensing devices, particularly affects analogue spinning Light Detection and Ranging (LiDAR) systems such as those developed by Velodyne [15]. This effect arises due to the sequential capturing of image lines rather than the simultaneous acquisition of the entire scene, leading to distortions when either the sensor or the objects within the scene are in motion [67]. This can result in geometric distortions such as skewed or curved objects in dynamic scenes, especially during fast sensor or object motion. Such distortions can significantly impact the fidelity and accuracy of the data collected by these LiDAR systems, posing challenges for applications that rely heavily on precise spatial measurements and reconstructions, including autonomous driving, 3D mapping,

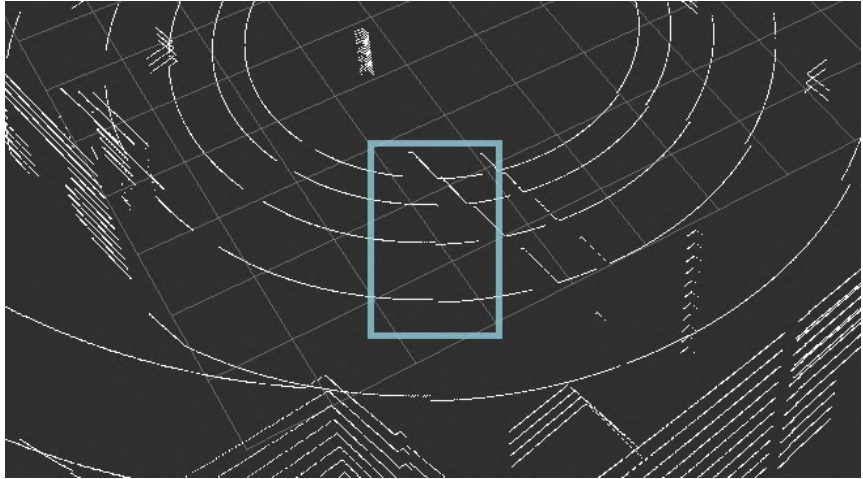

**Figure 2.3: The rolling shutter effect** observed in the real-world point cloud data from a Velodyne VLP-16 LiDAR. Image courtesy of Sobczak *et al.* [2].

and environment monitoring.

As shown in Figure 2.3, the rolling shutter effect observed in the point cloud data from a Velodyne VLP-16 LiDAR [2] is minor when driving around a laboratory environment. This minimal impact is due to the limited space and low achievable speeds in indoor settings. However, the effect becomes significantly pronounced in high-speed outdoor autonomous driving scenarios. Therefore, it is crucial to account for the rolling shutter effect in such conditions to ensure accurate data interpretation and system performance.

In the context of autonomous driving and related research, analogue spinning LiDAR sensors have been extensively utilized across a multitude of existing datasets, underscoring their critical role in the development and evaluation of perception algorithms. Notable examples include the KITTI dataset by Geiger *et al.* [15], the Sydney Urban Objects dataset by Quadros *et al.* [66], and the more recent nuScenes dataset [5] among others [3, 57–61]. These datasets, which have become benchmarks in the field, leverage data from Velodyne scanners to provide rich, real-world environments for testing and developing autonomous vehicle systems, robot navigation, and scene understanding technologies.

The pervasive issue of the rolling shutter effect in these datasets highlights the necessity for advanced calibration and processing techniques that can mitigate its impact. Researchers and engineers have developed various methods to correct for or minimize rolling shutter distortions, ensuring that the spatial information captured by LiDAR sensors remains accurate and reliable for downstream applications. One approach is to apply motion correction algorithms using the velocity of the sensor, often estimated through inertial measurements or fusion with camera data. These methods can correct for distortions by compensating for the sensor motion during data acquisition. For example, iterative algorithms [68] can register distorted LiDAR scans to undistorted maps by unwarping them based on inferred motion states. Another common technique [69] involves fusing LiDAR data with high-resolution camera inputs, leveraging Kalman filters to estimate and correct for both ego-motion and object motion.

Furthermore, the solid-state LiDAR addresses this challenge by providing a high update frequency and precise control over the laser beams, which minimizes the impact of the rolling shutter effect. By eliminating moving parts, solid-state LiDAR systems reduce the latency and improve the synchronization of data capture, leading to more accurate and reliable 3D point cloud [51]. This capability is essential for the continuous improvement of LiDAR-based sensing systems, enabling more robust and efficient autonomous perception. As the field progresses, the continuous improvement of LiDAR technologies and the development of robust corrective algorithms are crucial for harnessing the full potential of LiDAR-based sensing in autonomous systems and beyond.

In contrast, the Ouster OS1-128 LiDAR that we use eliminates temporal mismatch and rolling shutter effects by capturing all depth, intensity, and ambient data layers simultaneously, ensuring perfect spatial correlation [70]. Its digital LiDAR technology employs solid-state components and vertical cavity surface emitting lasers (VCSELs) [55], which fire laser pulses simultaneously rather than sequentially, preventing motion-induced distortions. Additionally, the OS1-128 high vertical angular resolution and uniform spacing throughout its FOV further enhance data accuracy and reliability. These features make it ideal for real-time applications such as autonomous driving and mapping.

### 2.2.4 Overcoming Adverse Weather

In adverse weather conditions, such as rain, fog, snow, and dust, the performance of LiDAR systems is significantly compromised. The presence of opaque particles in these conditions distorts and scatters the LiDAR light pulses, leading to reduced visibility and increased transmission loss [71]. Specifically, adverse weather not only increases the attenuation of the LiDAR signals but also weakens the reflectivity of objects in the environment, resulting in a diminished return signal. The reduction in received signal energy directly impacts the ability of LiDAR system to accurately generate fine-grained point clouds, which are crucial for autonomous driving applications and environmental sensing.

To mitigate the effects of adverse weather on LiDAR performance, researchers have explored a variety of strategies. Traditionally, datasets aimed at supporting autonomous vehicle development have incorporated radar technology due to its ability to penetrate fog, rain, and other particulate-filled environments [5,61,72]. Although radar systems offer an advantage in such conditions, they typically provide much lower spatial resolution compared to LiDAR, limiting their effectiveness for detailed environmental mapping and object detection tasks.

One promising solution for enhancing LiDAR performance in harsh weather conditions involves the use of Short-Wavelength Infrared (SWIR) lasers, such as those operating at wavelengths around 1,550 nm. These SWIR-based LiDAR systems are capable of achieving higher transmission power and, consequently, are less susceptible to atmospheric absorption and scattering. This characteristic allows SWIR LiDAR systems to maintain better performance during adverse weather events, offering a potential solution to the limitations faced by traditional LiDAR technologies.

Our contribution to this area of research is the introduction of the DurLAR dataset, which not only includes high-resolution LiDAR point clouds but also provides ambient (near infrared) and reflectivity images [71]. These additional modalities are captured using sensors with extreme sensitivity to low-light conditions, making them particularly robust against the challenges posed by poor illumination and adverse weather. The ambient imagery, capturing near-infrared wavelengths, and reflectivity data, detailing the surface characteristics of objects, enriches the dataset with crucial information that complements the spatial data provided by LiDAR. This multi-modal approach, as summarized in Table 2.2, ensures that the DurLAR dataset is uniquely positioned to support the development of advanced perception systems capable of operating reliably in a wide range of environmental conditions, including those where traditional LiDAR systems falter.

In conclusion, overcoming the challenges posed by adverse weather conditions to LiDAR-based perception is essential for advancing autonomous vehicle technologies and other applications reliant on accurate environmental sensing. Through the integration of radar, SWIR LiDAR technologies, and innovative dataset enhancements such as DurLAR, the research community continues to make strides toward achieving this goal.

### 2.2.5 Data Diversity

Data diversity within any dataset helps the generation of more universally trained models that can operate successfully under a variety of scenarios. Previous work considers the diversity in their dataset curation [5, 57, 61, 64], but fails to collect data under diverse conditions over the same driving route (see Table 2.2), *e.g.*, traffic level, times of day, weathers, *etc.*

### 2.2.6 Ground Truth Depth

While many existing datasets, such as the Stanford Track Collection [65], Sydney Urban Objects [66], Cityscapes [9], Oxford RobotCar [64], LiVi-Set [60], nuScenes [5] and H3D [58], include LiDAR data, they do not provide ground truth depth data directly usable for depth estimation tasks. This limitation arises from several factors related to how LiDAR data is captured and processed.

LiDAR data often suffers from sparsity. A LiDAR sensor typically emits laser pulses in a specific pattern, capturing only discrete points in space rather than continuous depth information. This results in point clouds with varying density, which may not cover all surfaces and objects in the environment uniformly (Section 2.2.1). Consequently, the resulting data may be too sparse to serve as ground truth for dense depth estimation models, which require comprehensive and consistent depth information across the entire scene.

LiDAR data is subject to noise and inaccuracies due to environmental factors. Factors such as surface reflectivity, the presence of transparent or absorptive materials (Section 2.2.2), and weather conditions (Section 2.2.4) can introduce errors in the measured distances. These inaccuracies can degrade the quality of the depth information, making it unreliable as ground truth without extensive preprocessing and manual annotating.

The alignment of LiDAR data with other sensor modalities, such as cameras, is crucial for creating accurate depth maps. This process, known as sensor fusion, requires precise calibration (Section 3.5) to ensure that the LiDAR points correspond correctly to the pixels in the camera images. Misalignments can lead to incorrect depth values, further complicating the use of raw LiDAR data as ground truth.

The temporal aspect of LiDAR data collection poses challenges. LiDAR sensors capture the environment over time, and any movement by the vehicle or objects within the scene can introduce motion distortions, known as the rolling shutter effect (Section 2.2.3). This effect can distort the spatial relationships within the point cloud, complicating the extraction of accurate depth information.

On the contrary, we introduce the DurLAR dataset (Chapter 3), which includes high-resolution ground truth depth data captured with a 128-channel high-fidelity LiDAR. This dataset addresses the limitations of existing datasets [5, 9, 58, 60, 64–66] by providing detailed and accurate depth information (Section 2.2.1), and precise multi-sensor calibrations (Section 3.5). The inclusion of ground truth depth in DurLAR enhances the training and evaluation of depth estimation models, improving their accuracy and robustness. This is particularly relevant in our following research as it allows us to develop and benchmark contemporary depth estimation methodologies (Section 3.6) that leverage this high-fidelity data to achieve better performance in monocular depth estimation and other future perception tasks.

## 2.3 Point Cloud Representations and Embeddings

Point Cloud Representations and Embeddings serve as the cornerstone for a myriad of LiDAR-based tasks, underpinning the advancements in how machines perceive and interact with the 3D world. The rich, spatially encoded data obtained from LiDAR sensors can be intricately represented in two primary forms: point-based and voxel-based representations [19, 20, 38, 46, 47, 73, 74]. Each representation offers distinct advantages tailored to leverage the unique characteristics of LiDAR data, catering to specific application requirements.

Point-based representations maintain the raw, unstructured nature of LiDAR data, capturing the essence of the environment with high fidelity. This form has been extensively explored, with notable methods such as PointNet and its successors demonstrating remarkable success in extracting features directly from point clouds without the need for pre-processing or rasterization [46, 47]. These methods excel in preserving the geometric details of scenes, making them particularly suited for tasks requiring precise localization and object identification.

On the contrary, voxel-based representations offer a paradigm shift by structuring the inherently sparse and irregular LiDAR data into a regular grid format [20, 75–77]. This approach simplifies the processing pipeline by transforming the data into a format more amenable to conventional 3D convolutional operations, thereby facilitating a range of applications from object detection to scene segmentation. However, the discretization process inherent in voxelization may lead to the loss of critical details, especially when dealing with fine structures or distant objects.

To bridge the gap between these representations, extensive research [20, 21] has been conducted on developing neural architectures capable of efficiently processing voxelized data. Feedforward Neural Network (FNN) transforms pointwise features into voxel-wise embeddings by aggregating points within each voxel [19, 20, 38]. This process involves voxelization, where the 3D space is divided into a grid of voxels, and each voxel aggregates the features of all contained points [20]. FNN is particularly suitable for this purpose because it process information in a straightforward manner from input to output without the complexities of recurrent or feedback connections. This straightforward processing is ideal for aggregating and transforming data within each voxel. The structured representation created by voxelizing point clouds facilitates the application of convolutional operations typical in 3D CNN, which rely on regular grid structures to perform efficient and effective computations [78]. Other neural network architectures, such as recurrent neural networks (RNN) [79, 80] or networks with complex feedback loops, are less suited for this task because they introduce unnecessary complexity and are designed for sequential or cyclic data processing, rather than the direct spatial aggregation required in voxelization. Despite its widespread use, FNN often struggles with maintaining the integrity of the original detail of point cloud. The compression and aggregation process can dilute fine-grained features, while the volumetric nature of voxel data poses computational challenges due to its high dimensionality and sparsity [21, 81].

To address these limitations, our work (Chapter 5) introduces a novel approach that harmonizes the strengths of both point and voxel-based methodologies. By partitioning the point cloud into voxel grids and employing a class-aware double nested Autoencoder (AE) enhanced with self-attention mechanisms, we offer a robust solution [21]. This innovative strategy ensures the retention of detailed information by concentrating on the local structural nuances within each voxel grid. Moreover, it employs a high-dimensional feature compression technique, significantly ameliorating computational efficiency. Our method not only preserves the granularity of point-based representations but also leverages the structured nature of voxel-based approaches, setting a new precedent for processing LiDAR data with enhanced detail preservation and operational efficiency.

## 2.4 Monocular Depth Estimation

Monocular depth estimation represents a pivotal challenge in computer vision, aiming to reconstruct a detailed depth map for each pixel from a single RGB image. This task is fundamental for various applications, including autonomous driving [82, 83], augmented reality, and robotics [83], where understanding the three-dimensional structure of the environment from a single viewpoint is crucial.

**Self-supervised methods** have emerged as a powerful strategy for training depth estimation models without the need for explicitly labeled depth maps. These methods exploit consistency constraints within monocular RGB image sequences [1, 82–85, 85–89], stereo image pairs [90–93], or synthetic data [94, 95]. By harnessing temporal dynamics in multi-frame architectures [1, 83, 96–101], these approaches leverage temporal information to refine depth predictions dynamically. This adaptation requires complex calculations across multiple frames, significantly increasing the computational demand.

Moreover, the advent of multi-view stereo (MVS) techniques [1, 102–109] has introduced capabilities for depth estimation from unordered image collections [1], which means the images do not need to be captured in a specific sequence or from a fixed set of stereo pairs. These self-supervised MVS methods utilize cost volumes to assimilate sequences of frames, allows for greater flexibility in acquiring images from various viewpoints without the constraint of predefined camera arrangements/poses [1].

**Supervised methods**, on the other hand, directly leverage ground truth depth data from depth sensors such as LiDAR [89, 110–113] and RGB-D cameras [114, 115]. These methods benefit from the precise depth information these sensors provide, enabling the training of highly accurate Convolutional Neural Network (CNN) architectures [114–116]. Following the success of CNN, residual-learning approaches [117–119] have been introduced to model the transformation between color images and their depth maps, exploiting deeper networks for enhanced accuracy. Nonetheless, the efficacy of these supervised methods is often constrained by the availability and resolution of ground truth depth data, highlighting a critical dependency on high-quality datasets for training.

Recent advancements in monocular depth estimation have introduced state-of-the-art supervised models such as MiDaS [120, 121] and Marigold [122]. MiDaS v3.1 [120] focuses on leveraging various encoder backbones, including vision transformers such as BEiT [123] and SwinV2 [124], alongside convolutional networks, which have significantly improved depth estimation accuracy and runtime efficiency. The MiDaS architecture allows for effective zero-shot cross-dataset transfer, making it particularly valuable for real-world applications where high-quality training data is scarce [120]. On the other hand, Marigold [122], a diffusion-based model, offers a novel approach by repurposing image generators for monocular depth estimation. This method excels in zero-shot transfer capabilities, fine-tuned using synthetic data, and achieves remarkable performance on unseen datasets. Its diffusion-based approach enables highly detailed depth maps with minimal inference time, providing a promising alternative to traditional CNN-based architectures.

In summary, monocular depth estimation continues to evolve with advancements in self-supervised and supervised methodologies. The development of advanced models and algorithms, driven by innovative uses of available data, underscores the progression toward more accurate and efficient depth prediction performance. As the quest for improved performance persists, the integration of diverse data modality and the exploration of novel deep learning frameworks remain at the forefront of computer vision research.

Overall, one of key challenges within contemporary autonomous driving task evaluation is the lack of high fidelity (vertical resolution) depth datasets in order to facilitate effective evaluation of geometric scene understanding tasks, such as monocular depth estimation. Based on our high-fidelity DurLAR dataset (Chapter 3), we consider the impact of abundant high-resolution ground truth depth data on three state-of-the-art contemporary monocular depth estimation architectures (MonoDepth2 [85], Depth-hints [92], ManyDepth [1]) through the use of our novel joint supervised/semi-supervised loss formulation (Section 3.7).

## 2.5 Temporal Redundancy

Temporal redundancy is a significant characteristic of both video and radar sequences, underscored by the frequent similarity between neighboring frames, especially in autonomous driving datasets such as KITTI [8], nuScenes [5], and Waymo [37]. This similarity arises due to the minimal temporal interval between successive frames—often as short as 0.1 seconds—resulting in a homogeneous scene diversity across different dataset splits. Such redundancy inherently leads to analogous detection and segmentation performance when employing identical training iterations, as observed in recent studies [125]. To mitigate this issue, prior research efforts [126,127] have adopted uniform sampling techniques across the entire Waymo training set to generate varied fine-tuning splits, exemplified by strategies such as selecting every alternate frame to achieve a 50% data subset.

In the domain of semi-supervised 3D LiDAR segmentation, prevalent methodologies [4, 18] typically rely on a passive, uniform sampling approach to sift through unlabeled points within a fully-labeled point cloud dataset. Conversely, active learning frameworks endeavor to efficiently navigate through this redundancy. They aim to minimize necessary annotation or training efforts by strategically choosing informative and diverse sub-scenes for labeling, thus effectively leveraging the underlying data redundancy [128–130]. These frameworks underscore the potential of reducing manual labeling workload and improving model performance by focusing on variably informative data points, thereby aligning with the objectives of efficient data utilization and enhanced learning efficacy.

Building upon these insights, we introduce an innovative temporal-redundancy-based sampling strategy. This approach is designed to not only retain the time efficiency comparable to that of uniform sampling but also significantly reduce inter-frame spatio-temporal redundancy. By maximizing data diversity through our proposed method, we aim to enhance the quality of the sampled dataset, ensuring that the training process benefits from a broader spectrum of scenarios and conditions present within the data. This strategy is poised to contribute substantially to the field of autonomous driving, where leveraging every nuance of the captured data can lead to significant improvements in perception systems. Our method aligns with the ongoing efforts to refine data processing and utilization strategies, thereby facilitating more effective and efficient training processes for autonomous driving models, and potentially setting a new benchmark in the management and utilization of temporally redundant data.

## 2.6 Invariant Features

The rapid advancement in 3D sensing technologies has led to an exponential increase in the use of point cloud data. A critical challenge in leveraging point cloud data effectively is developing methods that are invariant to transformations, particularly rotations and translations [34, 131, 132].

Invariant features refer to properties or characteristics of the data that **remain unchanged under transformations** such as rotation [32, 131–134], translation [135], or reflection [34, 35]. By extracting invariant features, algorithms can recognize and understand the underlying structure of the data without being affected by how it is presented. This capability is essential for tasks that rely on accurate object recognition, localization, and mapping, as it enables consistent performance regardless of changes in the viewpoint or configuration of the sensors collecting the data.

### 2.6.1 Transformation-Invariant Features

Yu *et al.* [131] introduce a rotation-invariant transformer for point cloud matching, proposing an architecture that achieves extrinsic rotation invariance by learning to describe local patches from rotation-variant inputs. Similarly, Jiang *et al.* [132] propose a Center-aware Feature (CF) descriptor that possesses transformation-invariance properties. Liu *et al.* [133] propose efficient global point cloud registration by matching rotation invariant features, leveraging branch-and-bound (BnB) optimization for global registration. Xu *et al.* [134] propose a novel method based on learning strictly rotation-invariant local feature descriptors for point cloud patches. This approach is crucial for tasks requiring high precision in feature matching and object recognition. Kim *et al.* [32] introduce a representation for 3D point cloud classification that is rotation-invariant, leveraging a graph convolution network for contextual relationship reasoning. Melia *et al.* [136] introduce a rotation-invariant feature that struggles with generalization across diverse point cloud densities and scales due to its computational cost and susceptibility to outdoor noise. Long *et al.* [135] introduce an unsupervised point cloud pre-training method using transformation invariance in clustering, which is an unsupervised representation learning scheme leveraging transformation invariance for point cloud pre-training. This method aims to enhance the model ability to recognize and classify point cloud data without extensive labeled datasets.

### 2.6.2 Isometry Invariant Features

Isometry invariant features are specific types of features used in the analysis of geometric data, such as point clouds, that remain unchanged under isometric transformations. These transformations include **rotations, translations, and reflections**, which preserve distances and angles within the point-wise data [34, 35]. Isometry invariant features are particularly valuable in computer vision, chemistry, and similar fields where the exact positioning and orientation of objects may vary, but their fundamental geometric properties do not.

Pointwise Distance Distribution (PDD) captures the local context of each point in a unit cell by enumerating distances to neighboring points in order. It is an isometry invariant proposed by Widdowson & Kurlin [35] to resolve the data ambiguity for periodic crystals, demonstrated through extensive pairwise comparisons across atomic 3D clouds from high-level periodic crystals of periodic structures [33–35]. Though the effectiveness of PDD in periodic crystals and atomic clouds has been proved by the aforementioned studies, to date, no work has applied PDD features to outdoor 3D point clouds. In outdoor settings, common invariant features [136–139] often face issues due to the irregular and sparse nature of the data, which is compounded by increased noise and environmental complexities [137,139]. Furthermore, the computational demands make them less suitable for the vast scale of outdoor settings [136,138].

Recognizing these limitations, we identify an opportunity to leverage the PDD features in a new domain, where representing the local context of neighboring points in a transformation invariant and geometrically robust manner is crucial. Drawing from the advantages of the PDD design, we propose the RAPiD feature, tailored specifically for LiDAR-based point clouds, to capture the localized geometry of neighboring structures.

## 2.7 LiDAR-Based Semantic Segmentation

LiDAR-based semantic segmentation involves the classification of 3D point clouds obtained from LiDAR data [3,5,15,17] into various categories [3] such as buildings, vehicles, roads, vegetation, pedestrian, *etc.*

### 2.7.1 Computationally Efficient Segmentation

Recent advancements in LiDAR segmentation have focused on improving efficiency while maintaining or enhancing segmentation accuracy. These efforts can be categorized based on their contributions, such as novel network architectures, efficient data representations, and advanced augmentation strategies.

**Efficient network architectures** are important in advancing computationally efficient LiDAR segmentation. For instance, the Center Focusing Network (CFNet) [140] leverages the center focusing feature encoding (CFFE) mechanism to explicitly model relationships between LiDAR points and virtual instance centers, significantly enhancing real-time segmentation performance. CFNet also incorporates a fast center deduplication module (CDM) to streamline instance detection, outperforming previous methods in both efficiency and accuracy on SemanticKITTI and nuScenes datasets.

Depthwise Separable Convolution (DSC) [141] consists of a depthwise convolution followed by a pointwise convolution. This structure reduces both model size and com-plexity, making it a more computationally efficient alternative to standard convolution. It is widely employed in mobile applications [142–144] and hardware accelerators [145]. Additionally, depthwise separable convolution serves as a fundamental component of Xception [146], a deep convolutional neural network architecture that achieves state-of-the-art performance on the ImageNet classification task [147] through more efficient model parameterization. In our research (Chapter 4), we extend the concept of DSC by introducing a sparse variant. This new approach retains the efficiency benefits of depthwise separable convolution while incorporating the advantages of sparse convolution for processing spatially-sparse data [75]. By leveraging this novel sparse variant, we aim to enhance the performance and efficiency of 3D semantic segmentation networks, achieving superior results with reduced computational resources (more efficient memory usage and computational load).

**Efficient data representations** are crucial for optimizing LiDAR segmentation. RPVNet [148] introduces a deep and efficient range-point-voxel fusion network that integrates multiple representations of LiDAR data. By combining voxel-based, point-based, and range-based branches within a unified framework, RPVNet achieves superior segmentation accuracy and efficiency. This multi-branch, multi-modal approach enables comprehensive feature extraction while maintaining computational efficiency and minimizing the overhead associated with using multiple sensors. Additionally, the 2DPASS [149] integrates 2D priors to assist semantic segmentation of LiDAR point clouds, enabling a more streamlined and efficient processing pipeline.

**Advanced augmentation strategies** are widely used to reduce the computational burden and data preparation overhead of LiDAR segmentation. Ryu *et al.* introduce the Instant Domain Augmentation (IDA) [150], a novel approach that generates augmented data on-the-fly during training. This technique utilizes domain randomization principles, applying various transformations such as noise injection, rotation, and scaling to LiDAR data to create diverse training samples. By dynamically generating augmented samples, IDA reduces the need for extensive pre-collected and manually augmented datasets, making the training process more efficient. This approach not only lowers the data preparation overhead but also ensures that models are exposed to a richer variety of scenarios, improving their performance in real-world applications. Similarly, LiDAL [129], incorporates inter-frame uncertainty to facilitate active learning, effectively reducing the labeling burden while maintaining high segmentation accuracy.

### 2.7.2 Semi-Supervised Learning (SSL) LiDAR Segmentation

Semi-Supervised Learning (SSL) for LiDAR semantic segmentation involves using a small amount of labeled or weakly-labeled (*e.g.*, scribble annotations [19]) point cloud data in conjunction with a large amount of unlabeled point cloud data during training, thereby leveraging aspects of both fully supervised learning and weak supervision. Numerous approaches have been explored for LiDAR semantic segmentation. Projection-based approaches [18, 42–44, 151–153] make full use of 2D-convolution kernels by using range or other 2D image-based spherical coordinate representations of point clouds. Conversely, voxel-based approaches [4, 18, 20, 154] transform irregular point clouds to regular 3D grids and then apply 3D convolutional neural networks with a better balance of the efficiency and effectiveness. Pseudo-labeling is generally applied to alleviate the side effect of intra-class negative pairs in feature learning from the teacher network [4, 18, 155, 156]. However, such methods only utilize samples with reliable predictions and thus ignore the valuable information that unreliable predictions carry. In our work, we combined a novel SSL framework with the mean teacher paradigm [157], demonstrating the utilization of unreliable pseudo-labels to improve segmentation performance.

### 2.7.3 Fully-Supervised Learning (FSL) LiDAR Segmentation

LiDAR-Based Semantic Segmentation [16, 18–20, 38, 148, 153–156, 158–167] is fundamental for LiDAR-driven scene perception, aiming to label each point in a point cloud sequence. The majority of the approaches [20, 38, 154, 156, 162] solely rely on the point-based features of the point cloud, such as SPVCNN [154] which introduces a point-to-voxel branch, using combined point-voxel features for segmentation. Cylinder3D [20] proposes cylindrical partitioning with a UNet [77] backbone variant. LiM3D [16] utilizes coordinates combined with surface reflectivity attributes. Overall, such prior work relies solely on point-based features such as coordinates and intensity of the points, lacking an effective fusion mechanism, resulting in suboptimal performance [168–170]. They are also susceptible to changes in viewpoint, distance, and point sparsity [171] due to the lack of isometry across all inter-point distances.

## 2.8  Evaluations and Metrics

In the realm of computer vision, particularly in depth estimation and 3D semantic segmentation, systematic analysis of evaluation metrics is crucial for enhancing model performance and applicability. By examining these metrics, researchers can identify the strengths and weaknesses of their models, facilitating iterative improvements to ensure robust and reliable outputs in practical scenarios. Evaluating depth estimation encompasses understanding various aspects of model performance, such as error magnitude and prediction accuracy. Similarly, the evaluation of 3D semantic segmentation focuses on segmentation accuracy across all classes, addressing imbalanced data distributions and ensuring fair performance assessment. These evaluations not only quantify model performance but also guide the development of advanced algorithms and corrective strategies.

**The evaluation of depth estimation methods** is critical to understanding and improving the accuracy and reliability of predictions. We delve into the various metrics employed to assess the performance of depth estimation models. Accurate evaluation metrics are essential as they provide quantitative measures of how closely the predicted depth values align with the ground truth. We introduce principal metrics such as Absolute Relative Error (Abs Rel), Squared Relative Error (Sq Rel), Root Mean Squared Error (RMSE), RMSE log, and Threshold Accuracy $\delta$, each serving a unique purpose in the comprehensive evaluation of model performance.

Consider the notation used in the metrics (Equations (2.1) to (2.5)), where $d_i$ represents the ground truth depth value at the $i$-th pixel, $\hat{d}_i$ denotes the predicted depth value at the $i$-th pixel, and $n$ is the total number of pixels for evaluation. These variables are used to define the principal metrics in depth estimation:

- **Absolute Relative Error** (Abs Rel): the metric in Equation (2.1) calculates the mean absolute error relative to the ground truth depth values. It provides a sense of how much the predicted depth values deviate from the actual values in relative terms.

$$\text{AbsRel} = \frac{1}{n} \sum_{i=1}^{n} \frac{\left| d_i - \hat{d}_i \right|}{d_i}. \tag{2.1}$$

- **Squared Relative Error** (Sq Rel): the metric in Equation (2.2) takes the square of the relative differences between the predicted and ground truth depth values. It is more sensitive to larger errors and penalizes them more heavily, making it useful for identifying significant deviations.

$$\text{SqRel} = \frac{1}{n} \sum_{i=1}^{n} \frac{\left( d_i - \hat{d}_i \right)^2}{d_i}. \tag{2.2}$$

- **Root Mean Squared Error** (RMSE): the metric in Equation (2.3) computes the square root of the average of the squared differences between the predicted and ground truth depths. It gives a general idea of the prediction error magnitude, emphasizing larger errors due to squaring.

$$\text{RMSE} = \sqrt{\frac{1}{n} \sum_{i=1}^{n} \left( d_i - \hat{d}_i \right)^2}. \tag{2.3}$$

- **RMSE log**: the metric in Equation (2.4) is similar to RMSE in Equation (2.3) but applied to the logarithm of the depth values. It reduces the impact of large depth values and provides a balanced evaluation across different depth ranges.

$$\text{RMSE log} = \sqrt{\frac{1}{n} \sum_{i=1}^{n} \left( \log d_i - \log \hat{d}_i \right)^2}. \tag{2.4}$$

- **Threshold Accuracy** $\delta$: the metric in Equation (2.5) evaluates the percentage of predicted depth values that fall within a certain threshold $\delta'$ of the ground truth, where $\mathbb{1}(\cdot)$ represents the indicator function which returns $1$ if the specified condition (the function input) is true and $0$ if it is false. Commonly used thresholds are $\delta' < 1.25$, $\delta' < 1.25^2$, and $\delta' < 1.25^3$. It measures how many predictions are reasonably close to the ground truth and provides an intuitive understanding of the model accuracy.

$$\delta = \frac{1}{n} \sum_{i=1}^{n} \mathbb{1}\left(\max\left(\frac{d_i}{\hat{d}_i}, \frac{\hat{d}_i}{d_i}\right) < \delta'\right).$$ (2.5)

**The evaluation of 3D semantic segmentation methods** relies heavily on the mean Intersection-over-Union (mIoU) metric [3, 172]. mIoU is a standard evaluation measure that quantifies the overlap between predicted and ground truth segments, making it a crucial indicator of segmentation accuracy. It is the average Intersection-over-Union (IoU) across all classes, where IoU for each class is the ratio of the intersection to the union of the predicted and ground truth segments. It measures the accuracy of a model in segmenting an image or point cloud by comparing the predicted segmentation with the ground truth segmentation. This metric is particularly effective in handling imbalanced datasets by giving a balanced representation of performance across all classes.

Specifically, mIoU is calculated in Equation (2.6) as the mean of the IoU for each class,

$$\text{mIoU} = \frac{1}{C} \sum_{i=1}^{C} \frac{\text{TP}_i}{\text{TP}_i + \text{FP}_i + \text{FN}_i},$$ (2.6)

where $C$ is the total number of classes, $\text{TP}_i$ is the number of true positive pixels for class $i$, $\text{FP}_i$ is the number of false positive pixels for class $i$, and $\text{FN}_i$ is the number of false negative pixels for class $i$.

As shown in Figure 2.4, using the ratio of the intersection of the predicted and ground truth areas to their union in the IoU metric effectively measures performance because it captures both false positives and false negatives in a single value. The intersection represents the correctly predicted area (namely the true positives that overlapped), while the union includes all the areas covered by the prediction and the ground truth, thus accounting for false positives (areas incorrectly predicted as the target class) and false negatives (actual target areas missed by the prediction). This comprehensive consideration ensures that the IoU provides a balanced evaluation of the segmentation accuracy, reflecting both the precision (correctness of positive predictions) and recall (completeness in capturing all target areas).

For simplicity, the illustration in Figure 2.4 pertains specifically to the scenario of 2D semantic segmentation. In 3D semantic segmentation task, the metric of mIoU is an extension of the metric used in the 2D image segmentation scenario. The mIoU metric for 3D semantic segmentation measures the overlap between the predicted and ground truth segments in a three-dimensional space, providing a comprehensive evaluation of the segmentation performance in 3D environments.

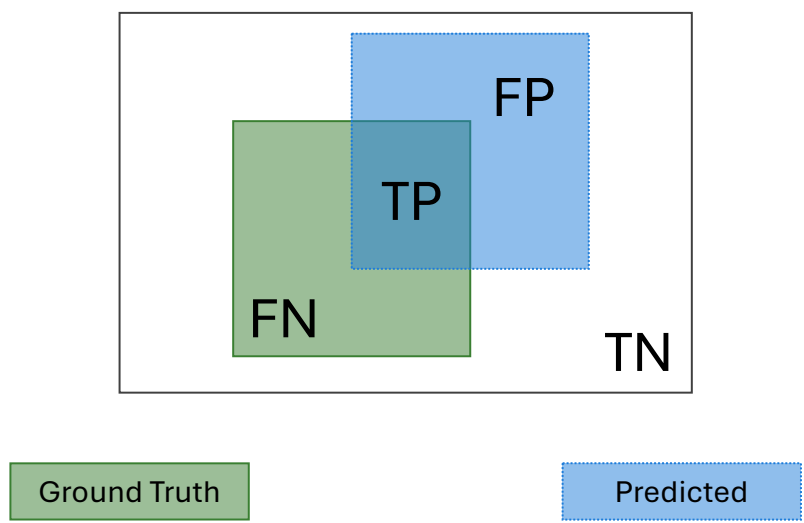

**Figure 2.4: Illustration of IoU calculation** for semantic segmentation. The green region represents the ground truth segmentation, while the blue region represents the predicted segmentation. True Positives (TP) are the overlapping area between ground truth and prediction, False Positives (FP) are the predicted area not overlapping with the ground truth, and False Negatives (FN) are the ground truth area not covered by the prediction. True Negatives (TN) are the areas correctly identified as not belonging to the target class. The IoU is calculated as the ratio of the TP area to the union of TP, FP, and FN areas.

## 2.9   Relevance to Contributions

Based on the overview of the current literature on the variety of subjects presented in this chapter, we will outline the novel contributions of this thesis in the following chapters.

In Chapter 3, we present DurLAR, a high-fidelity 128-channel 3D LiDAR dataset with panoramic ambient (near infrared) and reflectivity imagery. Existing datasets on autonomous driving with high-fidelity LiDAR are very limited (Section 2.2). Our driving platform is equipped with a high resolution 128 channel LiDAR, a 2MPix stereo camera, a lux meter, and a GNSS/INS system. Ambient and reflectivity images are made available along with the LiDAR point clouds to facilitate multi-modal use of concurrent ambient and reflectivity scene information.

In Section 3.6, leveraging DurLAR, with a resolution exceeding that of prior benchmarks (Section 2.2), we consider the task of monocular depth estimation (Section 2.4) and use this increased availability of higher resolution, yet sparse ground truth scene depth information to propose a novel joint supervised/self-supervised loss formulation. We compare performance over both our new DurLAR dataset, the established KITTI benchmark, and the Cityscapes dataset. Our evaluation shows our joint use of supervised and self-supervised loss terms, enabled via the superior ground truth resolution and availability within DurLAR improves the quantitative and qualitative performance of leading contemporary monocular depth estimation approaches (RMSE = 3.639, SqRel = 0.936).

Due to the important role of LiDAR-based semantic segmentation tasks in the field of autonomous driving (Section 2.7), we propose semi-supervised and fully supervised methods to enhance the performance of semantic segmentation tasks. Our semi-supervised method (Chapter 4) mainly aims to achieve faster and more efficient point cloud segmentation, while our fully supervised method (Chapter 5) is designed to achieve more accurate point cloud segmentation.

Specifically, in Chapter 4, we propose a novel pipeline using a smaller architecture that needs fewer ground-truth annotations to outperform current methods in segmentation accuracy. Whilst the availability of 3D LiDAR point cloud data has significantly grown in recent years (Section 2.2), annotation remains expensive and time-consuming, leading to a demand for semi-supervised semantic segmentation methods with application domains such as autonomous driving (Section 2.7.2). Existing work very often employs relatively large segmentation backbone networks to improve segmentation accuracy, at the expense of computational costs (Chapter 4). In addition, many use uniform sampling to reduce ground truth data requirements for learning needed, often resulting in sub-optimal performance. To address these issues, we propose a new pipeline that employs a smaller architecture, requiring fewer ground-truth annotations to achieve superior segmentation accuracy compared to contemporary approaches. This is facilitated via a novel Sparse Depthwise Separable Convolution (SDSC) module that significantly reduces the network parameter count while retaining overall task performance. To effectively sub-sample our training data, we propose a new Spatio-Temporal Redundant Frame Downsampling (ST-RFD) method that leverages knowledge of sensor motion within the environment to extract a more diverse subset of training data frame samples. To leverage the use of limited annotated data samples, we further propose a soft pseudo-label method informed by LiDAR reflectivity. Our method outperforms contemporary semi-supervised work in terms of mIoU, using less labeled data, on the SemanticKITTI (59.5@5%) and ScribbleKITTI (58.1@5%) benchmark datasets, based on a $2.3\times$ reduction in model parameters and $641\times$ fewer multiply-add operations whilst also demonstrating significant performance improvement on limited training data (*i.e.*, Less is More).

We also cover a comprehensive range of topics within the domain of accurate and fully-supervised 3D LiDAR technology (Chapter 5) and its applications in autonomous driving, focusing on critical aspects such as data collection (Sections 2.1 and 2.2), processing (Sections 2.3 and 2.6), and perception (Section 2.7). These subsection of the literature review is aligned with and form the basis for the novel contributions of Chapter 5. The introduction of the RAPiD features ensures robustness to transformations and viewpoints through isometry-invariant metrics (Section 2.6), while the RAPiD embedding method with RAPiD AE optimizes high-dimensional feature embeddings (Section 2.3). Additionally, the novel open-source network architecture RAPiD-Seg achieves SOTA performance in LiDAR segmentation, demonstrating significant advancements in LiDAR segmentation accuracy (Section 2.7.3). These contributions are deeply informed by and build upon the discussed literature, underscoring their relevance and impact in the field of autonomous driving.

DurLAR: A High-Fidelity LiDAR Dataset

Portions of this chapter have previously been published in the following peer-reviewed publication [17]:

- **Li, L.**, Ismail, K. N., Shum, H. P., & Breckon, T. P., "Durlar: A High-Fidelity 128-Channel LiDAR Dataset with Panoramic Ambient and Reflectivity Imagery for Multi-Modal Autonomous Driving Applications." In International Conference on 3D Vision (3DV). IEEE, 2021.

In this chapter, we present DurLAR, a high-fidelity 128-channel 3D LiDAR dataset with panoramic ambient (near infrared) and reflectivity imagery, as well as a sample benchmark task using depth estimation for autonomous driving applications. Our driving platform is equipped with a high resolution 128 channel LiDAR (see Figure 3.1), a 2MPix stereo camera, a lux meter and a GNSS/INS system. Ambient and reflectivity images are made available along with the LiDAR point clouds to facilitate multi-modal use of concurrent ambient and reflectivity scene information. Leveraging DurLAR, with a resolution exceeding that of prior benchmarks, we consider the task of monocular depth estimation and use this increased availability of higher resolution, yet sparse ground truth scene depth information to propose a novel joint supervised/self-supervised loss formulation. We compare performance over both our new DurLAR dataset, the established KITTI, and Cityscapes dataset. Our evaluation shows our joint use of supervised and self-supervised loss terms, enabled via the superior ground truth resolution and availability within DurLAR improves the quantitative and qualitative performance of leading contemporary monocular depth estimation approaches (RMSE = 3.639, Sq Rel = 0.936).

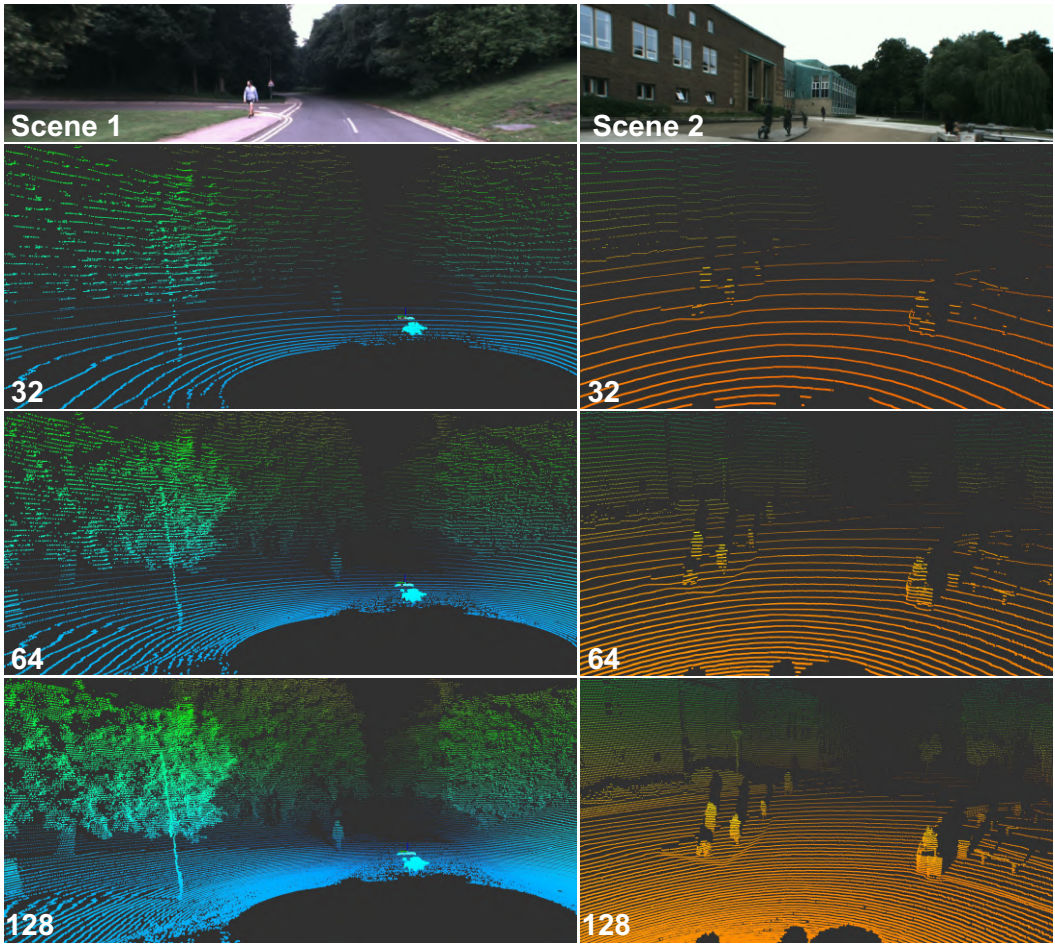

**Figure 3.1: LiDAR point clouds from two exemplar scenes** with differing vertical LiDAR resolution (top to bottom: color RGB images, [32 → 64 → 128] LiDAR channels).

## 3.1 Introduction

LiDAR (Light Detection and Ranging) is one of the core perception technologies enabling future self-driving vehicles and ADAS. Multiple datasets featuring LiDAR have been proposed to evaluate semantic in geometric scene understanding tasks such as depth estimation, object detection, visual odometry, optical flow and tracking [5, 15, 57, 58, 60–62, 64–66]. Based on this existing dataset provision, various architectures have been proposed for LiDAR based scene understanding in this domain [72, 85, 112, 173–177]. Moreover, benchmarks and evaluation metrics have emerged to facilitate the comparison of varies techniques and datasets [3, 8, 178–180].

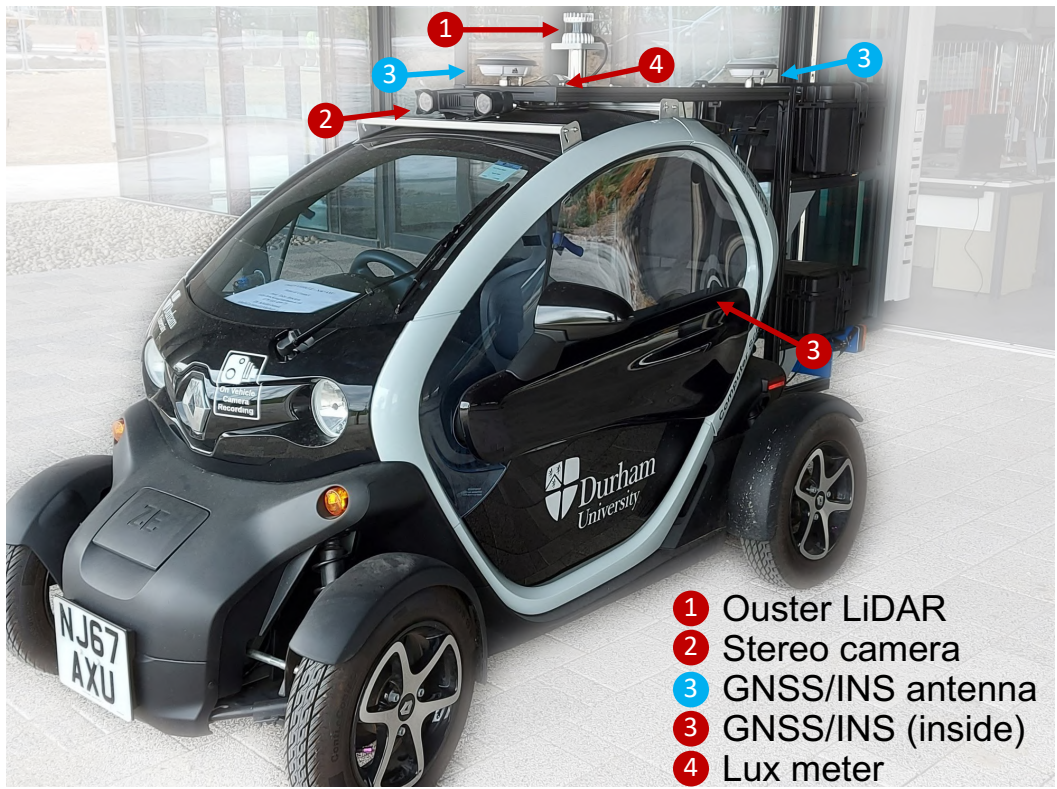

**Figure 3.2: Test vehicle** (Renault Twizy): equipped with a long-range stereo camera, a LiDAR, a lux meter and a combined GNSS/INS inertial navigation system.

In these datasets [3, 8, 178–180], LiDAR range data corresponding to the color image of the environment is provided as the ground-truth depth information. Such ground truth can be relatively sparse compared to the sampling of the corresponding color camera imagery — typically as low as 16 to 64 channels of depth (see Figure 3.1, *e.g*, 16-64 horizontal scanlines of depth information, spanning 360 degrees from the vehicle over a 50-200 m range). Here, the terminology *channel* refers to the vertical resolution of the LiDAR scanner, and has a one-to-one correspondence to the laser beam as it is referred to in some studies. With this in mind, current datasets and their associated metric-driven benchmarks [3, 5, 8, 37] are significantly limited (Section 2.2) when compared to the contemporary availability of high-resolution LiDAR data (Section 2.2.1) as we pursue in this chapter.

On the contrary, we propose a large-scale high-fidelity LiDAR dataset[1] based on the use of a 128 channel LiDAR unit mounted on our Renault Twizy test vehicle (Figure 3.2).

Subsequently, our dataset is presented in a KITTI-compatible format [15], ensuring that the data can be parsed using both our DurLAR development kit and the official KITTI tools, as well as third-party KITTI tools.

Compared to existing autonomous driving task datasets (Table 2.2), DurLAR has the following novel features:

- **High vertical resolution LiDAR** with 128 channels, which is twice that of any existing datasets (Table 2.2), full $360°$ depth, range accuracy to $\pm 2\,\mathrm{cm}$ at 20-50 m.

- **Ambient illumination (near infrared) and reflectivity panoramic imagery** (Section 2.2.2) are made available in the Mono16 format ($2048 \times 128$ resolution), with this being only dataset to make this provision (Table 2.2).

- **Zero temporal mismatch or shutter effects**, as our flash LiDAR captures all 128 channels simultaneously, and the data layers are perfectly spatially correlated [70].

- **Ambient illumination data** is recorded via an onboard lux meter, which is again not available in previous datasets (Table 2.2).

- **High-fidelity GNSS/INS** available via an onboard OxTS navigation unit operating at 100 Hz and receiving position and timing data from multiple GNSS constellations in addition to GPS.

- **KITTI data format** adopted as the *de facto* dataset format such that it can be parsed using both the DurLAR development kit and existing KITTI-compatible tools.

- **Diversity over repeated locations** such that the dataset has been collected under diverse environmental and weather conditions over the same driving route with additional variations in the time of day relative to environmental conditions (*e.g.* traffic, pedestrian occurrence, ambient illumination, see Table 2.2).

---

[1]Access to the dataset: `https://github.com/l1997i/DurLAR`. Refer to Appendix B for more details.

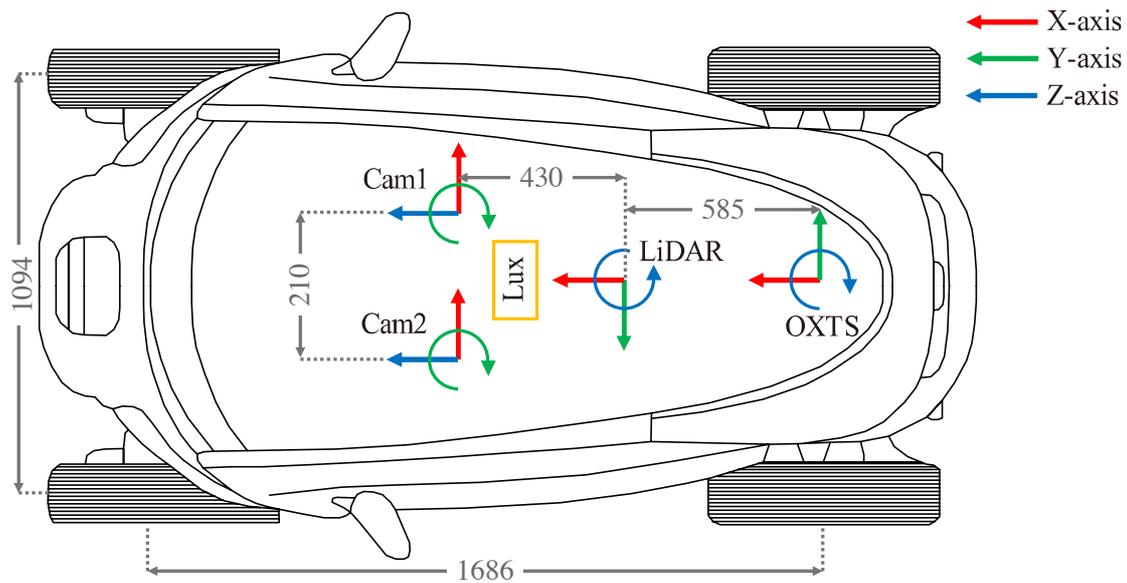

**Figure 3.3: Sensor placements**, top view. All coordinate axes follow the right-hand rule (sizes in mm).

## 3.2 Sensor Setup

The dataset is collected [2] using a Renault Twizy vehicle (Figure 3.2) equipped with the following sensor configuration (as illustrated in Figure 3.3):

- **LiDAR**: Ouster OS1-128 LiDAR sensor with 128 channels vertical resolution, 865 nm laser wavelength, 100 m @ >90% detection probability and 120 m @ >50% detection probability (100 klx sunlight, 80% Lambertian reflectivity, 2048 @ 10 Hz rotation rate mode), 0.3 cm range resolution, 360° horizontal FOV and 45° (−22.5° to +22.5°) vertical FOV, mounted height ∼ 1.62 m.

- **Stereo Camera**: Carnegie Robotics MultiSense S21 stereo camera with grayscale, color, and IR enhanced imagers, 0.4 m minimum range, 2048 × 1088@ 2 MP resolution, up to 30 Hz frame rate and 115°× 68° FOV, 21 cm baseline, factory calibrated, mounted height ∼ 1.42 m.

- **GNSS/INS**: OxTS RT3000v3 global navigation satellite and inertial navigation system, with 0.03° pitch/roll accuracy, 0.1-1.5 m position accuracy, 0.15° slip angle

---

accuracy, 250 Hz maximum data output rate, supporting positioning from GPS, GLONASS, BeiDou, Galileo, PPP and SBAS constellations.

- **Lux Meter**: Yocto Light V3, a USB ambient light sensor (lux meter), measuring ambient light up to 100,000 lux, hence indirectly representing the conditions of the external environment via ambient illumination conditions.

## 3.3  Data Collection and Description

To ensure the dataset has diverse weather and varying density of pedestrian and traffic occurrences, we collect the data over a variety of conditions. These includes different types of environments, times of day, weather and repeated locations along the test route with data collected for the key time periods and environments shown in Table 3.1. As shown in Figures 3.4 and 3.5, our dataset mainly contains suburban, highway, city center and campus areas.

|  | Avg. Speed | Day. | Peak times | Night |
|---|---|---|---|---|
| City | 20.4 km/h | [3] \| [3] | [3] \| [3] | [2] \| [3] |
| Campus | 26.4 km/h | [1] \| [1] | [1] \| [2] | [1] \| [1] |
| Residential | 31.2 km/h | [1] \| [2] | [2] \| [2] | [1] \| [1] |
| Suburb | 43.6 km/h | [1] \| [1] | [1] \| [1] | [1] \| [1] |

**Table 3.1: Key time periods and environmental conditions**. The value is expressed in the form of [traffic density] | [population density], using a qualitative scale of [3 - high, 2 - normal, 1 - low].

All the data is provided in the *de facto* KITTI data formats, with the exception of the ambient light data (lux) which is not provided by KITTI and is hence published in a simple plain text format with aligned timestamp.

## 3.4  Ambient and Reflectivity Panoramic Imagery

The proposed DurLAR dataset is the first autonomous driving task dataset to additionally provide high-resolution ambient and reflectivity panoramic 360-degree imagery (refer to Section 2.2.2). As shown in Figure 2.2, the ambient imagery can be captured even in low light conditions (near infrared, 800-2500 nm), while the reflectivity imagery pertains to the material property of the scene object and its reflectivity of the 850 nm LiDAR signal in use (Ouster OS1-128). These characteristics, combined with a superior vertical resolution when compared to other datasets, enable these images to offer great benefit when dealing with unfavorable illumination conditions and coherent scene object identification.

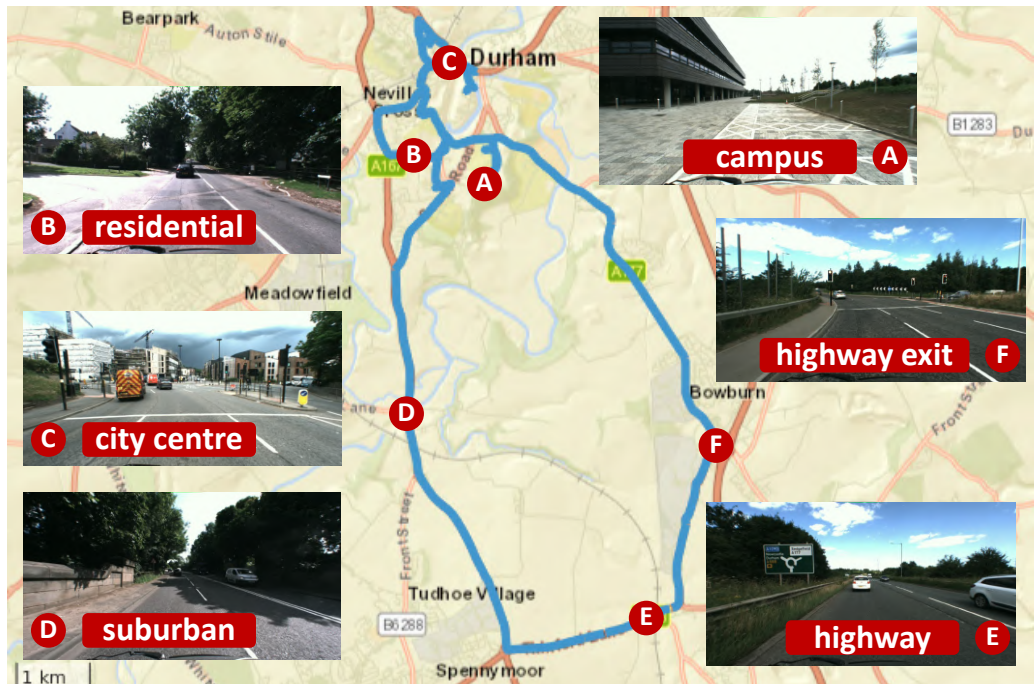

**Figure 3.4: The route** (blue curves) used for dataset collection showing a variety of driving environments.

**Ambient images** (Figure 2.2 (a)) offer day/night scene visibility in the near-infrared spectrum. The photon counting ASIC (Application Specific Integrated Circuit) of our sensor has particularly strong illumination sensitivity, so that the ambient images can be captured even in low light conditions. This is extremely practical in designing techniques that are specifically appropriate for adverse illumination conditions, such as nocturnal and adverse weather conditions.

**Reflectivity images** (Figure 2.2 (b)) contain information indicative of the material properties of the object itself and offer good consistency across illumination conditions and range. However, the Ouster OS1-128 LiDAR does not collect the true reflectivity data directly due to sensor limitations. Instead, an estimation of the reflectivity data is used to calculate the reflectivity images from the LiDAR intensity and range data. LiDAR intensity is the return signal strength of the laser pulse that recorded the range reading. According to the inverse square law (Equation (3.1)) for Lambertian objects in the far field, the intensity per unit area varies inversely proportional to the square of the distance [181],

$$I = \frac{S}{4\pi r^2},$$ (3.1)

where $I$ is the intensity, $r$ is the range (namely the distance of the object to the sensor) and $S$ is the source strength.

The calculation of reflectivity assumes that it is proportional to the source strength, which is also proportional to the product of intensity and the square of the range,

$$\text{Reflectivity} \propto S \propto Ir^2.$$ (3.2)

Exemplar ambient (near infrared) and reflectivity panoramic imagery is shown in Figure 3.6. In Figure 3.6 (a) and (c), clouds and shadows of objects can be distinguished (expressed as shades of grayscale). These pictures are very close to the images of grayscale or RGB camera. In Figure 3.6 (b) and (d), the reflectivity of the same object or material will remain constant regardless of the distance to the sensor, weather, light illumination and other conditions, since reflectivity is the intrinsic property of the object itself. The pillars of the building (Figure 3.6 (d)) have almost the same reflectivity (*i.e.* the same white color in the figure) regardless of their distance to the LiDAR sensor.

## 3.5 Calibration and Synchronization

LiDAR-to-camera calibration is performed using [182, 183]. With the custom calibration pattern shown in Figure 3.7, the calibration procedure is composed of two stages (following publication, we implement more precise and advanced calibrations – please refer to Appendix C). Firstly, a pair of two ArUco markers [184] are detected from the left frame of the stereo camera such that the transformation matrix $[R|t]$, containing rotation $R$ and translation $t$ parameters, between the camera and the center of the ArUco marker can be calculated (as shown in the overlays of Figure 3.8). Secondly, the edges of the orientated calibration boards are identified in the corresponding LiDAR data frame projection by orientated edge detection. Finally, the optimal rigid transformation between the LiDAR and the camera is found using RANSAC based optimization [182].

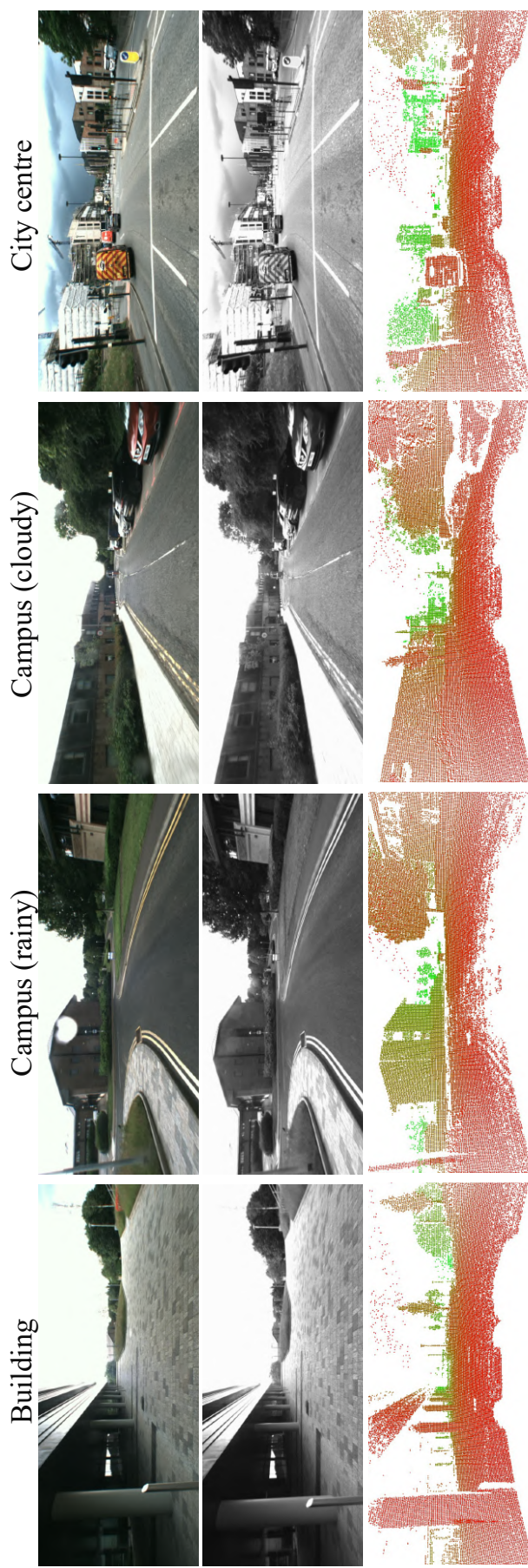

**Figure 3.5: Examples from DurLAR which demonstrate the diversity in our dataset.** From top to bottom, RGB left camera images (top), grayscale right camera images (center) and LiDAR point cloud (bottom). The point cloud is projected onto the 2D image plane using the LiDAR-to-left-camera external calibration, and the color varies with the distance from the LiDAR (near:=red → far:=green).

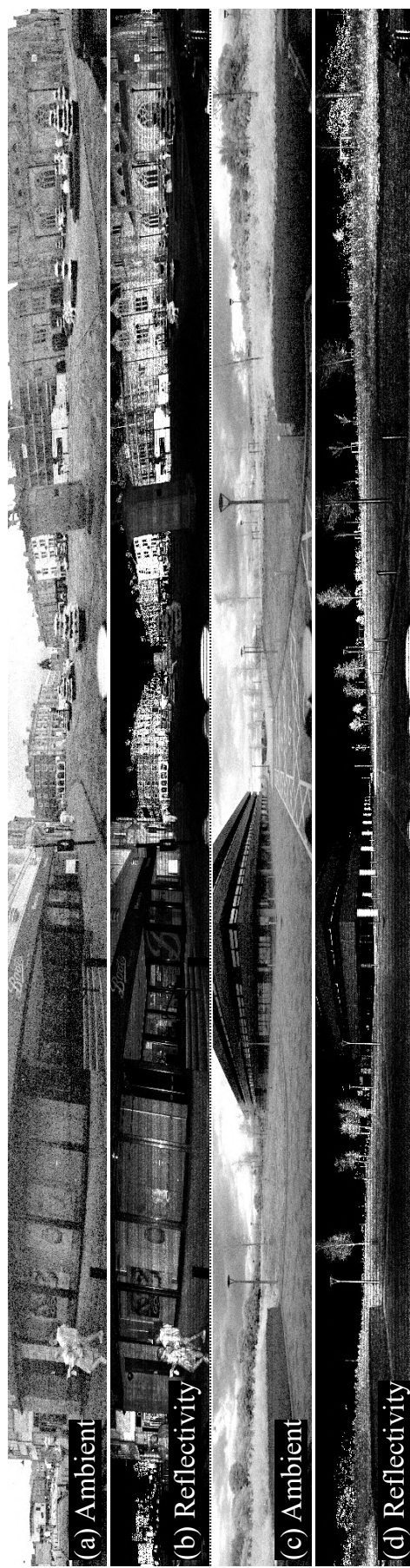

**Figure 3.6: Example of ambient (near infrared) and reflectivity panoramic images in real time, all without a camera.**

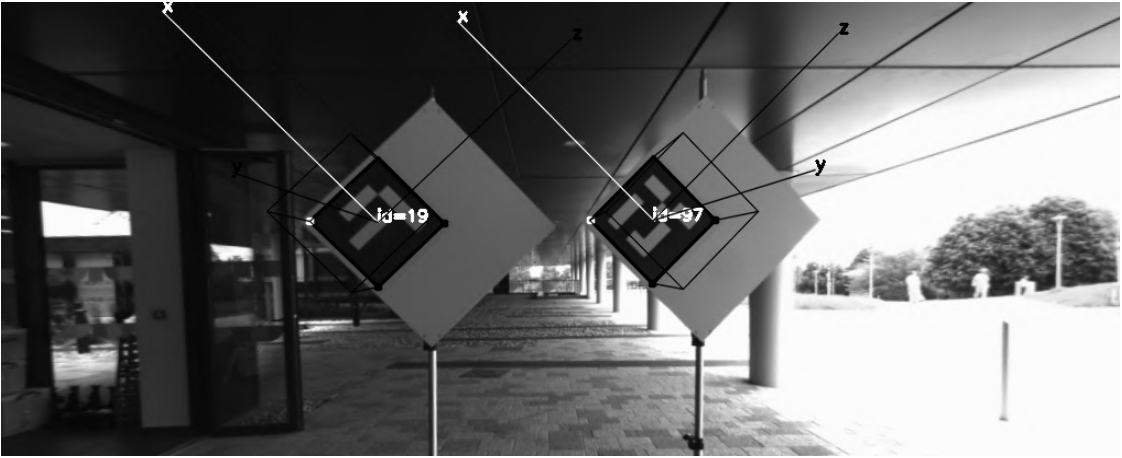

**Figure 3.7: Camera to LiDAR custom calibration pattern** with extrinsic parameter estimation overlay shown.

Stereo camera calibration is based on the manufacturer factory instructions for intrinsic and extrinsic settings. Calibration of the GNSS/INS is performed using the manufacturers recommended approach. The GNSS/INS with respect to the LiDAR is registered following [185].

All sensor synchronisation is performed at a rate of 10 Hz, using Robot Operating System (ROS, version Noetic) timestamps operating over a Gigabit Ethernet backbone to a common host (Intel Core i5-6300U, 16 GB RAM). More sophisticated methods like Precision Time Protocol (PTP) are available for applications needing higher synchronization accuracy.

The synchronization of sensors using ROS timestamps offers significant advantages, primarily due to its ease of implementation, cost-effectiveness, and flexibility. By leveraging the existing network infrastructure, ROS timestamps enable seamless and efficient synchronization across a wide range of sensors and devices without requiring additional hardware. This method simplifies the setup process and reduces costs, making it accessible for various applications. However, while ROS timestamps can be affected by network latency and jitter, introducing some variability in precision, the impact is generally manageable for many applications. In comparison, hardware synchronization provides higher precision and consistency by using dedicated components, making it ideal for applications requiring exact temporal accuracy, although it is more complex and costly to implement. We choose the appropriate synchronization strategy according to the trade-offs between these methods and our application requirements.

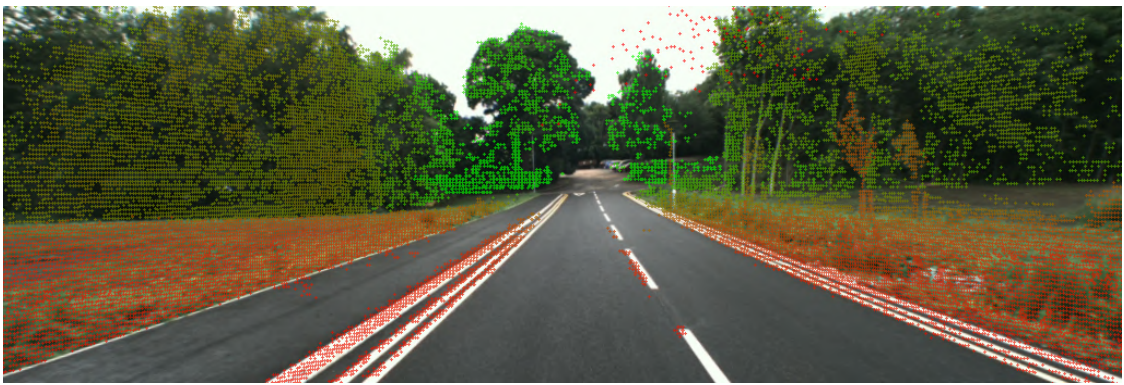

**Figure 3.8: Illustrative LiDAR 3D point cloud overlay** onto the right stereo image (color) using the calibration obtained.

## 3.6 Monocular Depth Estimation

Leveraging the higher vertical LiDAR resolution of our DurLAR dataset, we adopt monocular depth estimation as an illustrative benchmark task. We thus evaluate the relative performance of contemporary monocular depth estimation architectures [1, 85, 92], by leveraging the higher resolution LiDAR capability within DurLAR to facilitate more effective use of depth supervision, for which we propose a novel joint supervised/self-supervised loss formulation (Section 3.7). More broadly, the illumination-independent sensing capabilities of high-resolution 3D LiDAR additionally enable the evaluation of a range of driving tasks [60, 186] under varying environmental conditions spanning both extreme weather and illumination changes using our dataset.

We select ManyDepth [1] as a leading approach for monocular depth estimation as it offers state-of-the-art performance on the leading KITTI [15] and Cityscapes [9] benchmarks. Whilst ManyDepth [1] is a self-supervised approach, here we seek to leverage the availability of high-fidelity depth within DurLAR via the introduction of a secondary supervised loss term to formulate a novel supervised/self-supervised loss formulation. As a result, we can assess the impact of the availability of abundant ground truth depth at training time on the performance of this leading contemporary approach.

To these ends, we introduce the reverse Huber (Berhu) loss $\mathcal{L}_{\text{Berhu}}$ [187] as our supervised depth loss term, due to its effectiveness in smoothing and blurring depth prediction edges on object boundaries,

$$\mathscr{L}_{\text{Berhu}}\left(d, d^*\right) = \begin{cases} |d - d^*| & \text{if } |d - d^*| \leq \delta, \\ \frac{(d-d^*)^2 + \delta^2}{2\delta} & \text{if } |d - d^*| > \delta, \end{cases} \tag{3.3}$$

where $d$ is the predicted depth, $d^*$ is the ground truth depth, and $\delta$ stands for the threshold. If $|d - d^*| \leq \delta$, the Berhu loss is equal to $\mathscr{L}_1$; else, the Berhu loss acts approximately as $\mathscr{L}_2$.

We hence construct a joint supervised/semi-supervised version of ManyDepth [1], which adds $\mathscr{L}_{\text{Berhu}}$ into the original ManyDepth loss function, as shown in Equation (3.4):

$$\mathscr{L} = (1 - M)\mathscr{L}_p + \mathscr{L}_{\text{consistency}} + \mathscr{L}_{\text{smooth}} + \mathscr{L}_{\text{Berhu}}, \tag{3.4}$$

where $\mathscr{L}_p$ is the photometric reprojection error and $\mathscr{L}_{\text{smooth}}$ is the smoothness loss, from [1, 85]. $\mathscr{L}_{\text{consistency}}$ is the consistency loss, as implemented from [1].

For extended comparison, we similarly introduce this additional supervised depth loss via this additional Berhu loss term to the contemporary MonoDepth2 [85] and Depthhints [92] approaches leaving the remainder of the architectures unchanged. We specify a randomly generated data split for the DurLAR dataset as well, comprising 90k training frames, 5k validation frames and 5k test frames for our evaluation.

## 3.7 Evaluation Results

Training was performed with all learning parameters set as per the original works [1, 85, 92], with Berhu threshold $\delta = 0.2$, on a Nvidia Tesla V100 GPU over 20 epochs.

### 3.7.1 Quantitative Evaluation

The varying performance of self-supervised depth estimation between the KITTI [15], Cityscapes [9] and proposed DurLAR dataset illustrates the varying levels of challenge and complexity afforded by variations within the datasets (Table 3.2, records with × in the +S column)

However, within our evaluation on the DurLAR dataset, we consistently observe supe-rior performance (lower RMSE, higher accuracy, *etc,* Table 3.2) with the use of additional depth supervision (*i.e.* joint supervised/semi-supervised loss, Table 3.2, records with ✓ in the +S column) across all three monocular depth estimation approaches considered and show overall state-of-the-art performance on monocular depth estimation using our joint supervised/self-supervised ManyDepth variant (DurLAR, Table 3.2 - as highlighted in bold).

**Table 3.2: Performance comparison** over the KITTI Eigen split [8], Cityscapes [9] (self-supervised only) and DurLAR datasets (joint supervised/self-supervised, +S *v.s.* self-supervised). All models are trained and tested on the same dataset, without cross-dataset evaluation. Depth evaluation metrics (Section 2.8) are shown in the top row. Red refers to superior performances indicated by low values, and green refers to superior performance indicated by a higher value. The best results in KITTI and DurLAR are in **bold**; the second best in DurLAR are underlined.

| Dataset | Method | +S | W × H | Abs Rel | Sq Rel | RMSE | RMSE log | $\delta < 1.25$ | $\delta < 1.25^2$ | $\delta < 1.25^3$ |
|---|---|---|---|---|---|---|---|---|---|---|
| KITTI [15] | ManyDepth (MR) [1] | × | 640 × 192 | 0.098 | 0.770 | 4.459 | 0.176 | 0.900 | 0.965 | **0.983** |
| | ManyDepth (HR) [1] | × | 1024 × 320 | **0.093** | **0.715** | **4.245** | **0.172** | **0.909** | **0.966** | **0.983** |
| Cityscapes [9] | ManyDepth [1] | × | 416 × 128 | 0.114 | 1.193 | 6.223 | 0.170 | 0.875 | 0.967 | 0.989 |
| DurLAR | Depth-hints [92] | × | 640 × 192 | 0.122 | 1.070 | 4.148 | 0.211 | 0.870 | 0.946 | 0.972 |
| | Depth-hints [92] | ✓ | 640 × 192 | 0.121 | 1.109 | 4.121 | 0.210 | 0.874 | 0.946 | 0.972 |
| | MonoDepth2 [85] | × | 640 × 192 | 0.111 | 1.114 | 4.002 | 0.187 | 0.895 | 0.960 | 0.981 |
| | MonoDepth2 [85] | ✓ | 640 × 192 | <u>0.108</u> | <u>1.010</u> | <u>3.804</u> | 0.185 | 0.898 | 0.963 | 0.982 |
| | ManyDepth (MR) [1] | × | 640 × 192 | 0.115 | 1.227 | 4.116 | 0.186 | 0.892 | 0.962 | 0.982 |
| | ManyDepth (MR) [1] | ✓ | 640 × 192 | 0.109 | **0.936** | 3.711 | <u>0.176</u> | 0.895 | 0.964 | <u>0.984</u> |
| | ManyDepth (HR) [1] | × | 1024 × 320 | 0.109 | 1.111 | 3.875 | 0.177 | <u>0.901</u> | <u>0.966</u> | <u>0.984</u> |
| | ManyDepth (HR) [1] | ✓ | 1024 × 320 | **0.104** | 0.936 | **3.639** | **0.171** | **0.906** | **0.969** | **0.986** |

### 3.7.2 Qualitative Evaluation

To qualitatively illustrate the difference between self-supervised and joint supervised/self-supervised ManyDepth with the addition of depth loss, we show exemplar results in Figure 3.9 with areas of superior depth estimation indicated (green).

Within these examples, we can see a clearer contour edge of the bus and resolution of the upper LED display board on the vehicle (Figure 3.9, top - self-supervised *v.s.* supervised/self-supervised). Furthermore, we see improved depth resolution of the building (Figure 3.9, middle - self-supervised *v.s.* supervised/self-supervised) whereby additional depth supervision enables the technique to correctly estimate the depth of the supporting building pillars and is even able to resolve the depth of the short stainless steel stub in the foreground. Finally, we can see improved estimation and clarity of both vehicle and pedestrian depth within a crowded urban scene (Figure 3.9, bottom - self-supervised *v.s.* supervised/self-supervised).

**Table 3.3: Ablation results on ManyDepth [1].** vRes := the vertical resolution of LiDAR ground truth depth. $\pm$S := supervised/self-supervised (+S) and self-supervised ManyDepth (-S) for consistency with Table 3.2. $\delta_1$, $\delta_2$ and $\delta_3$ refers to $\delta < 1.25$, $\delta < 1.2^2$ and $\delta < 1.25^3$ respectively.

| vRes | Abs Rel | Sq Rel | RMSE | RMSE log | $\delta_1$ | $\delta_2$ | $\delta_3$ |
|---|---|---|---|---|---|---|---|
| 32/+S | 0.115 | **0.908** | <u>3.677</u> | 0.179 | 0.888 | 0.966 | <u>0.985</u> |
| 64/+S | <u>0.107</u> | <u>0.918</u> | 3.735 | <u>0.175</u> | 0.895 | <u>0.967</u> | **0.986** |
| 128/-S | 0.109 | 1.111 | 3.875 | 0.177 | <u>0.901</u> | 0.966 | 0.984 |
| 128/+S | **0.104** | 0.936 | **3.639** | **0.171** | **0.906** | **0.969** | **0.986** |

### 3.7.3 Ablation Study

Our ablation study shows the side-by-side impact of our joint supervised/unsupervised loss formulation in addition to the performance impact of high-fidelity depth (higher vertical LiDAR resolution).

**Supervised depth**: We train the ManyDepth [1] with and without the Berhu loss (Equation 3.3), such that we can compare the original self-supervised performance with that of additional depth supervision (Table 3.3, 128/-S *v.s.* 128/+S).

**Ground truth depth resolution:** We simulate a reduction in vertical ground truth depth resolution by subsampling the depth values present by 50% (64 channels) and 75% (32 channels) along the vertical axis of the LiDAR ground truth projection (Table 3.3).

From Table 3.3, we can see the superior performance of our joint supervised/unsupervised loss formulation (128/-S *v.s.* 128/+S). From Table 3.3 (DurLAR), we can see the superior performance from higher vertical LiDAR resolution (32/64 *v.s.* 128/-S).

## 3.8 Summary

We present a high-fidelity 128-channel 3D LiDAR dataset with panoramic ambient (near infrared) and reflectivity imagery for autonomous driving applications (DurLAR). In addition, we present the exemplar benchmark task of depth estimation task whereby we show the impact of higher resolution LiDAR as a means to the supervised extension of leading contemporary monocular depth estimation approaches [1, 85, 92].

DurLAR, is a novel large-scale dataset comprising contemporary high-fidelity LiDAR, stereo/ambient/reflectivity imagery, GNSS/INS and environmental illumination information under repeated route, variable environment conditions (in the *de facto* KITTI dataset format). It is the first autonomous driving task dataset to additionally comprise usable ambiance and reflectivity LiDAR obtained imagery ($2048 \times 128$ resolution).

In our sample monocular depth estimation task, we show superior performance can be achieved by leveraging the high resolution LiDAR resolution afforded by DurLAR via the secondary introduction of an additional supervised loss term for depth. This is demonstrated across three state-of-the-art monocular depth estimation approaches [1, 85, 92]. We show that the recent availability of abundant high-resolution ground truth depth from sensors such as those used in DurLAR enable new research possibilities for supervised learning within this domain.

Further work will consider the provision of additional dataset annotation for extra tasks, semantic and geometric scene information, and the ambient together with reflectivity imagery will be further explored. The high-resolution point clouds collected in this chapter present additional challenges for deep learning training and annotation due to their significant data volume. Therefore, in the next chapter (Chapter 4), we will explore novel methods to address the challenges posed by the large-scale 3D point clouds and their associated data.

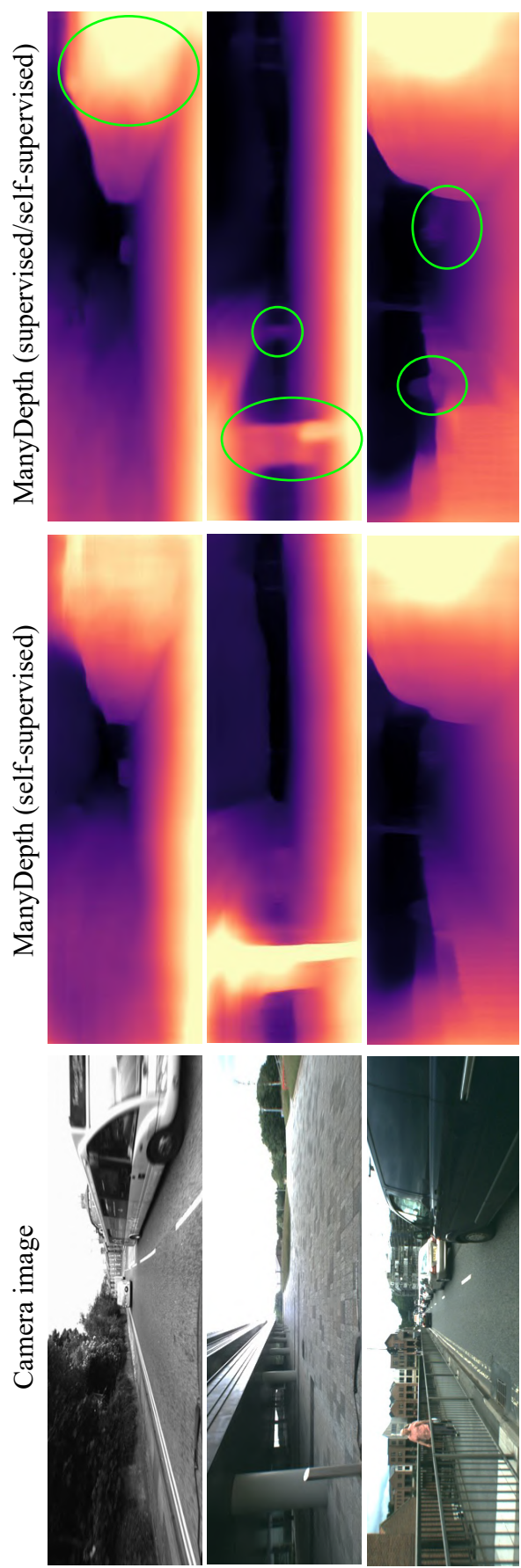

**Figure 3.9: Comparison of monocular depth estimation results** with areas of improvement highlighted with the use of depth supervision (green)

---

# Efficient 3D LiDAR Semantic Segmentation

---

Portions of this chapter have previously been published in the following peer-reviewed publication [16]:

- **Li, L.**, Shum, H. P., & Breckon, T. P., "Less is More: Reducing Task and Model Complexity for 3D Point Cloud Semantic Segmentation." In Conference on Computer Vision and Pattern Recognition (**CVPR**). IEEE, 2023.

While the availability of 3D LiDAR point cloud data has significantly increased in recent years, annotation remains expensive and time-consuming due to its unstructured nature and non-topological characteristics. The unstructured nature of LiDAR data refers to the random and scattered collection of points in space, which do not follow a regular grid or consistent pattern. Unlike structured data, such as images with fixed resolutions and orderly pixel arrangements, LiDAR point clouds lack inherent structure, making them challenging to process with conventional techniques. These have led to a demand for efficient semi-supervised semantic segmentation methods to support application domains such as autonomous driving. Existing work very often employs relatively large segmentation backbone networks to improve segmentation accuracy, at the expense of computational costs. In addition, many use uniformly sampled frames from videos to reduce ground truth data requirements for learning needed, often resulting in sub-optimal performance. To address these issues, we propose a new pipeline that employs a smaller architecture, requiring fewer ground-truth annotations to achieve superior segmentation accuracy compared to contemporary approaches. This is facilitated via a novel Sparse Depthwise Separable Convolution (SDSC) module that significantly reduces the network parameter count while retaining overall task performance. To effectively sub-sample our training data, we propose a new Spatio-Temporal Redundant Frame Downsampling (ST-RFD) method that leverages knowledge of sensor motion within the environment to extract a more diverse subset of training data frame samples. To leverage the use of limited annotated data samples, we further propose a soft pseudo-label method informed by LiDAR reflectivity. Our method outperforms contemporary semi-supervised work in terms of mIoU, using less labeled data, on the SemanticKITTI (59.5@5%) and ScribbleKITTI (58.1@5%) benchmark datasets, based on a 2.3× reduction in model parameters and 641× fewer multiply-add operations whilst also demonstrating significant performance improvement on limited training data (*i.e.*, *Less is More*, as per the old English proverb that implies a smaller quantity could lead to higher quality).

## 4.1 Introduction

Many contemporary methods on 3D semantic segmentation require relatively large backbone architectures with millions of trainable parameters requiring many hundred gigabytes of annotated data for training at a significant computational cost. Considering the time-consuming and costly nature of 3D LiDAR annotation, such methods have become less feasible for practical deployment.

Existing supervised 3D semantic segmentation methods [20,42–44,48,149,151,152,154] primarily focus on designing network architectures for densely annotated data. To reduce the need for large-scale data annotation, and inspired by similar work in 2D [188–190], recent 3D work proposes efficient ways to learn from weak supervision [4]. However, such methods still suffer from high training costs and inferior on-task performance. To reduce computational costs, a 2D projection-based point cloud representation is often considered [42–45, 151, 152, 191, 192], but again at the expense of significantly reduced on-task performance. As such, we observe a gap in the research literature for the design of semi or weakly supervised methodologies that employ a smaller-scale architectural backbone, hence facilitating improved training efficiency whilst also reducing their associated data annotation requirements.

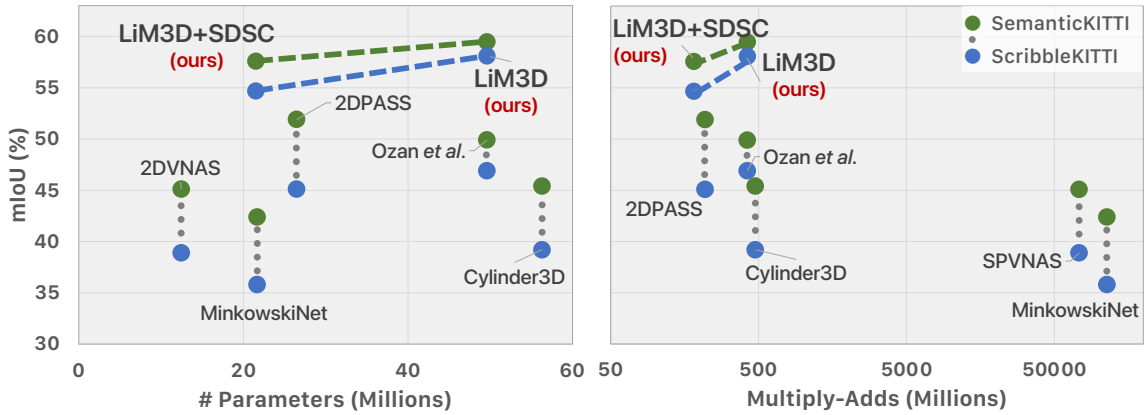

**Figure 4.1: mIoU performance (%) against parameters and multiply-add operations** on SemanticKITTI (fully annotated) and ScribbleKITTI (weakly annotated) under the 5% sampling protocol.

In this chapter, we propose a semi-supervised methodology for 3D LiDAR point cloud semantic segmentation. Facilitated by three novel design aspects, our *Less is More* (LiM) based methodologies require *less* training data and *less* training computation whilst offering (*more*) improved accuracy over contemporary state-of-the-art approaches (see Figure 4.1).

Firstly, from an architectural perspective, we propose a novel **Sparse Depthwise Separable Convolution (SDSC)** module, which substitutes traditional sparse 3D convolution into existing 3D semantic segmentation architectures, resulting in a significant reduction in trainable parameters and numerical computation whilst maintaining on-task performance (see Figure 4.1). Depthwise Separable Convolution has shown to be very effective within image classification tasks [146]. Here, we tailor a sparse variant of 3D Depthwise Separable Convolution for 3D sparse data by first applying a single submanifold sparse convolutional filter [75, 193] to each input channel with a subsequent pointwise convolution to create a linear combination of the sparse depthwise convolution outputs. This work is the first to introduce depthwise convolution into the 3D point cloud segmentation field as a conduit to reduce model size. Our SDSC module facilitates a 50% reduction in trainable network parameters without any loss in segmentation performance.

Secondly, from a training data perspective, we propose a novel **Spatio-Temporal Redundant Frame Downsampling (ST-RFD)** strategy that more effectively sub-samples a set of diverse frames from a continuously captured LiDAR sequence in order to maximize diversity within a minimal training set size. We observe that continuously captured LiDAR sequences often contain significant temporal redundancy, similar to that found in video [194], whereby temporally adjacent frames provide poor data variation. On this basis, we propose to compute the temporal correlation between adjacent frame pairs, and use this to select the most informative subset of LiDAR frames from a given sequence. Unlike passive sampling (*e.g.*, uniform or random sampling), our active sampling approach samples frames from each sequence such that redundancy is minimized and hence training set diversity is maximal. When compared to commonplace passive random sampling approaches [4, 18, 155], ST-RFD explicitly focuses on extracting a diverse set of training frames that will hence maximize model generalization.

Finally, in order to employ semi-supervised learning, we propose a soft pseudo-label method informed by the LiDAR reflectivity response, thus maximizing the use of any annotated data samples. Whilst directly using unreliable soft pseudo-labels generally results in performance deterioration [195], the voxels corresponding to the unreliable predictions can instead be effectively leveraged as negative samples of unlikely categories. Therefore, we use cross-entropy to separate all voxels into two groups, *i.e.*, a reliable and an unreliable group with low and high-entropy voxels respectively. We utilize predictions from the reliable group to derive positive pseudo-labels, while the remaining voxels from the unreliable group are pushed into a FIFO category-wise memory bank of negative samples [196]. To further assist semantic segmentation of varying materials in the situation where we have weak/unreliable/no labels, we append the reflectivity response features onto the point cloud features, which again improve segmentation results.

We evaluate our method on the SemanticKITTI [3] and ScribbleKITTI [4] *validation* set. Our method outperforms contemporary state-of-the-art semi- [18, 155] and weakly- [4] supervised methods and offers *more* in terms of performance on limited training data, whilst using *less* trainable parameters and *less* numerical operations (*Less is More*).

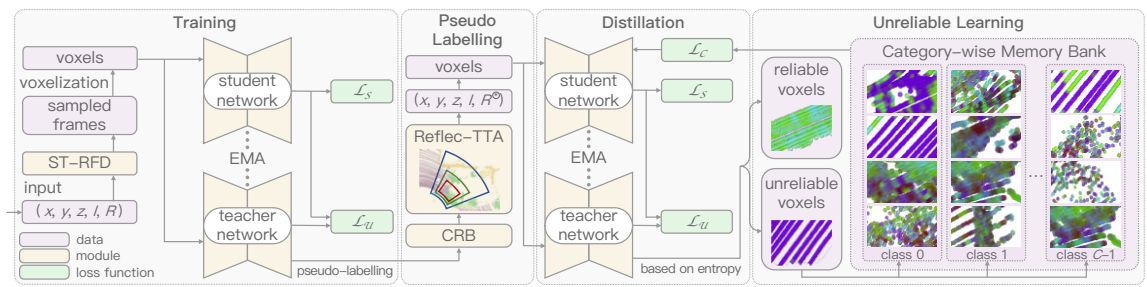

**Figure 4.2: Our proposed architecture** for unreliable pseudo-labels LiDAR semantic segmentation involves three stages: training, pseudo-labeling, and distillation with unreliable learning. We apply ST-RFD sampling before training the Mean Teacher on available annotations.

Overall, our contributions can be summarized as follows:

- A novel methodology for semi-supervised 3D LiDAR semantic segmentation that uses significantly *less* parameters and offers (*more*) superior accuracy.

- A novel Sparse Depthwise Separable Convolution (SDSC) module, to reduce trainable network parameters, and to both reduce the likelihood of over-fitting and facilitate a deeper network architecture.

- A novel Spatio-Temporal Redundant Frame Downsampling (ST-RFD) strategy, to extract a maximally diverse data subset for training by removing temporal redundancy and hence future annotation requirements.

- A novel soft pseudo-labeling method informed by LiDAR reflectivity as a proxy to in-scene object material properties, facilitating effective use of limited data annotation.

## 4.2 Overview

We first present an overview of the mean teacher framework we employ (Section 4.3) and then explain our use of unreliable pseudo-labels informed by LiDAR reflectivity for semi-supervised learning (Section 4.4). Subsequently, we detail our ST-RFD strategy for dataset diversity (Section 4.5) and finally our parameter-reducing SDSC module (Section 4.6).

Formally, given a LiDAR point cloud $P = \{\mathbf{p} \mid \mathbf{p} = (x, y, z, I, R) \in \mathbb{R}^5\}$ where $(x, y, z)$ is a 3D coordinate, $I$ is intensity and $R$ is reflectivity, our goal is to train a semantic segmentation model by leveraging both a large amount of unlabeled $U = \{\mathbf{p}_i^u\}_{i=1}^{N_u} \subsetneq P$ and a smaller set of labeled data $V = \{(\mathbf{p}_i^v, \mathbf{y}_i^v)\}_{i=1}^{N_v} \subsetneq P$.

Our overall architecture involves three stages (Figure 4.2): (1) **Training:** we utilize reflectivity-prior descriptors and adapt the Mean Teacher framework to generate high-quality pseudo-labels; (2) **Pseudo-labeling:** we fix the trained teacher model prediction in a class-range-balanced [4] manner, expanding dataset with Reflectivity-based Test Time Augmentation (Reflec-TTA) during test time; (3) **Distillation with unreliable predictions:** we train on the generated pseudo-labels, and utilize unreliable pseudo-labels in a category-wise memory bank for improved discrimination.

## 4.3 Mean Teacher Framework

We introduce weak supervision using the Mean Teacher framework [157], which avoids the prominent slow training issues associated with Temporal Ensembling [197]. This framework consists of two models of the same architecture known as the student and teacher respectively, for which we utilize a Cylinder3D-based [20] segmentation head $f$. The weights of the student model $\theta$ are updated via standard backpropagation, while the weights of the teacher model $\theta^*$ are updated by the student model through Exponential Moving Averaging:

$$\theta_{t+1}^* = \kappa \theta_t^* + (1 - \kappa)\theta_{t+1}, \quad t \in \{0, 1, \cdots T - 1\}, \tag{4.1}$$

where $\kappa$ denotes a smoothing coefficient to determine update speed, and $T$ is the maximum time step.

During training, we train a set of weakly-labeled point cloud frames with voxel-wise inputs generated via asymmetrical 3D convolution networks [20]. For every point cloud, our optimization target is to minimize the overall loss:

$$\mathcal{L} = \mathcal{L}_S + \lambda_U \mathcal{L}_U + g\lambda_C \mathcal{L}_C, \tag{4.2}$$

where $\mathcal{L}_S$ and $\mathcal{L}_U$ denote the losses applied to the supervised and unsupervised set of points respectively, $\mathcal{L}_C$ denotes the contrastive loss to make full use of unreliable pseudo-labels, $\lambda_U$ is the weight coefficient of $\mathcal{L}_U$ to balance the losses, and $g$ is the gated coefficient of $\mathcal{L}_C$. $g$ equals 1 if and only if it is in the distillation stage. We use the consistency loss, lovasz softmax loss [198], and the voxel-level InfoNCE [189] as $\mathcal{L}_U$, $\mathcal{L}_S$ and $\mathcal{L}_C$ respectively.

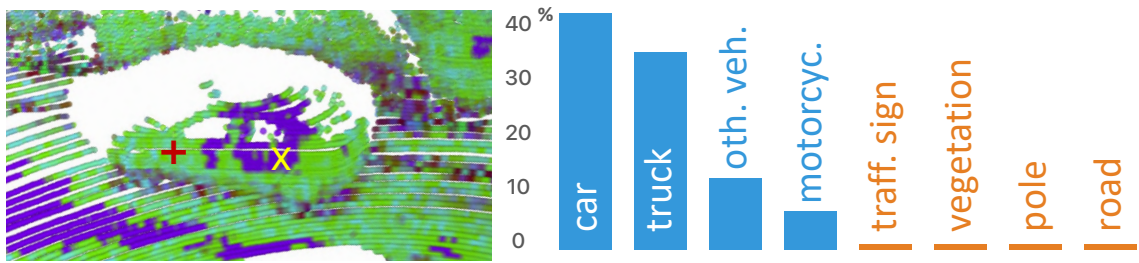

**Figure 4.3: Illustration on unreliable pseudo-labels**. Left: entropy predicted from an unlabeled point cloud, with lower entropy corresponding to greener color. Right: Category-wise probability of an unreliable prediction X, only top-4 and last-4 probabilities shown.

We first generate our pseudo-labels for the unlabeled points via the teacher model. Subsequently, we generate reliable pseudo-labels in a Class-Range-Balanced (CRB) [4] manner, and utilize the qualified unreliable pseudo-labels as negative samples in the distillation stage. Finally, we train the model with both reliable and qualified unreliable pseudo-labels to maximize the quality of the pseudo-labels.

## 4.4 Learning from Unreliable Pseudo-Labels

Unreliable pseudo-labels are frequently eliminated from semi-supervised learning tasks or have their weights decreased to minimize performance loss [4, 155, 199–202]. In line with this idea, we utilize CRB method [4] to first mask off unreliable pseudo-labels and then subsequently generate high-quality reliable pseudo-labels.

However, such a simplistic discarding of unreliable pseudo-labels may lead to valuable information loss as it is clear that unreliable pseudo-labels (*i.e.*, the corresponding voxels with high entropy) can offer information that helps in discrimination. Voxels that correlate to unreliable predictions can alternatively be thought as negative samples for improbable categories [190], although performance would suffer if such unreliable predictions are used as pseudo-labels directly [195]. As shown in Figure 4.3, the unreliable pseudo predictions have no confidence on `car` and `truck` classes, whilst being sure that that voxel cannot be `pole` or `road`. Subsequently, together with the use of CRB for high-quality reliable pseudo-labels, we also ideally want to make full use of these remaining unreliable pseudo-labels rather than simply discarding them. Following Wang *et al.* [190], we propose a method to leverage such unreliable pseudo-labels for 3D voxels as negative samples. However, to maintain a stable amount of negative samples, we utilize a category-wise memory bank $\mathcal{Q}_c$ (FIFO queue, [203]) to store all the negative samples for a given class $c$. As negative candidates in some specific categories are severely limited in a mini-batch due to the long-tailed class distribution of many tasks (*e.g.* autonomous driving), without such an approach in place we may instead see the gradual dominance of large and simple-to-learn classes within our generated pseudo-labels.

Following [189, 204], our method has three prerequisites, *i.e.*, anchor voxels, positive candidates, and negative candidates. They are obtained by sampling from a particular subset, constructed via Equation (4.3) and Equation (4.4), in order to reduce overall computation. In particular, the set of features of all candidate anchor voxels for class $c$ is denoted as:

$$\mathcal{A}_c = \left\{ \mathbf{E}_{a,b} \mid y_{a,b}^* = c, p_{a,b}(c) > \delta_p \right\},\tag{4.3}$$

where $\mathbf{E}_{a,b}$ is the feature embedding for the $a$-th point cloud frame at voxel $b$, $\delta_p$ is the positive threshold of all classes, $p_{a,b}(c)$ is the softmax probability by the segmentation head at $c$-th dimension. $y_{a,b}^*$ is set to the ground truth label $y_{a,b}^*$ if the ground truth is available, otherwise, $y_{a,b}^*$ is set to the pseudo label $\hat{y}_{a,b}$, due to the absence of ground truth.

The positive sample is the common embedding center of all possible anchors, which is the same for all anchors from the same category, shown in Equation (4.4).

$$\mathbf{E}_{\mathbf{c}}^+ = \frac{1}{|\mathcal{A}_c|} \sum_{\mathbf{E}_c \in \mathcal{A}_c} \mathbf{E}_c.\tag{4.4}$$

Following [190], we similarly construct multiple negative samples $\mathbf{E}_c^-$ for each anchor voxel.

Finally, for each anchor voxel containing one positive sample and $N - 1$ negative samples, we propose the voxel-level InfoNCE loss [189] (a variant of contrastive loss) $\mathcal{L}_C$ in Equation (4.5) to encourage maximal similarity between the anchor voxel and the positive sample, and the minimal similarity between the anchor voxel and multiple negative samples.

$$\mathcal{L}_C = -\frac{1}{C} \sum_{c=0}^{C-1} \mathop{\mathbb{E}}_{\mathbf{E}_c} \left[ \log \frac{f\left(\mathbf{e}_c, \mathbf{e}_c^+, \tau\right)}{\sum_{\mathbf{e}_{c,j}^- \in \mathbf{E}_c^-} f\left(\mathbf{e}_c, \mathbf{e}_{c,j}^-, \tau\right)} \right]$$
$$= -\frac{1}{C} \sum_{c=0}^{C-1} \mathop{\mathbb{E}}_{\mathbf{E}_c} \left[ \log \frac{\exp\left(\langle \mathbf{e}_c, \mathbf{e}_c^+ \rangle / \tau\right)}{\exp\left(\langle \mathbf{e}_c, \mathbf{e}_c^+ \rangle / \tau\right) + \sum\limits_{j=1}^{N-1} \exp\left(\langle \mathbf{e}_c, \mathbf{e}_{c,j}^- \rangle / \tau\right)} \right].$$
$(4.5)$

where $\mathbf{e}_{c,j}^-$ denotes the embedding of the $j$-th negative sample of class $c$, and $\langle \cdot, \cdot \rangle$ denotes cosine similarity.

To obtain minimal accuracy degradation despite very few weak labels, *e.g.*, 1% weakly-labeled ScribbleKITTI [4] dataset, we propose a Test-Time Augmentation (TTA) that does not depend on any label, but only relies on a feature of the original LiDAR points themselves. Also included in almost every LiDAR benchmark dataset for autonomous driving [3–5, 8, 17, 61], is the intensity of light reflected from the surface of an object at each point. In the presence of limited data labels in the semi-supervised learning case, this property of the material surface, normalized by distance to obtain surface reflectivity in Equation (4.6), could readily act as auxiliary information to identify different semantic classes.

Our intuition is that reflectivity $R$, as a point-wise distance-normalized intensity feature, offers consistency across lighting conditions and ranges as:

$$R = Ir^2 = \frac{S}{4\pi r^2} \cdot r^2 \propto S, \tag{4.6}$$

where $S$ is the return strength of the LiDAR laser pulse, $I$ is the intensity and $r$ is the point distance from the source on the basis that scene objects with similar surface material, coating, and color characteristics will share similar $S$ returns.

On this basis, we define our novel reflectivity-based Test-Time Augmentation (Reflec-TTA) technique, as a substitute for label-dependent Pyramid Local Semantic-context (PLS) augmentation [4] during test-time as ground truth is not available. We append our point-wise reflectivity to the existing point features in order to enhance performance in the presence of false or non-existent pseudo-labels at the distillation stage. As shown in Figure 4.4, and following [4], we apply various sizes $s$ of bins in cylindrical coordinates to analyze the intrinsic point distribution of the LiDAR sensor at varying resolutions (shown in red, green and blue in Figure 4.4). For each bin $b_i$, we compute a coarse histogram, $\mathbf{h}_i$:

$$
\begin{aligned}
\mathbf{h}_i &= \left\{ h_i^{(k)} \mid k \in [1, N_b] \right\} \in \mathbb{R}^{N_b}, \quad i \in [1, s], \\
h_i^{(k)} &= \# \left\{ R_j \in r_k, \ \forall j \mid p_j \in b_i \right\}, \\
r_k &= \left[ \ (k-1)/N_b, \ k/N_b \ \right), \quad k \in \left[ \ 1, N_b \ \right].
\end{aligned}
\tag{4.7}
$$

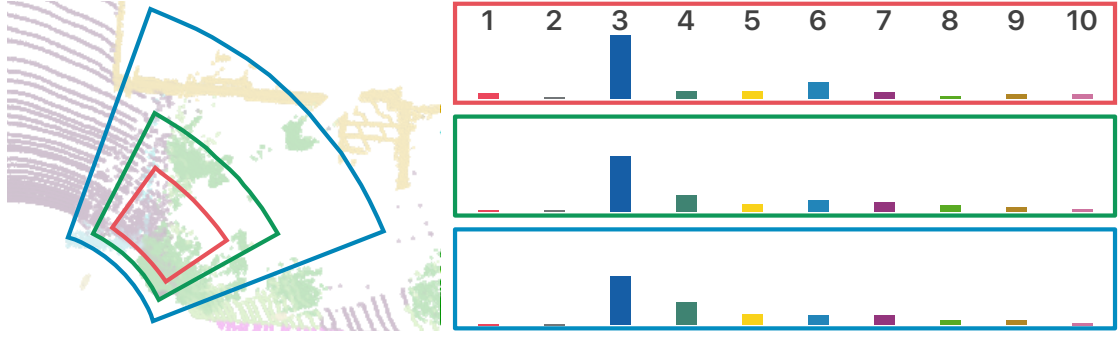

**Figure 4.4: Coarse histograms** of Reflec-TTA bins (not to scale).

The Reflec-TTA features $R^\circledR$ of all points $p_j \in b_j$ is further computed as the concatenation of the coarse histogram $\mathbf{h}_i$ of the normalized histogram:

$$
R^\circledR = \left\{ \mathbf{h}_i / \max \left( \mathbf{h}_i \right) \mid i \in [1, s] \right\} \in \mathbb{R}^{s N_b}
\tag{4.8}
$$

In the distillation stage, we append $R^\circledR$ to the input features and redefine the input LiDAR point cloud as the augmented set of points $P^\circledR = \left\{ p \mid (x, y, z, I, R^\circledR) \in \mathbb{R}^{s N_b + 4} \right\}$.

## 4.5 Spatio-Temporal Redundant Frame Downsampling

Due to the spatio-temporal correlation of LiDAR point cloud sequences often captured from vehicles in metropolitan locales, many large-scale point cloud datasets demonstrate significant redundancy. Common datasets employ a frame rate of 10Hz [3,5,8,17,37,61,72], and a number of concurrent laser channels of 32 [5], 64 [3,8,37,61,72] or 128 [17]. Faced with such large-scale, massively redundant training datasets, the popular practice of semi-supervised semantic segmentation approaches [18,157,188,205,206] is to uniformly sample 1%, 10%, 20%, or 50% of the available annotated training frames, without considering any redundancy attributable to temporary periods of stationary capture (*e.g.* due to traffic, Figure 4.5) or multi-pass repetition (*e.g.* due to loop closure).

To extract a diverse set of frames, we propose a novel algorithm called Spatio-Temporal Redundant Frame Downsampling (ST-RFD, Algorithm 1) that determines spatio-temporal redundancy by analyzing the spatial-overlap within time-continuous LiDAR frame sequences. The key idea is that if spatial-overlap among some continuous frame sequences is high due to spatio-temporal redundancy, multiple representative frames can be sub-sampled for training, significantly reducing both training dataset size and redundant training computation.

Figure 4.6 shows an overview of ST-RFD. It is conducted inside each temporal continuous LiDAR sequence $e$. First, we evenly divide $p$ point cloud frames in each sequence into $\lceil p/q \rceil$ subsets (containing $q$ frames). For each frame at time $t$ inside the subset, we find its corresponding RGB camera image in the dataset at time $t$ and $t + 1$. To detect the spatio-temporal redundancy at time $t$, the similarity $\psi(t, t + 1)$ between temporally adjacent frames are then computed via the Structural Similarity Index Measure (SSIM, [207]). We utilize the mean value of similarity scores between all adjacent frames in the current subset as a proxy to estimate the spatio-temporal redundancy present. A sampling rate is then determined according to this mean similarity for frame selection within this subset. This is repeated for all subsets in every sequence to construct our final set of sub-sampled LiDAR frames for training.

Concretely, as shown in Algorithm 1, we implement a ST-RFD supervisor that determines the most informative assignments (*i.e.*, the key point cloud frames) that the teacher and student networks should train on respectively. The ST-RFD supervisor has an empirical supervisor function $\upsilon$, which decides the number of assignments, *i.e.*, the sampling rate corresponding to the extent of spatio-temporal redundancy. Using SSIM [207] as the redundancy function $\psi$ to measure the similarity between the RGB images associated with two adjacent point clouds, we define the empirical supervisor function $\upsilon$ with decay property $\upsilon(x) = \exp(-\beta x)$, where $\beta \in (0, +\infty)$ is the decay coefficient, and $x$ is the redundancy calculated from $\psi$. In this way, the higher the degree of spatio-temporal redundancy (as $\psi \to 1$), the lower the sampling rate our ST-RFD supervisor will allocate,

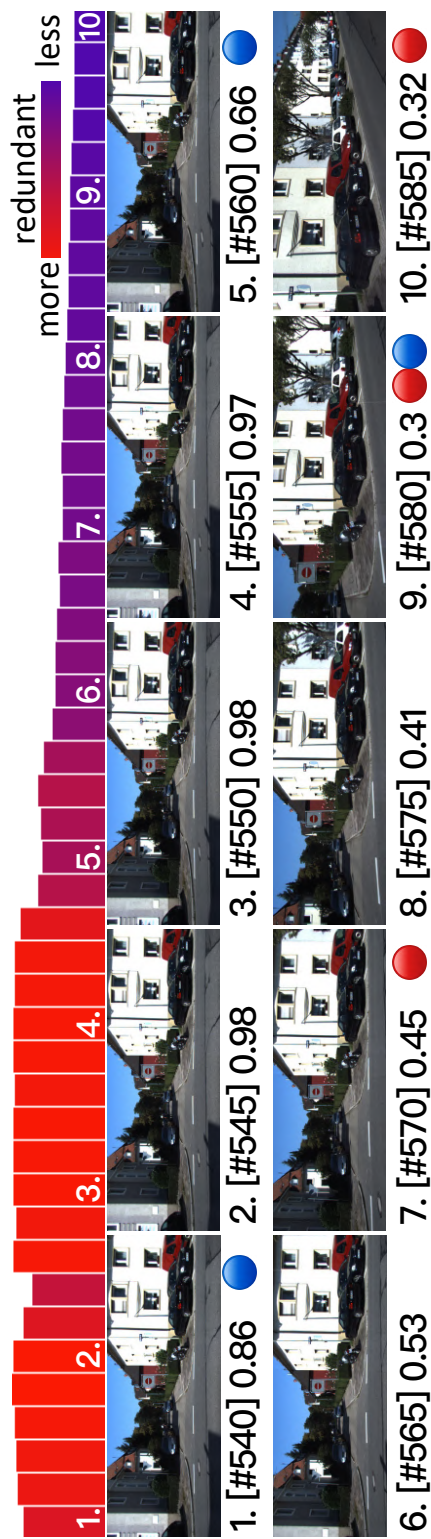

**Figure 4.5: Illustration of LiDAR frame temporal correlation as [# frame ID] redundancy** with 5% sampling on SemanticKITTI [3] (sequence 00) using uniform sampling (selected frames in ●) and ST-RFD strategy (●).

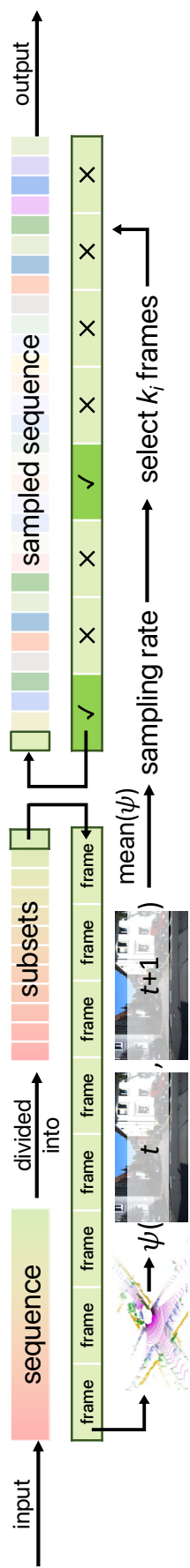

**Figure 4.6: Overview** of our proposed Spatio-Temporal Redundant Frame Downsampling approach.

---

**Algorithm 1:** Spatio-Temporal Redundant Frame Downsampling.

**Input:** Point cloud frames pool $P$ (size of $p$), subset size $q$, redundancy function $\psi \in [0, 1]$ and empirical supervisor function $\upsilon$.

**1** Divide $P$ evenly into $\lceil p/q \rceil$ subsets $Q$.

**2** $D \leftarrow$ empty dictionary.

**3 forall** $e \leftarrow 0 : n_e - 1$ **do**
    // loop for all sequences

**4**     **forall** $i \leftarrow 0 : \lceil p/q \rceil - 1$ **do**
        // loop for subsets $Q$

**5**         $C_i \leftarrow \varnothing$       // chosen point cloud frames

**6**         $Q_{i,j} \leftarrow j$-th frame in subset $Q_i$.

**7**         $\overline{M_i} \leftarrow \frac{1}{s} \sum_{j=0}^{s-1} \psi(Q_{i,j})$.     // redundancy

**8**         $k_i \leftarrow \lceil \upsilon(\overline{M_i}) \cdot s \rceil$.

**9**         $T_i \leftarrow$ select $k_i$ frames in $Q_i$ with the smallest $M_i$.

**10**         $C_i \leftarrow C_i \cup T_i$.

**11**     Append key-value pair $(e, C_i)$ into $D$.

**Return:** Dictionary $D$.

---

hence reducing the training set requirements for teacher and student alike.

## 4.6   Sparse Depthwise Separable Convolution

Existing LiDAR point cloud semantic segmentation methods generally rely on a large-scale backbone architecture with tens of millions of trainable parameters [4, 20, 36, 38–40] due to the requirement for 3D (voxel-based) convolution operations, to operate on the voxelized topology of the otherwise unstructured LiDAR point cloud representation, which suffer from both high computational training demands and the risk of overfitting. Based on the observation that Depthwise Separable Convolution has shown results comparable with regular convolution in tasks such as image classification but with significantly fewer trainable parameters [142–146, 208], here we pursue the use of such an approach within 3D point cloud semantic segmentation.

As such we propose the first formulation of sparse variant Depthwise Separable Convolution [142] applied to 3D point clouds, namely Sparse Depthwise Separable Convolution (SDSC). SDSC combines the established computational advantages of sparse convolution for point cloud segmentation [75], with the significant trainable parameter reduction offered by Depthwise Separable Convolution [146].

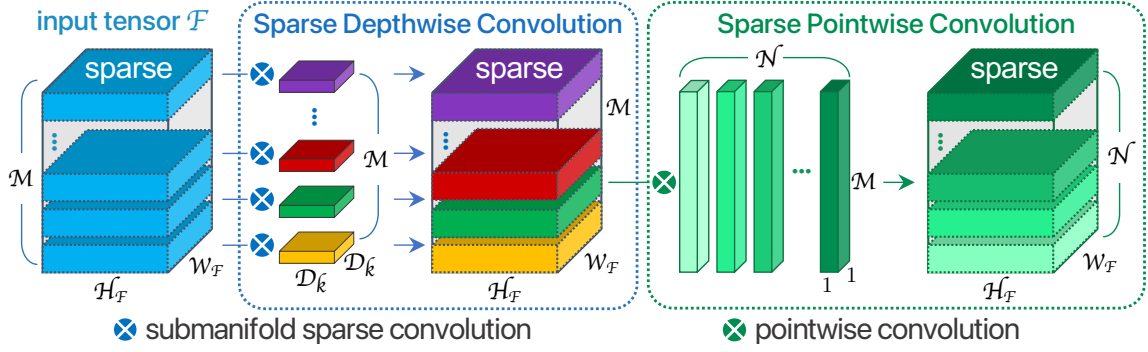

**Figure 4.7: Illustration of the SDSC** convolution module.

### 4.6.1 Sparse Depthwise Separable Convolution Module

Our SDSC module, as outlined in Figure 4.7, initially takes a tensor $F \in \mathbb{R}^{H_F \times W_F \times M}$ as input, where $H_F$, $W_F$ and $M$ denote the height, width and channels respectively. Firstly, a sparse depthwise convolution SDC($M, M, D_k, s = 1$) is applied, with $M$ input and output feature planes, a kernel size of $D_k$ and stride $s$ in order to output a tensor $T \in \mathbb{R}^{H_F \times W_F \times M}$. Inside our sparse depthwise convolution, $M$ sparse spatial convolutions are performed independently over each input channel using submanifold sparse convolution [75] due to its tensor shape preserving property at no computational or memory overhead. Secondly, the sparse pointwise convolution SPC($M, N, 1, s = 1$) projects the channels output $T$ by the sparse depthwise convolution onto a new channel space, to mix the information across different channels. As a result, the sparse Depthwise Separable Convolution SDSC($M, N, D_k, s = 1$) is the compound of the sparse depthwise convolution and the sparse pointwise convolution, namely SDSC($M, N, D_k, s = 1$) = SDC ○ SPC.

Using a sparse voxelized input representation similar to [209], and a series of such SDSC sub-modules we construct the popular Cylinder3D [20] sub-architectures within our overall Mean Teacher architectural design (Figure 4.2).

### 4.6.2 Parameters and Computation Costs Analysis

Given a Tensor $F \in \mathbb{R}^{H_F \times W_F \times L_F \times M}$, where $H_F$, $W_F$, $L_F$ and $M$ denote the radius, azimuth, height in the cylinder coordinate [20] and channels respectively. Applying convolution operation only for the active site of the sparse 3D point cloud, the computational cost (in FLOPs) of submanifold sparse convolution (SSC, [75]) is $a \times M \times N$ for the active site, where $M$ is the number of input channels as defined previously, and $N$ is the number of output channels. $a$ is the number of active inputs to the spatial location defined in [75]. The computational cost for the inactive site is $0$.

Since our SDSC sub-module consists of a sparse depthwise convolution (SDC) and a sparse pointwise convolution (SPC), the computational cost for SDSC is the sum cost of those two parts. SDC has a computational cost of

$$a \times M \times H_F \times W_F \times L_F. \tag{4.9}$$

SPC computes a linear combination of the SDC output via a $1 \times 1$ convolution, which has the computational cost of

$$M \times N \times H_F \times W_F \times L_F. \tag{4.10}$$

As a result, the computational cost of SDSC is the sum of Equations (4.9) and (4.10), *i.e.*,

$$a \times M \times H_F \times W_F \times L_F + M \times N \times H_F \times W_F \times L_F. \tag{4.11}$$

The ratio of computational cost of SDSC to SSC [75] for active site, *i.e.*, cost(SDSC) : cost(SSC), is:

$$\frac{a \times M \times H_F \times W_F \times L_F + M \times N \times H_F \times W_F \times L_F}{a \times M \times N \times H_F \times W_F \times L_F} \\ = \frac{1}{N} + \frac{1}{a} \approx \frac{1}{N} \quad (a \gg N). \tag{4.12}$$

Similar to the computational cost analysis, the parameters of SDSC is also the sum of SDC and SPC. The ratio of model parameters of SDSC to SSC [75] is:

$$\frac{\mathbf{D_K} \times M \times H_F \times W_F \times L_F + M \times N \times H_F \times W_F \times L_F}{\mathbf{D_K} \times M \times N \times H_F \times W_F \times L_F} \\ = \frac{1}{N} + \frac{1}{\mathbf{D_K}} \approx \frac{1}{\mathbf{D_K}} \quad (N \gg \mathbf{D_K}), \tag{4.13}$$

where $\mathbf{D_K}$ is the dimension of convolution kernel $K$ of size $D_{K,1} \times D_{K,2} \times D_{K,3}$, *i.e.*, $\mathbf{D_K} = D_{K,1} \times D_{K,2} \times D_{K,3}$.

LiM3D+SDSC uses SDSC sub-module as the basic building block for constructing other convolution-based modules in Cylinder3D (*e.g.*, residual block, upsample block, and downsample block). Take the residual block as an example, SDSC uses approximately $32\times$, $64\times$, $\cdots$, $512\times$ less computation than SSC for active sites when $N = \{32, 64, \cdots, 512\}$ (Equation (4.12)). SDSC-based residual block with a kernel size of $1 \times 3 \times 1$ has $3\times$ fewer parameters than the SSC-based residual block with the same kernel size, and $9\times$ fewer parameters with $3 \times 1 \times 3$ kernel size (Equation (4.13)).

## 4.7 Evaluation

We evaluate our proposed *Less is More 3D* (LiM3D) approach against state-of-the-art 3D point cloud semantic segmentation approaches using the SemanticKITTI [3] and ScribbleKITTI [4] benchmark datasets.

### 4.7.1 Experimental Setup

We detail the datasets, evaluation protocol, and implementation for LiDAR segmentation. We select a fully-labeled (SemanticKITTI) and weakly-labeled dataset (ScribbleKITTI) to compare performance under different label availability. The evaluation utilizes mIoU metric to compare SOTA methods, involving computational costs and hyperparameter analyses.

**SemanticKITTI** [3] is a large-scale 3D point cloud dataset for semantic scene understanding with 20 semantic classes consisting of 22 sequences - [`00` to `10` as *training*-split (of which `08` as *validation*-split), and `11` to `21` as *test*-split].

**ScribbleKITTI** [4] is the first scribble (*i.e.* sparsely) annotated dataset for LiDAR semantic segmentation providing sparse annotations for the *training* split of SemanticKITTI for 19 classes, with only 8.06% of points from the full SemanticKITTI dataset annotated.

**Evaluation Protocol:** Following previous work [4, 18, 20, 155], we report performance on the SemanticKITTI and ScribbleKITTI *training* set for intermediate training steps, as this metric provides an indication of the pseudo-labeling quality, and on the *validation* set to assess the performance benefits of each individual component. Performance is reported using the mean Intersection-over-Union (mIoU, as %) metric. For semi-supervised training, we report over both the benchmarks using the SemanticKITTI and ScribbleKITTI *validation* set under 5%, 10%, 20%, and 40% partitioning. We further report the relative performance of semi-supervised or scribble-supervised for ScribbleKITTI (SS) training to the fully supervised upper-bound (FS) in percentages (SS/FS) to further analyze semi-supervised performance and report the results for the fully-supervised training on both *validation* sets for reference. The trainable parameter count and number of multiply-adds (multi-adds) are additionally provided as a metric of computational cost.

**Table 4.1:** **Comparative mIoU for semi-supervised methods** with Range- and Voxel-based representation under uniform sampling (U), sequential partition (P) and ST-RFD sampling (S): bold/2nd best = **best**/2nd best; † = 1% labeled training frames only; ‡ = 50% labeled training frames only; * denotes locally reproduced result; – denotes missing result due to unavailability of reference implementation from authors.

| Repr. | Samp. | Method | | | SemanticKITTI [3] | | | | | | ScribbleKITTI [4] | | | | | |
|---|---|---|---|---|---|---|---|---|---|---|---|---|---|---|---|---|
| | | | | 5% | 10% | 20% | 40% | 50% | 100% | 5% | 10% | 20% | 40% | 50% | 100% |
| Range | U | LaserMix [18] | (2022) | – | 58.8 | 59.4 | – | 61.4 | – | – | 54.4 | 55.6 | – | 58.7 | – |
| Voxel | U | Cylinder3D [20] | (CVPR'21) | 45.4 | 56.1 | 57.8 | 58.7 | – | 67.8 | 39.2 | 48.0 | 52.1 | 53.8 | – | 56.3 |
| | U | LaserMix [18] | (2022) | – | 60.0 | <u>61.9</u> | – | 62.3 | – | – | 53.7 | 55.1 | – | 56.8 | – |
| | P | Jiang et al. [155] | (ICCV'21) | 41.8 | 49.9 | 58.8 | 59.9 | – | 65.8 | – | – | – | – | – | – |
| | U | Unal et al. [4] | (CVPR'22) | 49.9* | 58.7* | 59.1* | 60.9 | – | 68.2* | 46.9* | 54.2* | 56.5* | 58.6* | – | <u>61.3</u> |
| | S | LiM3D+SDSC (ours) | | 57.6 | 61.0 | 61.7 | 62.1 | 62.7 | 67.5 | 56.1 | 56.9 | 57.2 | 58.9 | <u>59.3</u> | 60.7 |
| | S | LiM3D (ours) | | **59.5** | **62.2** | **63.1** | **63.3** | **63.6** | **69.5** | **58.1** | **61.0** | **61.2** | **62.0** | **62.1** | **62.4** |

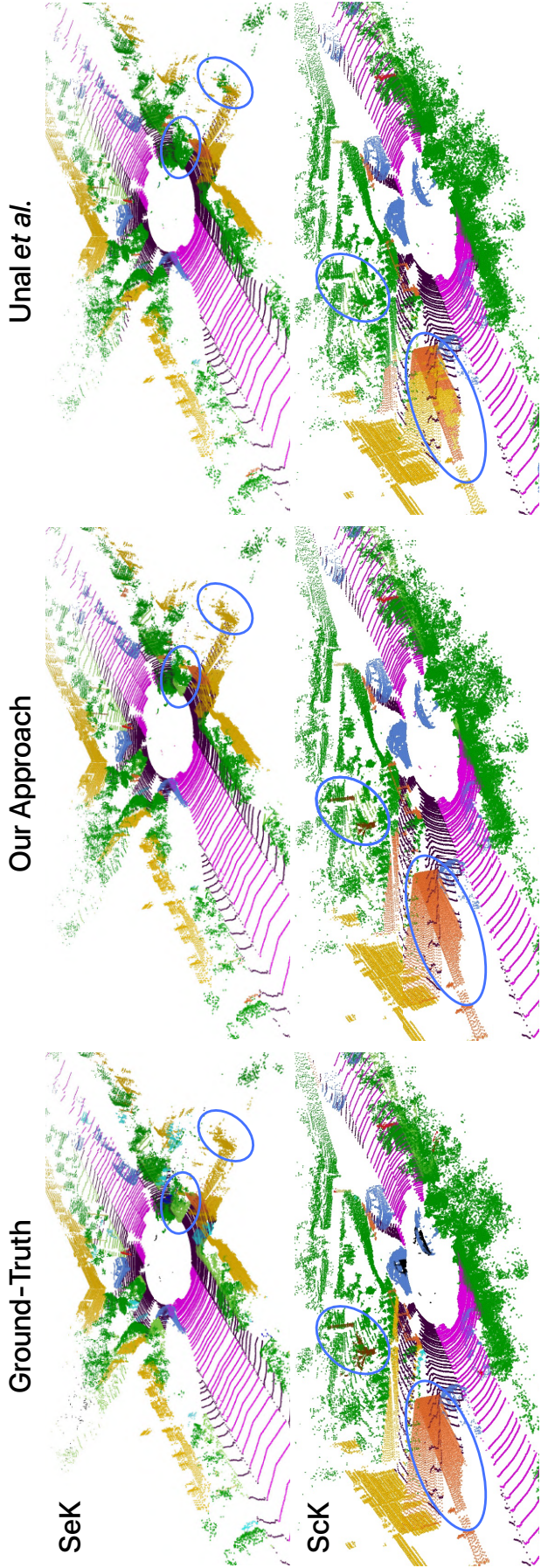

**Figure 4.8:** **Comparing the 10% sampling split** of SemanticKITTI (SeK, first row) and ScribbleKITTI (ScK, second row) *validation* set with ground-truth (left), our approach (middle) and Unal et al. [4] (right) with areas of improvement highlighted.

**Table 4.2: Component-wise ablation of LiM3D (mIoU as %, and #parameters in millions, M)** on SemanticKITTI [3] *training* and *validation* sets where UP, RF, RT, ST, SD denote Unreliable Pseudo-labeling, Reflectivity Feature, Reflec-TTA, ST-RFD, and SDSC module respectively.

| UP | RF | RT | ST | SD | Training mIoU (%) | | | | Validation mIoU (%) | | | | #Params |
|---|---|---|---|---|---|---|---|---|---|---|---|---|---|
| | | | | | 5% | 10% | 20% | 40% | 5% | 10% | 20% | 40% | (M) |
| | | | | | 82.8 | 87.5 | 87.8 | 88.2 | 54.8 | 58.1 | 59.3 | 60.8 | 49.6 |
| ✓ | | | | | – | – | – | – | 55.9 | 58.8 | 59.9 | 61.2 | 49.6 |
| ✓ | ✓ | | | | 83.6 | 88.3 | 88.7 | 89.1 | 56.8 | 59.6 | 60.5 | 61.4 | 49.6 |
| ✓ | | ✓ | | | – | – | – | – | 57.5 | 59.8 | 61.2 | 62.6 | 49.6 |
| ✓ | ✓ | ✓ | | | – | – | – | – | 58.7 | 61.3 | 62.4 | 62.8 | 49.6 |
| ✓ | ✓ | ✓ | ✓ | | **85.2** | **89.1** | **89.5** | **89.7** | **59.5** | **62.2** | **63.1** | **63.3** | 49.6 |
| ✓ | ✓ | ✓ | ✓ | ✓ | 83.8 | 88.6 | 89.0 | 89.2 | 57.6 | 61.0 | 61.7 | 62.1 | **21.5** |

**Implementation Details:** Training is performed using $4\times$ NVIDIA A100 80GB GPU without pre-trained weights with a DDP shared training strategy [210] to maintain GPU scaling efficiency, whilst reducing memory overhead significantly. Specific hyper-parameters are set as follows - Mean Teacher: $\kappa = 0.99$; unreliable pseudo-labeling: $\lambda_C = 0.3$, $\tau = 0.5$; ST-RFD: $\beta = \{7.45, 5.72, 4.00, 2.28, 0\}$ for sampling $\{5\%, 10\%, 20\%, 40\%, 100\%\}$ labeled training frames, assuming the remainder as unlabeled; Reflec-TTA: $N_b = 10$, $s = 3$ various bin sizes, following [4], we set each bin $b_i = (\rho, \phi) \in \{(20, 40), (40, 80), (80, 120)\}$.

## 4.7.2 Experimental Results

In this section, we comprehensively evaluate the performance of our proposed method on 3D semantic segmentation tasks. We highlight overall results, performance across different backbones, and qualitative assessments through visual examples, thereby validating the improvements in accuracy and efficiency.

**Overall Results on 3D Semantic Segmentation**

In Table 4.1, we present the performance of our *Less is More* 3D (LiM3D) point cloud semantic segmentation approach both with (LiM3D+SDSC) and without (LiM3D) SDSC in a side-by-side comparison with leading contemporary state-of-the-art approaches on the SemanticKITTI and ScribbleKITTI benchmark *validation* sets to illustrate our approach offers superior or comparable (within 1% mIoU) performance across all sampling ratios.

On SemanticKITTI, with a lack of available supervision, LiM3D shows a relative performance (SS/FS) from $85.6\%$ (5%-fully-supervised) to $91.1\%$ (40%-fully-supervised), and LiM3D+SDSC from $85.3\%$ to $92.0\%$, compared to their respective fully supervised upper-bound. LiM3D/LiM3D+SDSC performance is also less sensitive to reduced labeled data sampling compared with other methods.

Our model significantly outperforms on small ratio sampling splits, *e.g.*, $5\%$ and $10\%$. LiM3D shows up to $19.8\%$ and $18.9\%$ mIoU improvements whilst, with a smaller model size LiM3D+SDSC again shows significant mIoU improvements by up to $16.4\%$ and $15.5\%$ when compared with other range and voxel-based methods respectively.

Besides {5%, 10%, 20%, 40%} labeled frames training, we also report our results with less than 5% label frames shown in Table 4.3. By applying our proposed architecture for semi-supervised and scribble-supervised 3D semantic segmentation, LiM3D and LiM3D+SDSC achieve higher than 80% relative performances (SS/FS) comparing with fully-supervised methods with less than only 1% labeled, *i.e.*, 191 frames (Table 4.3).

**Performance using Different Backbones**

In Table 4.4, alongside Cylinder3D [20], we also implement our architecture with popular backbone networks [48, 154] widely-used in 3D semantic segmentation.

**Qualitative Results**

Furthermore, we present supporting qualitative results in Figure 4.8. We also show a higher-resolution version of qualitative results in Figures 4.9 and 4.10 that our method has superior performance. Figure 4.11 compares {5%, 10%, 20%, 40%} sampling splits of SemanticKITTI [3] using LiM3D (ours). Note that using our semi-supervised methodology, the results training with very few ground-truth labels (*e.g.*, 5% and 10%) can achieve comparable performance to the training with a large number of labels (see Figure 4.11, the 1-st and 2-nd rows *v.s.* 4-th row), with only subtle differences shown in the green and red circle. In Figure 4.12, the magnification of regional details shows that our method can achieve better segmentation results than other methods, especially in the category of `vegetation`, `fence`, `sidewalk`, *etc.*

**Table 4.3: 3D semantic segmentation results** of LiM3D (ours) evaluated on SemanticKITTI [3] and ScribbleKITTI [4] *valid*-set with %1 and 2% labeled data. Alongside the per-class metrics, we show the relative performance of the semi-supervised approach against the fully supervised (SS/FS). S: with SDSC sub-module (✓) or without SDSC sub-module, *i.e.*, with normal sparse convolution.

| % labeled | Dataset | S | mIoU | SS/FS | car | bicycle | motorcycle | truck | other vehicle | person | bicyclist | motorcyclist | road | parking | sidewalk | other ground | building | fence | vegetation | trunk | terrain | pole | traffic sign |
|---|---|---|---|---|---|---|---|---|---|---|---|---|---|---|---|---|---|---|---|---|---|---|---|
| LiM3D (ours) 2% / 383 frames | Semantic | | 59.3 | 85.3 | 95.6 | 37.6 | 50.1 | 54.4 | 46.0 | 68.8 | 77.1 | 0.0 | 87.9 | 32.8 | 76.6 | 2.0 | 91.4 | 54.8 | 89.5 | 69.5 | 77.1 | 66.2 | 49.4 |
| | Semantic | ✓ | 58.7 | 84.5 | 94.8 | 37.1 | 55.0 | 56.2 | 45.2 | 66.1 | 75.4 | 0.0 | 87.0 | 32.5 | 75.9 | 2.1 | 89.3 | 49.7 | 89.3 | 68.0 | 76.0 | 65.8 | 45.9 |
| | Scribble | | 58.2 | 83.7 | 92.5 | 35.6 | 52.0 | 57.1 | 49.4 | 66.8 | 78.5 | 0.0 | 85.6 | 30.2 | 74.4 | 2.4 | 89.7 | 54.0 | 88.2 | 66.9 | 74.4 | 63.7 | 47.7 |
| | Scribble | ✓ | 56.8 | 81.7 | 93.1 | 34.9 | 47.0 | 50.9 | 43.8 | 64.1 | 75.6 | 0.0 | 85.2 | 29.5 | 73.9 | 2.0 | 88.8 | 49.5 | 88.2 | 66.9 | 74.8 | 64.8 | 47.4 |
| LiM3D (ours) 1% / 191 frames | Semantic | | 58.4 | 84.0 | 92.6 | 37.5 | 51.2 | 50.4 | 47.9 | 68.6 | 80.3 | 0.0 | 86.3 | 33.5 | 74.7 | 3.9 | 88.9 | 51.4 | 88.3 | 67.4 | 75.1 | 64.8 | 45.8 |
| | Semantic | ✓ | 57.2 | 82.3 | 92.6 | 34.5 | 47.2 | 54.5 | 44.3 | 65.5 | 76.6 | 0.0 | 85.3 | 29.2 | 74.3 | 2.5 | 88.9 | 49.7 | 88.1 | 67.1 | 74.9 | 63.9 | 47.0 |
| | Scribble | | 57.0 | 82.0 | 93.1 | 31.7 | 46.8 | 55.4 | 45.2 | 65.2 | 71.8 | 0.0 | 85.3 | 29.8 | 74.0 | 2.7 | 87.3 | 50.9 | 88.3 | 67.8 | 75.4 | 64.2 | 45.9 |
| | Scribble | ✓ | 55.8 | 80.3 | 92.7 | 27.6 | 43.6 | 50.6 | 42.3 | 60.6 | 73.9 | 0.0 | 85.3 | 29.1 | 74.2 | 2.6 | 87.3 | 49.7 | 87.2 | 68.5 | 70.2 | 64.5 | 43.6 |

**Table 4.4: 3D semantic segmentation results** of LiM3D (ours) evaluated on SemanticKITTI [3] and ScribbleKITTI [4] *valid*-set with 10% labeled data, using different backbones. Alongside the per-class metrics, we show the relative performance of the semi-supervised approach against the fully supervised (SS/FS). S: with SDSC sub-module (✓) or without SDSC sub-module, *i.e.*, with normal sparse convolution.

| Model | Dataset | S | mIoU | SS/FS | car | bicycle | motorcycle | truck | other vehicle | person | bicyclist | motorcyclist | road | parking | sidewalk | other ground | building | fence | vegetation | trunk | terrain | pole | traffic sign |
|---|---|---|---|---|---|---|---|---|---|---|---|---|---|---|---|---|---|---|---|---|---|---|---|
| LiM3D (ours) + Cylinder3D [20] | Semantic | | 62.2 | 89.5 | 95.5 | 47.6 | 65.2 | 60.7 | 42.4 | 75.2 | 84.3 | 0.0 | 94.2 | 42.4 | 80.7 | 5.4 | 91.0 | 61.1 | 86.6 | 65.7 | 70.7 | 64.0 | 49.2 |
| | Semantic | ✓ | 61.0 | 87.8 | 95.3 | 43.9 | 59.2 | 46.2 | 47.0 | 71.4 | 79.8 | 1.6 | 93.8 | 44.0 | 80.0 | 4.4 | 90.8 | 60.4 | 87.7 | 64.5 | 73.5 | 63.8 | 51.4 |
| | Scribble | | 61.0 | 87.8 | 95.0 | 34.5 | 52.9 | 61.5 | 41.3 | 71.0 | 85.6 | 0.0 | 93.7 | 44.4 | 79.9 | 0.4 | 90.1 | 58.1 | 87.9 | 61.5 | 74.8 | 65.3 | 44.3 |
| | Scribble | ✓ | 56.7 | 81.6 | 95.6 | 48.9 | 45.2 | 16.0 | 43.1 | 66.9 | 81.8 | 0.0 | 91.8 | 30.9 | 75.7 | 1.8 | 90.0 | 59.2 | 86.6 | 62.6 | 69.8 | 63.4 | 46.8 |
| LiM3D (ours) + MinkowskiNet [48] | Semantic | | 60.4 | 86.9 | 94.6 | 44.3 | 47.1 | 70.2 | 29.5 | 68.7 | 80.8 | 0.0 | 93.6 | 38.4 | 79.6 | 0.1 | 90.2 | 58.7 | 88.2 | 65.5 | 75.6 | 65.5 | 58.9 |
| | Semantic | ✓ | 59.4 | 85.5 | 94.5 | 43.3 | 47.1 | 70.2 | 29.5 | 66.7 | 76.0 | 0.0 | 93.5 | 38.1 | 79.2 | 0.1 | 90.1 | 58.4 | 87.8 | 65.1 | 74.3 | 64.1 | 50.1 |
| | Scribble | | 56.2 | 80.9 | 93.8 | 42.7 | 37.6 | 68.4 | 33.7 | 54.2 | 63.5 | 0.0 | 91.9 | 39.9 | 76.7 | 0.1 | 88.9 | 61.9 | 86.9 | 67.3 | 74.3 | 59.1 | 27.7 |
| | Scribble | ✓ | 50.3 | 72.4 | 92.8 | 08.8 | 30.3 | 68.2 | 36.3 | 27.2 | 56.6 | 1.7 | 88.0 | 47.9 | 71.8 | 2.7 | 85.3 | 63.9 | 86.7 | 71.2 | 74.7 | 56.6 | 13.6 |
| LiM3D (ours) + SPVCNN [154] | Semantic | | 62.1 | 89.4 | 94.6 | 48.3 | 59.1 | 72.2 | 45.9 | 62.2 | 68.1 | 0.0 | 91.7 | 59.2 | 79.0 | 1.7 | 91.6 | 65.0 | 86.6 | 70.9 | 71.4 | 64.2 | 42.8 |
| | Semantic | ✓ | 60.8 | 87.5 | 92.7 | 29.6 | 64.8 | 72.8 | 51.2 | 47.1 | 0.0 | 0.0 | 89.0 | 58.2 | 75.2 | 1.9 | 88.8 | 68.6 | 86.4 | 73.4 | 77.8 | 63.8 | 30.8 |
| | Scribble | | 59.6 | 85.8 | 92.5 | 29.4 | 63.7 | 72.8 | 50.8 | 52.2 | 43.2 | 0.0 | 89.8 | 56.3 | 75.1 | 0.6 | 88.6 | 66.7 | 87.6 | 71.0 | 74.3 | 62.3 | 29.6 |
| | Scribble | ✓ | 54.7 | 78.7 | 91.0 | 23.1 | 61.8 | 73.1 | 45.5 | 30.0 | 36.3 | 1.6 | 87.5 | 56.2 | 72.7 | 0.7 | 86.7 | 68.3 | 87.7 | 72.8 | 75.8 | 61.3 | 19.3 |

### 4.7.3   Ablation Studies

Ablation studies are conducted to comprehensively evaluate the effectiveness of various components, proposed modules and strategies within our proposed methodologies.

**Effectiveness of Components**

In Table 4.2 we ablate each component of LiM3D step by step and report the performance on the SemanticKITTI *training* set at the end of training as an overall indicator of pseudo-labeling quality in addition to the corresponding *validation* set.

As shown in Table 4.2, adding unreliable pseudo-labeling (UP) in the distillation stage, we can increase the $valid$ mIoU by $+0.7\%$ on average in *validation* set. Appending reflectivity features (RF) in the training stage, we further improve the mIoU on the *training* set by $+0.7\%$ on average. Due to the improvements in training, the model can generate a higher quality of pseudo-labels, which results in a $+0.5\%$ increase in mIoU in the *validation* set. If we disable reflectivity features in the training stage, applying Reflec-TTA in the distillation stage alone, we then get an average improvement of $+1.3\%$ compared with pseudo-labeling only. On the whole, enabling all reflectivity-based components (RF+RT) shows great improvements of up to $+2.8\%$ in $validation$ mIoU.

Substituting the uniform sampling with our ST-RFD strategy, we observe further average improvements of $+1.0\%$ and $+0.8\%$ on $training$ and $validation$ respectively (Table 4.2).

Our SDSC module reduces the trainable parameters of our model by $57\%$, with a performance cost of $-0.7\%$ and $-1.4\%$ mIoU on $training$ and $validation$ respectively (Table 4.2). Finally, we provide two models, one without SDSC (LiM3D) and one with (LiM3D+SDSC), corresponding to the bottom two rows of Table 4.2.

**Table 4.5: The computation cost and mIoU (in percentage)** under 5%-labeled training results on SemanticKITTI (SeK) and ScribbleKITTI (ScK) *validation* set.

| Method | # Parameters | # Mult-Adds | SeK [3] | ScK [4] |
|---|---|---|---|---|
| Cylider3D [20] | 56.3 | 476.9M | 45.4 | 39.2 |
| Unal *et al.* [4] | 49.6 | 420.2M | 49.9 | 46.9 |
| 2DPASS [149] | 26.5 | <u>217.4M</u> | 51.7 | 45.1 |
| MinkowskiNet [48] | 21.7 | 114.0G | 42.4 | 35.8 |
| SPVNAS [154] | **12.5** | 73.8G | 45.1 | 38.9 |
| LiM3D+SDSC (ours) | <u>21.5</u> | **182.0M** | <u>57.6</u> | <u>54.7</u> |
| LiM3D (ours) | 49.6 | 420.2M | **59.5** | **58.1** |

**Effectiveness of SDSC Module**

In Table 4.5, we compare our LiM3D and LiM3D+SDSC with recent state-of-the-art methods under 5%-labeled semi-supervised training on the SemanticKITTI and ScribbleKITTI *validation* sets. LiM3D+SDSC outperforms the voxel-based methods [4, 20] with at least a **2.3×** reduction in model size. Similarly, with comparable model size [48, 149, 154], LiM3D+SDSC has higher mIoU in both datasets and up to **641×** fewer multiply-add operations.

**Effectiveness of ST-RFD Strategy**

In Table 4.6, we illustrate the effectiveness of our ST-RFD strategy by comparing LiM3D with two widely-used strategies in semi-supervised training, *i.e.*, random sampling and uniform sampling on SemanticKITTI [3] and ScribbleKITTI [4] *validation* set. Whilst uniform and random sampling have comparable results on both *validation* sets, simply applying our ST-RFD strategy improves the baseline by +0.90%, +0.75%, +0.60% and +0.55% on SemanticKITTI under 5%, 10%, 20% and 40% sampling protocol respectively. Furthermore, using corresponding range images of the point cloud, rather than RGB images to compute the spatio-temporal redundancy within ST-RFD (see ST-RFD-R in Table 4.6), has no significant difference on the performance.

**Table 4.6: Effects of ST-RFD sampling** on SemanticKITTI and ScribbleKITTI *validation* set (mIoU as %).

| Sampling | SemanticKITTI [3] | | | | ScribbleKITTI [4] | | | |
|---|---|---|---|---|---|---|---|---|
| | 5% | 10% | 20% | 40% | 5% | 10% | 20% | 40% |
| Random | 58.5 | 61.6 | 62.6 | 62.7 | 57.1 | 60.3 | <u>60.5</u> | 60.9 |
| Uniform | 58.7 | 61.3 | 62.4 | 62.8 | 56.9 | 60.6 | 60.3 | 61.0 |
| ST-RFD-R | <u>59.1</u> | **62.4** | <u>62.9</u> | **63.4** | <u>58.0</u> | <u>60.7</u> | **61.2** | <u>61.8</u> |
| ST-RFD | **59.5** | <u>62.2</u> | **63.1** | <u>63.3</u> | **58.1** | **61.0** | **61.2** | **62.0** |

**Table 4.7: Effects of differing reliability using pseudo voxels** on SemanticKITTI *validation* set, measured by the entropy of voxel-wise prediction. *Unreliable* and *Reliable*: selecting negative candidates with top 20% highest entropy scores and bottom 20% counterpart respectively. *Random*: sampling randomly regardless of entropy.

| Ratio | Unreliable | | Reliable | | Random | |
|---|---|---|---|---|---|---|
| | mIoU | SS/FF | mIoU | SS/FF | mIoU | SS/FF |
| 5% | **59.5** | **85.6** | 57.2 | 82.3 | 56.4 | 81.2 |
| 10% | **62.2** | **89.5** | 60.8 | 87.5 | 59.7 | 85.9 |
| 20% | **63.1** | **90.8** | 61.4 | 88.3 | 60.5 | 87.1 |
| 40% | **63.3** | **91.1** | 62.8 | 90.4 | 61.3 | 88.2 |

**Table 4.8: Reflectivity (Reflec-TTA) vs. Intensity (intensity-based TTA)** on SemanticKITTI and ScribbleKITTI *validation* set (mIoU, %).

| TTA | SemanticKITTI [3] | | | | ScribbleKITTI [4] | | | |
|---|---|---|---|---|---|---|---|---|
| | 5% | 10% | 20% | 40% | 5% | 10% | 20% | 40% |
| Intensity | 56.2 | 59.1 | 59.8 | 60.9 | 55.7 | 57.5 | 57.9 | 59.2 |
| Reflectivity | **59.5** | **62.2** | **63.1** | **63.3** | **58.1** | **61.0** | **61.2** | **62.0** |

**Effectiveness of Unreliable Pseudo-Labeling**

In Table 4.7, we evaluate selecting negative candidates with different reliability to illustrate the improvements of using unreliable pseudo-labels in semi-supervised semantic segmentation. The *"Unreliable"* selecting of negative candidates outperforms other alternative methodologies, showing the positive performance impact of unreliable pseudo-labels.

**Effectiveness of Reflec-TTA**

In Table 4.2, we compare LiM3D performance with and without Reflec-TTA and further experiment on the SemanticKITTI and ScribbleKITTI *validation* set in Table 4.8. This demonstrates that the LiDAR point-wise intensity feature $I^\circledR$, in place of the distance-normalized reflectivity feature $R^\circledR$, offers inferior on-task performance.

## 4.8 Summary

We present an efficient semi-supervised architecture for 3D point cloud semantic segmentation, which achieves *more* in terms of performance with *less* computational costs, *less* annotations, and *less* trainable model parameters (*i.e.*, *Less is More*, LiM3D). Our architecture consists of three novel contributions: the SDSC convolution module, the ST-RFD sampling strategy, and the pseudo-labeling method informed by LiDAR reflectivity. These individual components can be applied to any 3D semantic segmentation architecture to reduce the gap between semi or weakly-supervised and fully-supervised learning on task performance, whilst managing model complexity and computation costs.

After exploring efficient semi-supervised methods for 3D point cloud semantic segmentation in this chapter, it is essential to further investigate more accurate segmentation approaches. While semi-supervised methods reduce the dependency on labeled data and demonstrate promising results, fully supervised techniques have the potential to achieve higher accuracy by fully leveraging labeled datasets. In the next chapter (Chapter 5), we focus on maximizing performance in complex 3D environments by leveraging our proposed range-aware pointwise distance distribution features, which exhibit rotational and translational invariance. We therefore aim to more effectively harness these invariant features to enhance model reliability and accuracy.

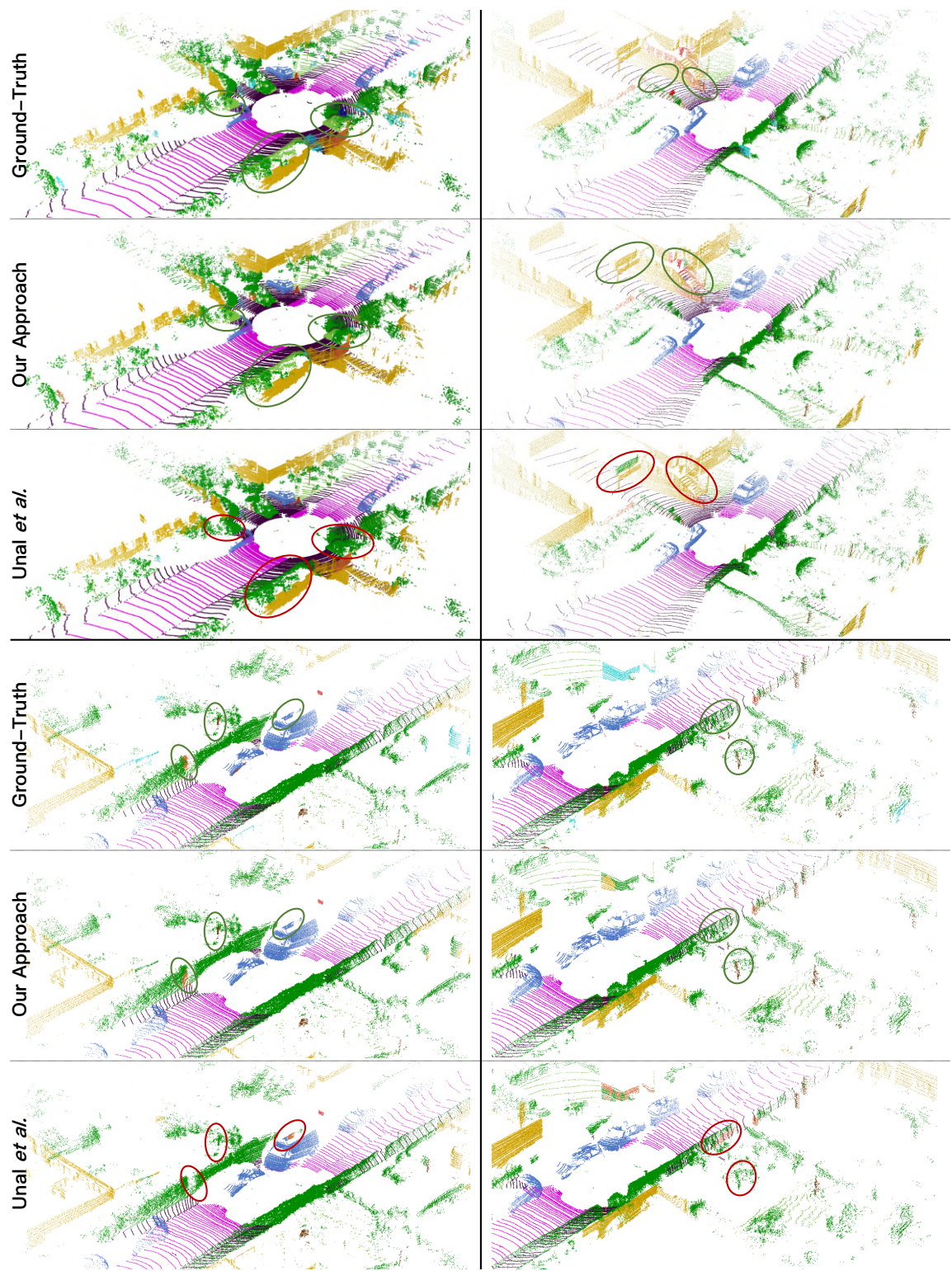

**Figure 4.9: Comparing the 10% sampling split of SemanticKITTI** [3] *validation* set with ground-truth (left), our approach (middle) and Unal *et al.* [4] (right) with areas of improvement highlighted in green, and areas of under-performance in red.

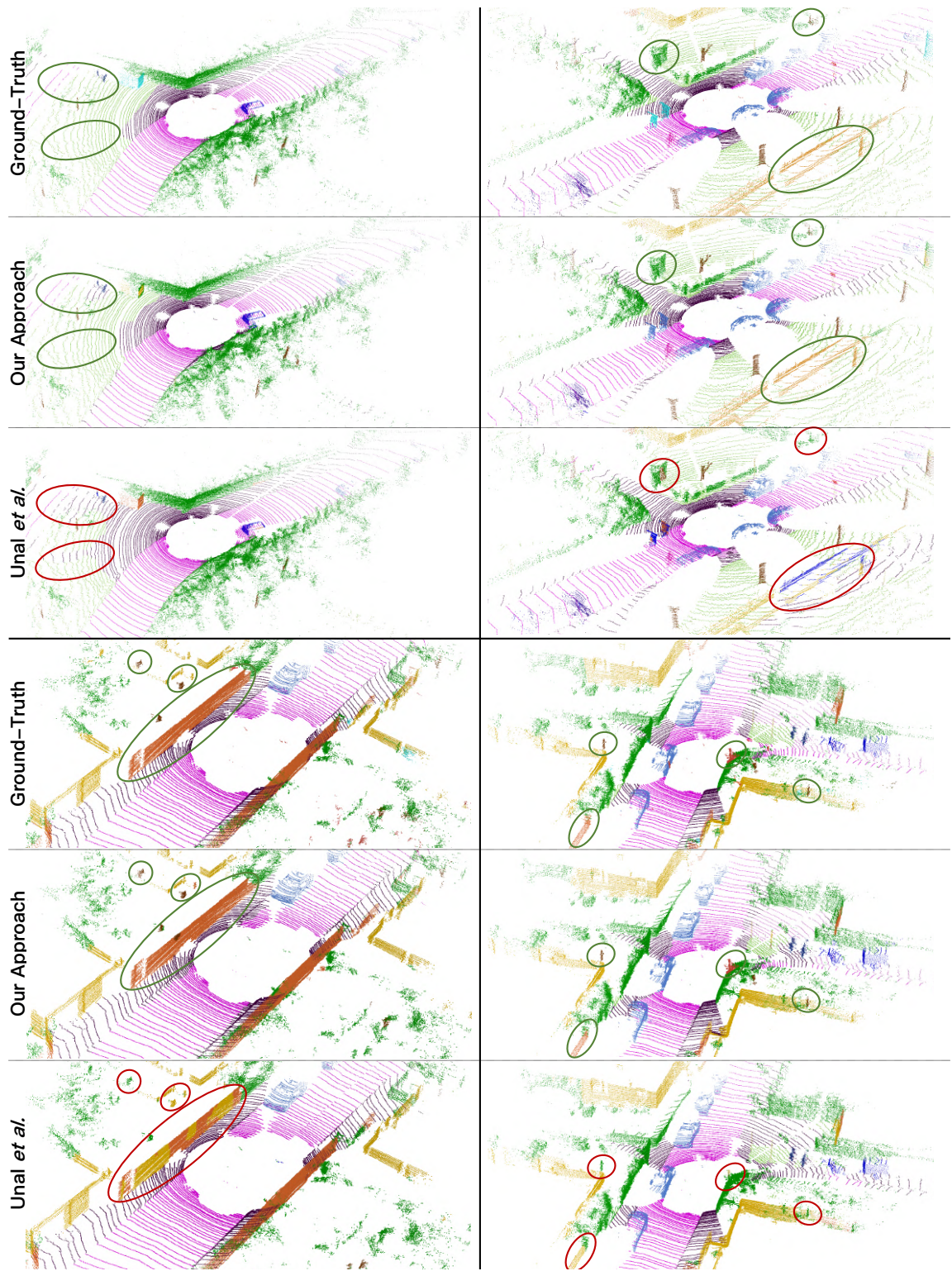

**Figure 4.10: Comparing the 10% sampling split of SemanticKITTI** [3] *validation* set with ground-truth (left), our approach (middle) and Unal *et al.* [4] (right) with areas of improvement highlighted in green, and areas of underperformance in red.

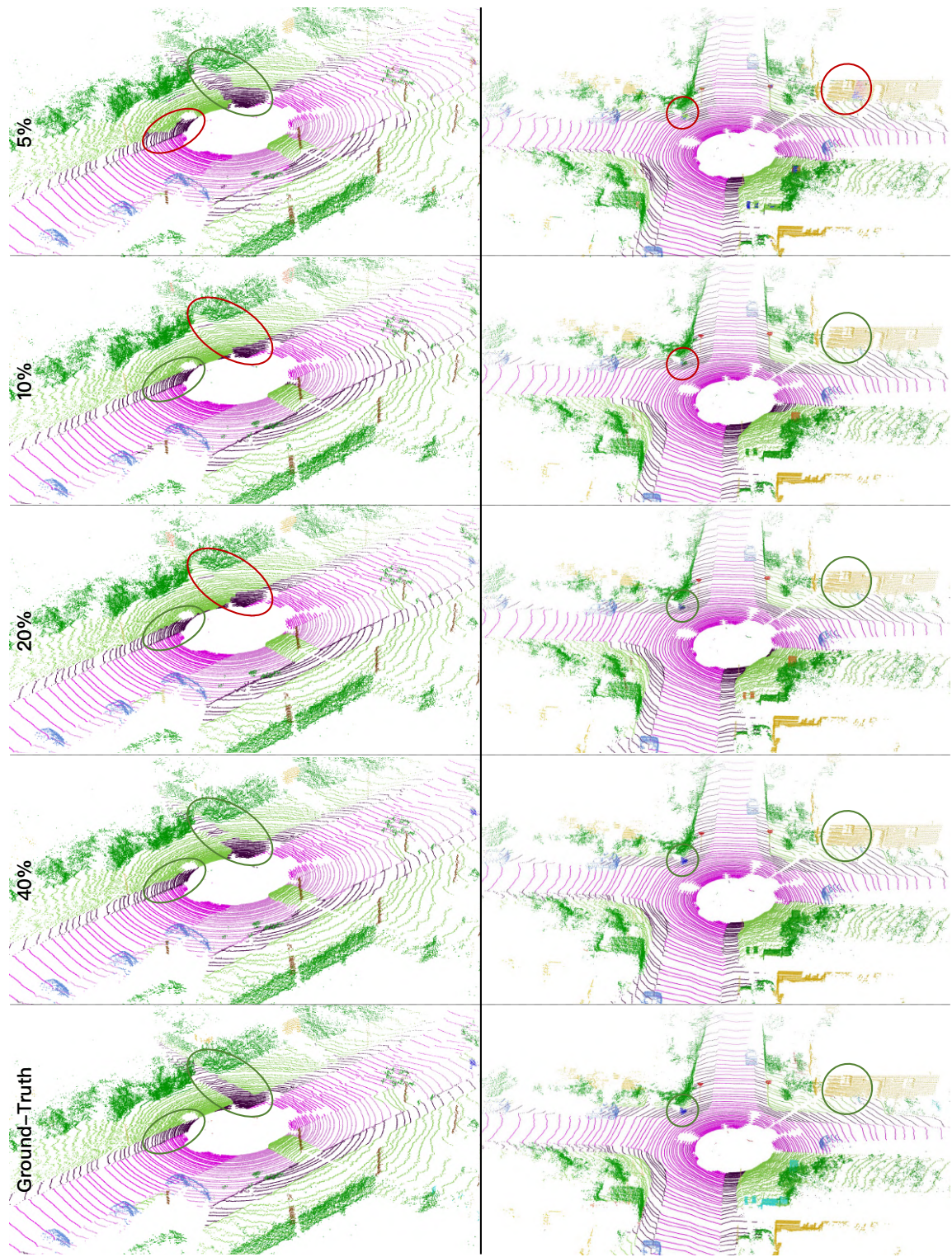

**Figure 4.11: Comparing the 5%, 10%, 20%, 40% sampling split** of SemanticKITTI [3] *validation* set with ground-truth (bottom) with areas of improvement highlighted in green, and areas of under-performance in red.

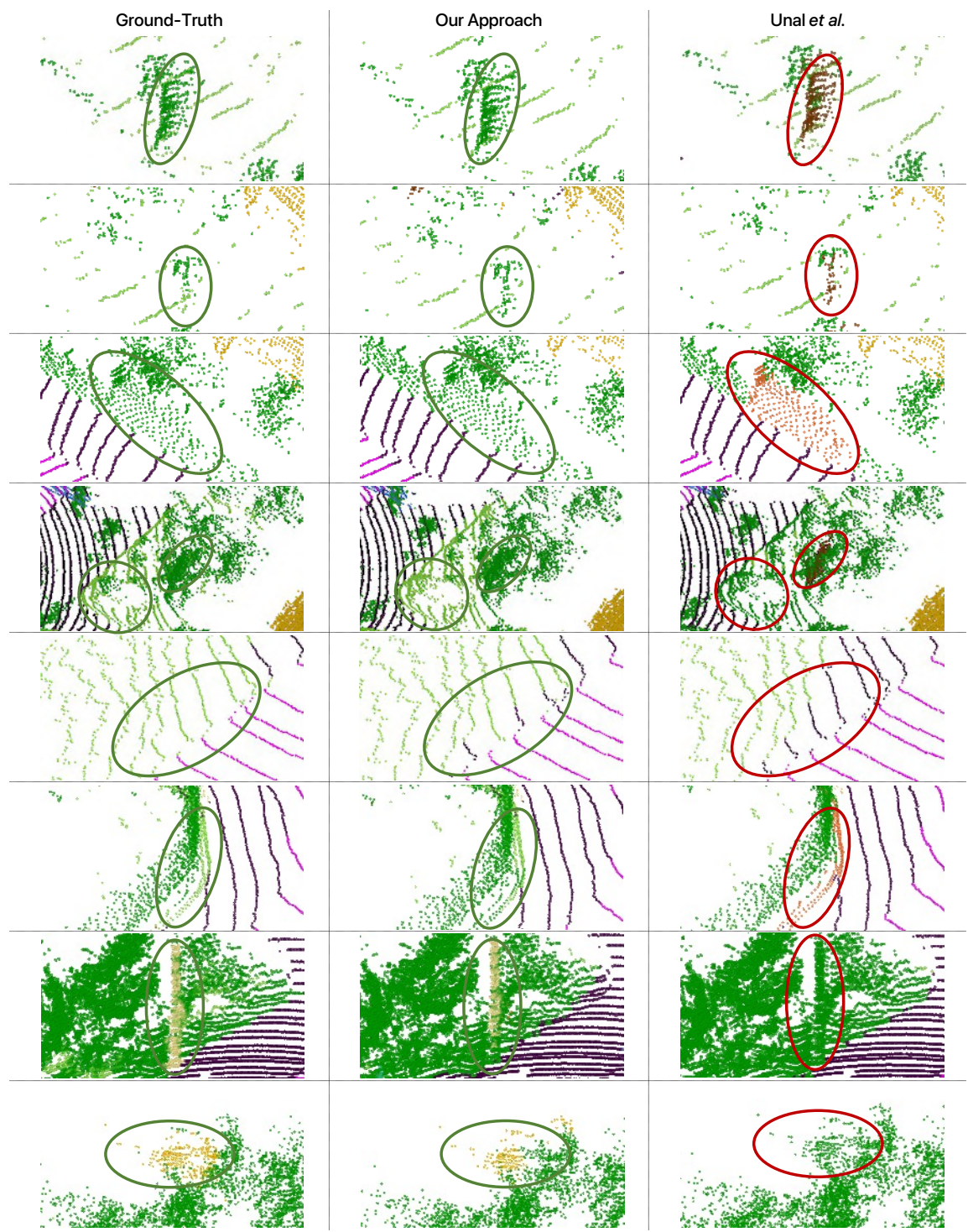

**Figure 4.12: Magnification of regional details**: comparing the 10% sampling split of SemanticKITTI [3] *validation* set with ground-truth (left), our approach (middle) and Unal *et al.* [4] (right) with areas of improvement highlighted in green, and areas of under-performance in red.

CHAPTER 5

---

## Accurate 3D LiDAR Semantic Segmentation

---

Portions of this chapter have previously been published in the following peer-reviewed publication (**oral presentation**):

- **Li, L.**, Shum, H. P., & Breckon, T. P., "RAPiD-Seg: Range-Aware Pointwise Distance Distribution Networks for 3D LiDAR Semantic Segmentation." In European Conference on Computer Vision (**ECCV**). Springer, 2024.

3D point clouds play a pivotal role in outdoor scene perception, especially in the context of autonomous driving. Recent advancements in 3D LiDAR segmentation often focus intensely on the spatial positioning and distribution of points for accurate segmentation. However, these methods, while robust in variable conditions, encounter challenges due to sole reliance on coordinates and point intensity, leading to poor isometric invariance and suboptimal segmentation. To tackle this challenge, our work introduces **R**ange-**A**ware **Poi**ntwise **D**istance Distribution (RAPiD) features and the associated RAPiD-Seg architecture. Our RAPiD features exhibit rigid transformation invariance and effectively adapt to variations in point density, with a design focus on capturing the localized geometry of neighboring structures. They utilize inherent LiDAR isotropic radiation and semantic categorization for enhanced local representation and computational efficiency, while incorporating a 4D distance metric that integrates geometric and surface material reflectivity for improved semantic segmentation. To effectively embed high-dimensional RAPiD features, we propose a double-nested autoencoder structure with a novel class-aware embedding objective to encode high-dimensional features into manageable voxel-wise embeddings. Additionally, we propose RAPiD-Seg which incorporates a channel-wise attention fusion and two effective RAPiD-Seg variants, further optimizing the embedding for enhanced performance and generalization. Our method outperforms contemporary LiDAR segmentation work in terms of mIoU on SemanticKITTI ($\mathbf{76.1}$) and nuScenes ($\mathbf{83.6}$) datasets (leaderboard rankings: $\mathbf{1^{st}}$ on both datasets [1] as of 10 November 2023).

## 5.1   Introduction

Recent literature in the domain of 3D LiDAR segmentation has witnessed significant strides, with myriad approaches attempting pointwise 3D semantic scene labeling. Leveraging color [158, 162, 163], range imagery [153, 160], and Birds Eye View (BEV) projections [153, 166], multi-modal methods integrate diverse data streams from LiDAR and other sensors to enhance feature representation and segmentation accuracy. Single-modal (LiDAR-only) methods [16, 20, 38, 148] efficiently exploit spatial data while avoiding challenges inherent to multi-modal systems, such as modality heterogeneity, the increased computational burden, and sensor overlap issues [211]. However, multi-modal approaches [163, 166] offer richer environmental insights by integrating complementary data. Our contribution, though developed for single-modal systems, can be adapted to multi-modal setups by enhancing calibration and fusion techniques [182, 212–214], improving sensor alignment and coherence, and ultimately boosting overall system performance.

Despite these advancements, a common challenge inherent to these methods is their poor isometric invariance, characterized by heavy reliance solely on coordinates and intensity data, which often results in suboptimal segmentation outcomes [16, 20]; this is primarily due to poor translational invariance and visibility (*e.g.*, occlusions), or sparse

---

observations (*e.g.*, at long range) [171], affecting data spatial distribution.

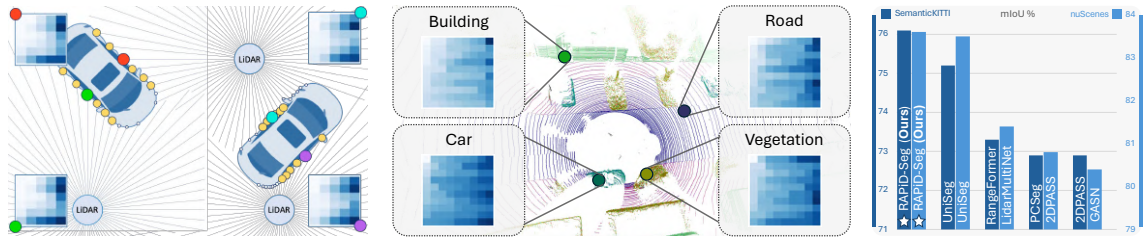

**Figure 5.1: Left:** RAPiD exhibits excellent viewpoint invariance and geometric stability, visualizing comparable features around the vehicle door structure at varying ranges and viewpoints (feature matrix plots inset). **Middle:** RAPiD is distinctive in different semantic classes, as visualized by the matrices. Embedded RAPiD patterns corresponding to different points are visualized using a spectrum of colors, showcasing their capacity to represent different classes. **Right:** Our RAPiD-Seg achieves superior results over SOTA methods on nuScenes [5] and SemanticKITTI [3].

In this chapter, we seek features that are (1) capable of capturing the localized geometric structure of neighboring points, (2) invariant to rotation and translation, and (3) applicable in noisy LiDAR outdoor environments. While numerous methods fulfill some of these requirements individually [136–138], they fall short of addressing them all [215]. Recognizing the need for higher-level features capable of capturing local geometry while potentially incorporating LiDAR point-specific attributes (*e.g.*, intensity and reflectivity), we instead turn our attention to Pointwise Distance Distribution (PDD) features [34, 35]. PDD features are recognized for their remarkable efficacy in providing a robust and informative geometric representation of point clouds, excelling in both rotational and translational invariance while including intricate local geometry details.

However, directly employing PDD features is impractical due to their high-dimensional nature, and their significant memory and storage requirements for large-scale point clouds. PDD also overlooks local features, as including distant points subsequently dilutes the focus on immediate neighborhoods.

As an enabler, we propose the **R**ange-**A**ware **Po**intwise **D**istance Distribution (RAPiD-Seg) solution for LiDAR segmentation. As shown in Figure 5.1, it utilizes invariance of RAPiD features to rigid transformations and point cloud sparsity variations, while concentrating on compact features within specific local neighborhoods. Specifically, our method leverages inherent LiDAR isotropic radiation and semantic categorization for enhanced local representations while reducing computational burdens. Moreover, our formulation computes a 4D distance, incorporating both 3D geometric and reflectivity differences to enhance semantic segmentation fidelity. To compress high-dimensional RAPiD features into tractable voxel-wise embeddings, we propose a novel embedding approach with our class-aware double-nested AutoEncoder (AE) module. We further incorporate a channel-wise attention fusion (Section 5.4.1) and two effective RAPiD-Seg variants (Section 5.4.2) to optimize the AE independently before integrating into the full network, enhancing performance and generalization capability.

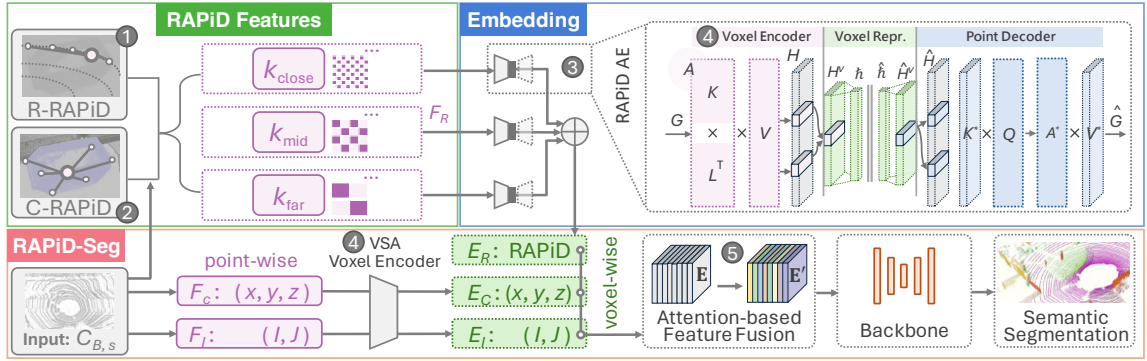

**Figure 5.2: Our proposed architecture** for 3D segmentation framework leverages RAPiD features from the point cloud. We encode pointwise features into voxel-wise embeddings via the voxel encoder and multiple RAPiD AutoEncoders (RAPiD AE). After attention-based feature fusion, these fused embeddings go through the backbone network for segmentation results.

We conduct extensive experiments on SemanticKITTI [3] and nuScenes [5] datasets, upon which our approach surpasses the state-of-the-art (SOTA) segmentation performance.

Overall, as shown in Figure 5.2, our contributions can be summarized as follows:

- A novel Range-Aware Pointwise Distance Distribution **feature (RAPiD)** for 3D LiDAR segmentation that ensures robustness to rigid transformations and viewpoints through isometry-invariant metrics within specific Regions of Interest, *i.e.*, intra-ring (R-RAPiD) and intra-class (C-RAPiD).

- A novel method for **embedding** RAPiD with class-aware double nested AE (RAPiD AE) to optimize the embedding of high-dimensional features, balancing efficiency and fidelity.

- A novel open-source network architecture **RAPiD-Seg** and supporting training methodology, to enable modular LiDAR segmentation that achieves SOTA performance on SemanticKITTI (mIoU: **76.1**) and nuScenes (mIoU: **83.6**).

## 5.2   Range-Aware Pointwise Distances (RAPiD)

In the context of LiDAR segmentation, a principal challenge lies in the lack of robustness against rigid transformations such as rotation and translation, as well as against variations in viewpoint, points sparsity, and occlusion [171, 211]. Traditional methods predominantly leverage 3D point coordinates to furnish spatial information [19, 20, 38]; however, they may be inadequate in scenes with poor visibility (*e.g.*, occlusions) or sparse observations (*e.g.*, at long range) [171]. Such reliance on coordinates alone could lead to inaccuracies in recognizing object transformations or occlusions [171]. Data augmentation, *e.g.*, random geometric transformation can improve robustness under rigid transformations [19, 20], but fail to guarantee comprehensive coverage of all potential transformations, resulting in vulnerability to previously unseen variations.

To achieve a transformation-invariant 3D data representation, we observe that distances within rigid bodies (*e.g.*, vehicles, roads, and buildings) remain constant under rigid transformations. In light of this, we focus on assimilating the principle of the isometric invariant into the LiDAR-driven point cloud perception. Specifically, we delve into the PDD [35] — an isometry invariant that quantifies distances between adjacent points. For outdoor point clouds, vanilla PDD features are computationally intensive and susceptible to noise and sparsity. Our Range-Aware Pointwise Distance Distribution (RAPiD), specifically designed for LiDAR data, instead calculates the PDD features for each point within specific Regions of Interest (RoI), which are typically associated with the intrinsic structure of LiDAR data.

### 5.2.1   Mathematical Formulation

Given a fixed number $k > 0$ representing the fixed number of point neighbors and a $u$-point cluster $P_{\mathrm{RoI}}$ comprising no fewer than $k$ points based on RoI, the RAPiD is a $u \times k$ matrix, which retains both spatial distances and LiDAR reflectivity [16] disparities between points. RAPiD are adapted to LiDAR sparsity at different distances by using range-specific parameters $k_{\mathrm{close}}$, $k_{\mathrm{mid}}$, $k_{\mathrm{far}}$ for close, mid, and far ranges, respectively.

The $k$-point RAPiD in region $P_{\text{RoI}}$ is defined as:

$$\text{RAPiD}\left(P_{\text{RoI}}; k\right) = \text{sort}\left(\left[\,\text{sort}\left(\left[\,\boldsymbol{\rho}_{j,1}, \ldots, \boldsymbol{\rho}_{j,k}\,\right]\right)\,\right]_{j=1}^{u}\right), \tag{5.1}$$

$\forall l \in \{1, \ldots, k\}, j \in \{1, \ldots, u\}$, $\boldsymbol{\rho}_{j,l}$ is given by:

$$\boldsymbol{\rho}_{j,l} = \left\|\left[\,\boldsymbol{p}_j - \boldsymbol{p}_{j,l}, \;\; g\left(r_j\right) - g\left(r_{j,l}\right)\,\right]\right\|_2, \tag{5.2}$$

where $\boldsymbol{p}_j$ and $\boldsymbol{p}_{j,l}$ denote the 3D coordinates of the $j$-th point and its $l$-th nearest neighbor within $P_{\text{RoI}}$, respectively; $r_j$ and $r_{j,l}$ represent the reflectivity values of $\boldsymbol{p}_j$ and $\boldsymbol{p}_{j,l}$, correspondingly; $\|\cdot\|_2$ is the Euclidean norm. $g : \mathbb{R} \to [D_{\min}, D_{\max}]$ is the reflectivity mapping function that maps the numerical range of reflectivity onto a consistent scale with the range of Euclidean distances between points:

$$g(r) = \left(\frac{r - r_{\min}}{r_{\max} - r_{\min}}\right)\left(D_{\max} - D_{\min}\right) + D_{\min}, \tag{5.3}$$

$$D_{\min} = \min_{j,l} \|\boldsymbol{p}_j - \boldsymbol{p}_{j,l}\|_2, \quad D_{\max} = \max_{j,l} \|\boldsymbol{p}_j - \boldsymbol{p}_{j,l}\|_2. \tag{5.4}$$

where $[D_{\min}, D_{\max}]$ is the range of the Euclidean norms of coordinate differences for all considered point pairs. The 4D distance in Equation (5.2) integrates material reflectivity into RAPiD features, enhancing feature representation and aiding in the discrimination of various materials and surfaces, which is crucial for accurate semantic segmentation.

For convenience to facilitate feature normalization and data alignment, we arrange RAPiD lexicographically by sorting Equation (5.1), where sort($\cdot$) on the inner and outer brackets sorts the elements $\boldsymbol{\rho}_{j,l}$ within each row $j$, and the sorted rows based on their first differing elements, both in ascending order [35]. It treats RAPiD matrices with the same geometric structure but different orders as identical. To further improve the data robustness, we normalize $\boldsymbol{\rho}_{j,l}$ after excluding outliers exceeding a distance threshold $\delta$, substituting these entries with the maximum of the normalized distribution to represent significant inter-point distances.

## 5.2.2 Range and Sparsity Awareness

**Preliminaries:** The parameter $k$ represents the number of neighboring points considered for each point in the cloud. For two point clouds with the same basic structure, a small $k$ value (*e.g.*, $k = 5$) will result in a small inter-point distance. A larger $k$ means that the local geometry of the point clouds must be similar over a larger radius.

Our RAPiD are adapted to LiDAR sparsity at different distances by using range-specific parameters $k_{close}$, $k_{mid}$ and $k_{far}$, for close, mid, and far ranges, respectively. Herein, we supplement the hyper-parameter analysis of various ranges $R$ and $k$.

As shown in Figure 5.3 (right), $\boldsymbol{p}_1$ and $\boldsymbol{p}_2$ are two adjacent points on the same LiDAR ring, with the LiDAR being at a range of $R$. According to the principles of trigonometric geometry, the inter-point distance between $\boldsymbol{p}_1$ and $\boldsymbol{p}_2$ is:

$$\|\boldsymbol{p}_1 - \boldsymbol{p}_2\|_2 = 2R\sin(\theta/2), \tag{5.5}$$

where $\theta$ is the LiDAR angular resolution (horizontal/azimuth). For the SemanticKITTI dataset, $\theta = 0.09 \deg$ [15]; for the nuScenes dataset, $\theta = 0.1 \sim 0.4 \deg$ [5].

We aim for our $k$-point RAPiD to focus on local geometric structures; therefore, in areas where the point cloud is sparse, we seek to avoid excessively large inter-point distances that can result from an overly large $k$. For the $k$-point RAPiD in the extreme case, we constrain the possible maximum inter-point distance from exceeding a certain threshold $\delta_{max}$:

$$\delta_{max} = (k-1)\|\boldsymbol{p}_1 - \boldsymbol{p}_2\|_2 = 2(k-1)R\sin(\theta/2). \tag{5.6}$$

Experimentally, we select $\delta_{max} = 0.25$, and subsequently determined the corresponding combinations of $k$ and $R$ via Equation (5.6). The proportional visualization in Figure 5.3 (left) demonstrates that our range division is rational, effectively distinguishing various point cloud sparsities based on range $R$, thereby facilitating the selection of appropriate $k$.

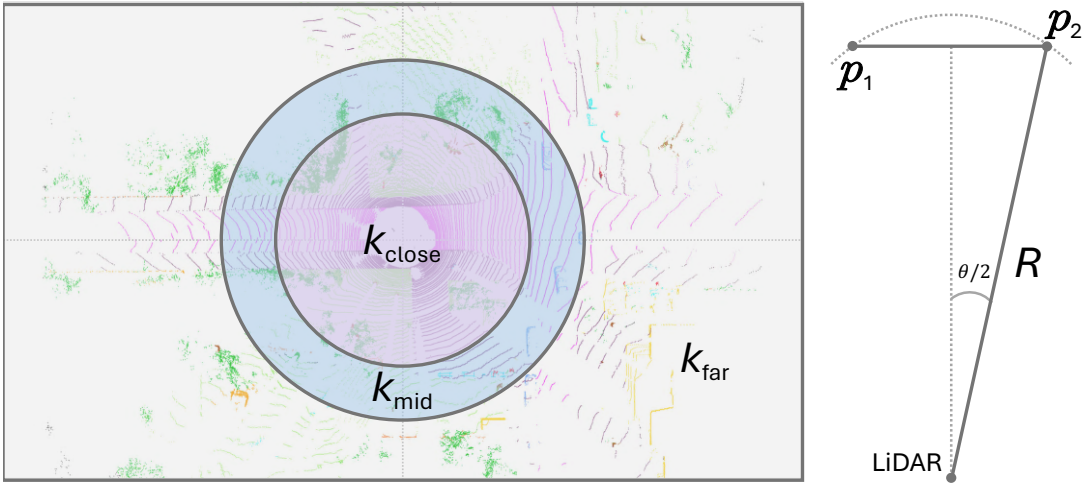

**Figure 5.3: The illustration** of range $R$ and $k$.

### 5.2.3 Intra-Ring and Intra-Class RAPiD

To enhance the RAPiD ability to capture local context, we propose R-RAPiD and C-RAPiD, based on the inherent structure of LiDAR data. The key benefit of C-RAPiD is its capacity to underscore the inherent traits of points within the same semantic class, achieved based on semantic labels before distance calculation. As a complementary, R-RAPiD is versatile, regardless of semantic label availability.

**Intra-Ring RAPiD (R-RAPiD)** is a specialized variant of RAPiD, which confines the RoI to the ring encompassing the anchor points, thereby serving to economize on computational overhead while capitalizing on the inherent structural characteristics of LiDAR data (Figure 5.2 ❶).

Given LiDAR sensors with fixed beam counts [3, 5, 17, 37], *e.g.*, $B \in \{32, 64, 128\}$, beams radiate isotropically around the vehicle at set angles. We segment the $m$-point LiDAR point cloud $C_{B,s}$ by beams, where $m = s \times B$, with $s$ representing measurements within a scan cycle and $B$ the number of laser beams. Suppose $\boldsymbol{p}_i$ with Cartesian coordinate $(x_i, y_i, z_i)$ is a point within $C_{B,s}$, its transformation to cylindrical coordinates $(\theta_i, \phi_i)$ is formulated as:

$$\theta_i = \left\lfloor \arctan 2(y_i, x_i)/\Delta\theta \right\rfloor, \tag{5.7}$$

$$\phi_i = \left\lfloor \arcsin \left[ z_i \left( x_i^2 + y_i^2 + z_i^2 \right)^{-1/2} \right] /\Delta\phi \right\rfloor, \tag{5.8}$$

where $\Delta\theta$ and $\Delta\phi$ denote the mean angular resolutions horizontally and vertically between adjacent beams.

$C_{B,s}$ is then partitioned to $B$ rings based on $\phi_b$, *i.e.*, $C_{B,s} = \bigcup_{b=0}^{B-1} R_b$, where each ring $R_b = \{(\theta_{b,i}, \phi_{b,i}) \mid \phi_{b,i} = \phi_b, \ \forall i \leq m\}$; $\bigcup$ signifies the cumulative union. R-RAPiD can thus be defined as the cumulative concatenate $\oplus$ of RAPiD in each ring $R_b$,

$$\text{R-RAPiD}(C_{B,s}; k) := \bigoplus_{b=0}^{B-1} \text{RAPiD}(R_b; k). \tag{5.9}$$

**Intra-Class RAPiD (C-RAPiD)** concentrates on extracting point features within the confines of each semantic class (Figure 5.2 ❷). It is complementary to R-RAPiD, preserving the embedding fidelity within individual semantic class.

Specifically, C-RAPiD concatenates RAPiD for each point $\boldsymbol{p}_j$ and its coresponding semantic label $y_j$ within semantic class $i$, across $N_c$ total classes:

$$\text{C-RAPiD}(\{C_{B,s}, \mathcal{Y}\}; k) := \bigoplus_{i=0}^{N_c-1} \text{RAPiD}(S_i; k), \tag{5.10}$$

where $S_i = \{\boldsymbol{p}_j \mid y_j = i, \ \forall j \leq m\}, \ \mathcal{Y} = [y_j]_{j=1}^{m}$.

## 5.3 RAPiD Embedding

The direct utilization of high-dimensional RAPiD features imposes a substantial computational burden. Efficient processing of large data volumes requires methods that can condense dimensionality while preserving data integrity [216].

We explore data embedding methods to reduce computational costs while providing feature learning capacity. AE is efficient in embedding high-dimensional features utilizing the reconstruction loss [217, 218]. However, conventional AE, limited by their sensitivity to input order and reliance on fixed-size misinputs [219], struggle to align with the unordered and variable size inherent in point clouds [21].

We propose a double nested AE structure with a novel class-aware embedding objective, using the Voxel-based Set Attention (VSA) module [21] as a building block. This facilitates superior contextual awareness and adaptability to unordered and variably-sized point cloud data.

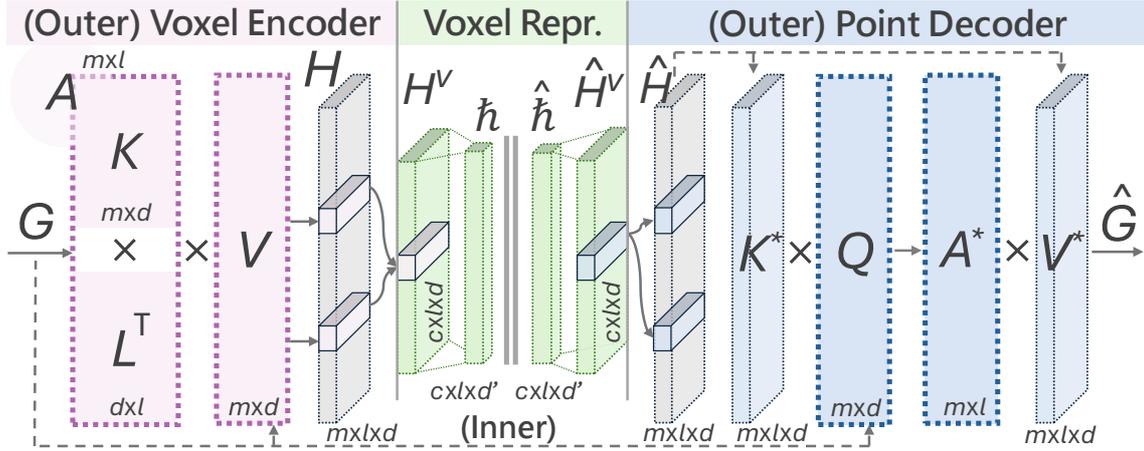

**Figure 5.4: Our RAPiD AE** consists of an Encoder, Convolution Layer, and Decoder module, aiming to reproduce the input features and generate the compressed voxel-wise RAPiD representation $\hbar$.

### 5.3.1 Nested RAPiD AE

Illustrated in Figure 5.2 ❸ and Figure 5.4, it is composed of two modules: the outer module is constituted by a VSA AE, primarily responsible for the point-to-voxel conversion. Within the inner module, which focuses on voxel-wise representation, we employ an additional AE specifically designed for dimensionality reduction. Specifically, considering an input of $m$-point RAPiD features $G \in \mathbb{R}^{m \times d}$, where each point encompasses $d$ features. We compress pointwise $G$ into a voxel-wise representation $H^v \in \mathbb{R}^{c \times l \times d}$ by an outer VSA Voxel Encoder (with $c$ reduced voxel-wise dimension and $l$ latent codes), then further reduces it to a compressed embedding $\hbar \in \mathbb{R}^{c \times l \times d'}$ with $d'$ features via an inner Encoder. The inner and outer VSA Point Decoders then reconstruct the output feature set $\hat{G}$ akin to $G$.

### 5.3.2 Inner Voxel-Wise Representation

Inner Voxel-Wise Representation is conducted via an inner AE, which takes $H^v$ from outer VSA AE as the input, yielding a lower-dimensional embedding $\hbar$ (dimension reduced from $d$ to $d'$). This process involves $u \times$ convolutional layers and a batch normalization layer [220]:

$$\hbar^{(i)} = \text{BatchNorm} \circ \text{Conv}(\hbar^{(i-1)}), \tag{5.11}$$

where $\hbar^{(0)} = H^v$, $\hbar^{(u-1)} = \hbar$. The Convolutional Feed-Forward Network (ConvFFN) then promotes voxel-level information exchange, enhancing spatial feature interactivity.

It maps reduced hidden features $\hbar$ to a 3D sparse tensor, indexed by voxel coordinates $X^v$, and employs dual depth-wise convolutions (DwConv) for spatial interactivity:

$$\hat{\hbar} = \mathrm{DwConv}^{(2)} \circ \zeta \circ \mathrm{DwConv}^{(1)}\left(\mathrm{SpT}(\hbar, X^v)\right), \tag{5.12}$$

where $\zeta$ and $\mathrm{SpT}(\cdot)$ represents the non-linear activation and sparse tensor construction. Subsequently, $\hat{\hbar} \in \mathbb{R}^{c \times l \times d'}$ is reconstructed into $\hat{H}^v \in \mathbb{R}^{c \times l \times d}$ through an inner Decoder, consisting of multiple DeConv Layers.

### 5.3.3 Outer VSA AE

Outer VSA AE consists of a voxel encoder and point encoder. The voxel encoder project RAPiD features into key-value spaces, forming $K$ and $V$, followed by a cross-attention mechanism with a latent query $L$, producing an attention matrix $A$. The cross-attention allows for an effective mechanism to query specific information from the voxelized input. By projecting the features into key-value $(K, V)$ spaces, the cross-attention mechanism lets a latent query $L$ selectively attend to the most relevant information from the voxel space. This helps the model focus on the most important aspects of the 3D environment, enabling better localization and understanding of objects in the scene.

The voxel-wise representation $H^v$ is the scatter sum [221] of pointwise $H$ to aggregate the value vectors $V$:

$$H^v = \mathrm{Sum}_{\mathrm{scatter}}(H, I^v), \quad H = \tilde{A}^\top V, \tag{5.13}$$

$$\tilde{A} = \mathrm{Softmax}_{\mathrm{scatter}}(KL^\top, I^v), \quad (Q, V) = \mathrm{Proj}(G), \tag{5.14}$$

where $\mathrm{Proj}(\cdot)$ represents the linear projection, and $I^v$ is the voxel indices. The Point Decoder reconstructs the output set from the enriched hidden features $\hat{H}^v$. We start by broadcasting $\hat{H}^v$ based on $I^v$, resulting in $\hat{H} \in \mathbb{R}^{m \times l \times d}$. The output $\hat{G}$ is analogous to the operational paradigm of Voxel Encoder:

$$\hat{G} = \tilde{A}^{*\top} V^*, \quad \tilde{A}^* = \mathrm{Softmax}(A^*), \tag{5.15}$$

$$A^* = \left[K_i^* Q_i^\top\right]_{i=1}^{m}, \quad (K^*, V^*) = \mathrm{Proj}(\hat{\mathcal{H}}). \tag{5.16}$$

### 5.3.4 Class-Aware Embedding Objective

Class-Aware Embedding Objective addresses the issue of non-uniqueness in embeddings produced by the AE, where various distinct inputs yield approximately the same embedding, leading to inaccurate representation [222, 223]. Our objective aims to facilitate the generation of AE embeddings that demonstrate robust semantic class discriminability.

Specifically, we introduce a novel class-aware contrastive loss $\mathcal{L}_{\text{contr}}$ in Equation (5.17), which aims to maximize the distance between embeddings of different semantic classes while minimizing the distance of the same class.

$$\mathcal{L}_{\text{contr}} = \frac{1}{m} \sum_{i=1}^{m} \left[ \frac{1}{|P(i)|} \sum_{p \in P(i)} \text{ReLU}(\alpha_p - \text{sim}(H_i, H_p)) \right.$$
$$\left. + \frac{1}{|N(i)|} \sum_{n \in N(i)} \text{ReLU}(\text{sim}(H_i, H_n) - \alpha_n) \right], \tag{5.17}$$

$$\mathcal{L}_{\text{recon}} = \frac{1}{m \cdot d} \sum_{i=1}^{m} \sum_{j=1}^{d} (G_{i,j} - \hat{G}_{i,j})^2, \tag{5.18}$$

where $i$ represents the point index; $P(i)$ and $N(i)$ are the indices of the nearest point in the same (Positive) and different class (Negative) relative to point $i$, respectively; $\text{sim}(\cdot)$ is a similarity measure; $\alpha_p$ and $\alpha_n$ is the positive and negative margin controlling separation between classes, respectively. The pointwise representations $H_i$, $H_p$, and $H_n$ are broadcasted from $\hbar$ based on voxel indices $I^v$.

We further combine this with the MSE [224] reconstruction loss in Equation (5.18). The overall loss function, with $\lambda$ balancing the fidelity of reconstruction with the distinctiveness of the embeddings, is formulated as: $\mathcal{L}_{\text{total}} = \mathcal{L}_{\text{recon}} + \lambda \mathcal{L}_{\text{contr}}$.

### 5.3.5 Multi-Neighboring-Point Stacked RAPiD

To facilitate the local representation of RAPiD and robustness against noise, we implement point-wise RAPiD features as multi-neighboring-point stacked (MNPS) features. Specifically, for an anchor point $\boldsymbol{p}$, the $k$-point RAPiD features are implemented by computing the $(k-1)$-point RAPiD among the sub-pointcloud formed by the $k$ nearest neighbors of the anchor point $\boldsymbol{p}$. This $(k-1)$-point RAPiD thus serves as the MNPS

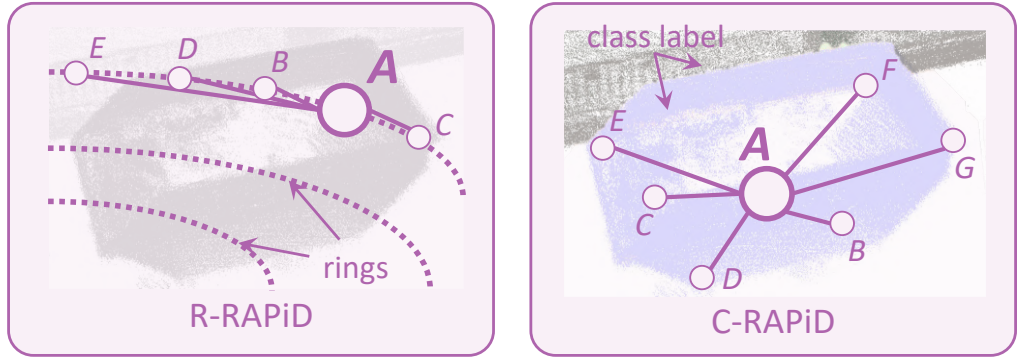

**Figure 5.5: Visual illustration** of R-RAPiD and C-RAPiD. R-RAPiD (left) confines the RoI to the ring surrounding anchor points A (*e.g.*, B, C, D, and E), optimizing computational efficiency by leveraging the structural characteristics of LiDAR data. C-RAPiD (right) focuses on point features within the same semantic class (*e.g.*, A, B, C, D, E, F, and G), preserving feature embedding fidelity.

implementation for the current anchor point $\boldsymbol{p}$.

Intuitively, our implementation of MNPS resembles a sliding window mechanism, wherein each anchor point and its neighboring points are considered a small window. This window slides over anchor points to compute the RAPiD features of the sub-windows. Consequently, our MNPS implementation inherits the advantages of sliding windows in extracting local features, flexibility (various window sizes and sliding steps), computational efficiency (reusing part of the results from the previous window), and noise reduction.

As shown in Figure 5.5, we take the computation of 5-point R-RAPiD (left) and 7-point C-RAPiD (right) as an example. For 5-point R-RAPiD of anchor point $A$, let point $B$, $C, \cdots, E$ be the nearest neighborhoods of $A$ in their belonged ring (the local distances in Figure 5.5 are magnified for better visualization). The MNPS implementation $\mathbf{M}_A$ of the 4-point R-RAPiD within a sub-pointcloud composed of $B, \cdots, E$ is:

$$\mathbf{M}_A = \begin{bmatrix} \boldsymbol{\rho}_{A,B}, & \boldsymbol{\rho}_{A,C}, & \boldsymbol{\rho}_{A,D}, & \boldsymbol{\rho}_{A,E} \\ \boldsymbol{\rho}_{B,A}, & \boldsymbol{\rho}_{B,D}, & \boldsymbol{\rho}_{B,C}, & \boldsymbol{\rho}_{B,E} \\ \boldsymbol{\rho}_{C,A}, & \boldsymbol{\rho}_{C,B}, & \boldsymbol{\rho}_{C,D}, & \boldsymbol{\rho}_{C,E} \\ \boldsymbol{\rho}_{D,B}, & \boldsymbol{\rho}_{D,E}, & \boldsymbol{\rho}_{D,A}, & \boldsymbol{\rho}_{D,C} \\ \boldsymbol{\rho}_{E,D}, & \boldsymbol{\rho}_{E,B}, & \boldsymbol{\rho}_{E,A}, & \boldsymbol{\rho}_{E,C}, \end{bmatrix}. \tag{5.19}$$

where $\boldsymbol{\rho}$ is the 4D distance defined in Equation (5.2). The computation of C-RAPiD follows a similar approach, with the only difference being that the selection of neighboring points is confined within the semantic category to which the anchor point belongs.

## 5.4 RAPiD-Seg for 3D LiDAR Segmentation

We present RAPiD-Seg, a 3D LiDAR segmentation network leveraging RAPiD features. It incorporates multiple complementary features via the channel-wise fusion mechanism [225]. RAPiD-Seg takes point cloud as input and performs single-modal (LiDAR-only) 3D semantic segmentation. Specifically, the input point cloud has three types of point-wise features: (a) **coordinates-based features** $F_C = \{p_C \mid p_C = (x, y, z)\}$; (b) **intensity-based features** $F_I = \{p_I \mid p_I = (I, J)\}$, where $I$ and $J$ are intensity and reflectivity; (c) **RAPiD features** $F_R = \{p_R \mid p_R = \text{RAPiD}(P_{\text{RoI}}; k)\}$. The RoI varies depending on whether it is R- or C-RAPiD.

As shown in Figure 5.2, $F_R$ and $F_C \oplus F_I$ are fed to a RAPiD AE ❸ and a VSA voxel encoder ❹ for voxelization, resulting in voxel-wise representation $E_R$, $E_C$, and $E_I$. Subsequently, we incorporate the complementary voxel-wise features via the channel-wise fusion mechanism (Section 5.4.1). The backbone net takes in them for LiDAR segmentation prediction. Additionally, we introduce two effective RAPiD-Seg variants to expedite the convergence of the extensive segmentation network towards an optimal solution.

### 5.4.1 RAPiD Channel-Wise Fusion with Attention

To combine various LiDAR point attributes, a highly efficient method for feature fusion is still a research topic [16, 19]. Current methods, which typically concatenate different LiDAR attribute embeddings (*e.g.*, intensity [162, 226, 227], reflectivity [16], PLS [19]) face issues with increased dimensionality and complexity, and the need for balancing weights to avoid biased training towards dominant attributes.

We thus propose a novel RAPiD Feature Fusion module (RAPiD-Fuse), pioneering the application of the channel-wise attention mechanism [225] in 3D LiDAR point attribute fusion. With RAPiD-Fuse, we effectively fuse coordinates (C), intensity & reflectivity (I), and RAPiD features (R) of all voxels, emphasizing informative features adaptively while suppressing less relevant ones across different channels.

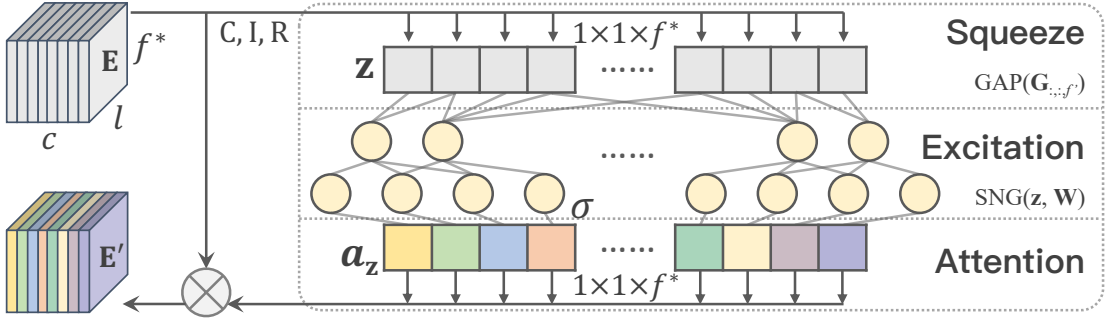

**Figure 5.6: Our feature fusion module** with channel-wise attention.

As shown in Figure 5.2 ⑤ and Figure 5.6, we initially concatenate 3 different types of features (C, I, R) along the voxel dimension, resulting in $\mathbf{E}$. In the squeeze operation, Global Average Pooling (GAP) [228] condenses each channel $f'$ of the tensor into a representative channel descriptor. These descriptors form a channel-level embedding $\mathbf{z} = [z_{f'}]_{f'=1}^{f^*}$:

$$\mathbf{z} = [\text{GAP}(\mathbf{E}_{:,:,f'})]_{f'=1}^{f^*} = \frac{1}{c \times l} \left[ \sum_{c'=0}^{c-1} \sum_{l'=0}^{l-1} \mathbf{E}_{c',l',f'} \right]_{f'=1}^{f^*}, \qquad (5.20)$$

where $f^*$ is the dimension of the concatenated feature. A channel-wise attention $\boldsymbol{a_z} \in \mathbb{R}^{1 \times 1 \times f^*}$ is then computed through the excitation step: $\boldsymbol{a_z} = \sigma \left( \mathbf{W}_2 \, \delta(\mathbf{W}_1 \mathbf{z}) \right)$, where $\delta$ and $\sigma$ represent the ReLU and Sigmoid activation. With each element in $\boldsymbol{a_z}$ reflects the attention allocated to the corresponding feature of the voxel, the channel-wise fused features $\mathbf{E}'$ can be computed as $\mathbf{E}' = \boldsymbol{a_z} \cdot \mathbf{E}$.

## 5.4.2 RAPiD-Seg Architectures for 3D Segmentation

The complexity inherent in 3D LiDAR-driven networks typically exacerbates the end-to-end training process [16, 19, 38], since the extensive parameter space increases the propensity for overfitting, slow convergence, and the possibility of settling into local minima [16].

As shown in Figure 5.7, we propose two novel and effective variants of RAPiD-Seg for quicker and better performance. R-RAPiD-Seg involves AE training from scratch, followed by C-RAPiD-Seg utilizing C-RAPiD features with the pre-trained AE and backbone for enhanced performance.

**R-RAPiD-Seg.** We first construct the lightweight R-RAPiD-Seg (Figure 5.7 (a)) for fast 3D

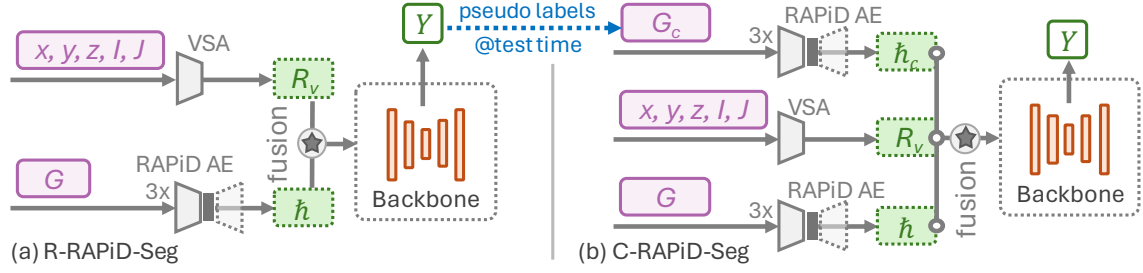

**Figure 5.7:** Two variants of RAPiD-Seg. (a) **R-RAPiD-Seg** utilizes R-RAPiD features, and (b) **C-RAPiD-Seg** utilizes both R- and C-RAPiD features for better performance.

segmentation. R-RAPiD-Seg adopts an early fusion scheme utilizing LiDAR original points and R-RAPiD features. Specifically, 3D coordinate $F_C$ and intensity-based features $F_I$ are voxelized to voxel-wise representations $R_v = E_C \oplus E_I$ based on VSA voxelization [21]. We get the compressed $d'$-dimension voxel-wise R-RAPiD representations $\hbar$ (Section 5.3) from RAPiD AE. $\hbar$ is then fused into the voxel-wise representations $R_{v*} = \text{FuAtten}(R_v, \hbar)$, where FuAtten($\cdot$) is Fusion with Attention in Section 5.4.1. The fused voxel-wise features are taken into the backbone network for 3D segmentation.

**C-RAPiD-Seg.** To facilitate the embedding fidelity within individual semantic classes, we also design a class-aware framework with C-RAPiD features, *i.e.*, C-RAPiD-Seg (Figure 5.7 (b)). Specifically, we fuse the voxel-wise C-RAPiD $\hbar_C$ into the representations $R_{v*} = \text{FuAtten}(R_v, \hbar \oplus \hbar_C)$. C-RAPiD requires class labels to compute the features regarding the semantic categories. Since the ground-truth labels are missing during the test time, we generate the reliable pseudo labels $\tilde{\mathcal{Y}}$ with a pretrained R-RAPiD-Seg based on confidence (Figure 5.7 (b) blue dotted arrow). Subsequently, the pseudo voxel-wise C-RAPiD $\hbar_C$ are generated through $\tilde{\mathcal{Y}}$. We fuse them into the voxel-wise representations for 3D segmentation with the bonebone network.

## 5.5 Evaluation

Following the popular practice of LiDAR segmentation methods [20, 38, 163], we evaluate our proposed RAPiD-Seg network against SOTA 3D LiDAR segmentation approaches on the SemanticKITTI [3] and nuScenes [5] datasets.

### 5.5.1 Experimental Setup

For the experiments, we employ mIoU metric for evaluation. We construct a point-voxel backbone based on Minkowski-UNet34 [48], and detail our training strategies and hardware.

**Datasets:** SemanticKITTI [3] comprises 22 point cloud sequences, with sequences `00-10`, `08`, and `11-21` for training, validation, and testing. 19 classes are chosen for training and evaluation by merging classes with similar motion statuses and discarding sparsely represented ones. Meanwhile, nuScenes [5] has 1000 driving scenes; 850 are for training and validation, with the remaining 150 for testing. 16 classes are used for LiDAR semantic segmentation, following the amalgamation of akin classes and the removal of rare ones.

**Evaluation Protocol:** Following the popular practice of [3,6,20], we adopt the Intersection-over-Union (IoU) of each class and mean IoU (mIoU) of all classes as the evaluation metric. The IoU of class $i$ is defined as $\text{IoU}_i = \text{TP}_i / (\text{TP}_i + \text{FP}_i + \text{FN}_i)$, where $\text{TP}_i$, $\text{FP}_i$ and $\text{FN}_i$ denote the true positive, false positive and false negative of class $i$, respectively.

**Implementation Details:** We construct the point-voxel backbone based on the Minkowski-UNet34 [48] (re-implemented by PCSeg [6] codebase), which is the open-access backbone with SOTA results to date. Before the AE training stage, we first generate pointwise RAPiD features (Section 5.2), which yield three RAPiD outputs for each frame, corresponding to different $k$ values based on range. Notably, this stage does not impose additional burdens on the subsequent overall training. The number of maximum training epochs for AE and the whole network is set as 100 and the initial learning rate is set as $10^{-3}$ with SGD optimizer. We use 2 epochs to warm up the network and adopt the cosine learning rate schedule for the remaining epochs. All experiments are conducted on $4\times$ NVIDIA A100 GPU ($1\times$ for inference).

### 5.5.2 Experimental Results

We conduct the performance evaluation of the proposed LiDAR semantic segmentation method, RAPiD-Seg, against state-of-the-art (SOTA) techniques. Focusing on the SemanticKITTI and nuScenes *test* set, the RAPiD-Seg method shows substantial improvements, particularly in segmenting rigid objects, due to its unique approach of integrating 3D geometry and material reflectivity. The study also explores cross-dataset evalua-tion, demonstrating the stability and adaptability across different LiDAR resolutions and datasets. Additionally, qualitative results include error maps and detailed visualizations, highlighting the accuracy and consistency in segmentation compared to baseline methods. The efficacy of the RAPiD is further evidenced through visualizations of their embeddings, indicating their stability and distinctiveness across semantic categories.

**Quantitative Results**

In this section, we provide quantitative results, including the comparison results of SOTA fully-supervised LiDAR semantic segmentation methods and results on cross-dataset performance.

**Comparing with SOTA Methods:** In Table 5.1 and Table 5.2, we showcase the performance of our RAPiD-Seg LiDAR segmentation method on SemanticKITTI and nuScenes *test* set in comparison with published leading contemporary SOTA approaches to demonstrate its superior efficacy. Our method significantly outperforms others, especially for rigid object categories (*e.g.*, `truck`, `o.veh`, `park`, *etc.*), primarily due to our fusion of both localized 3D geometry and material reflectivity within RAPiD features, which enable segmentation based on material properties and local rigid structures. Remarkably, our single-modal methodology outperforms multi-modal approaches [148, 153, 160, 163, 164, 166], suggesting superior efficacy of our RAPiD features over alternative modalities like RGB and range images. Our inference time (105ms per frame) is comparable to other contemporary approaches [20, 163, 164].

**Cross-Dataset Evaluation:** In Table 5.3, we showcase the self-adaptivity of our RAPiD features across different datasets. We start by simulating a downsampling in vertical resolution which uniformly sub-samples the beam by 50% (32 channels). The slight drop in mIoU from 73.02 in Conf. (a.1) to 72.60 in (a.2) suggests that the RAPiD features are stable and almost unaffected across different LiDAR resolutions and sparsity. Conf. (a.3) and (b.2) present a cross-dataset AE training with both SemanticKITTI and nuScenes datasets. When calculating the class-aware contrastive loss (*c.f.* Equation (5.17)), we merge the labels of semantically equivalent classes from the two datasets. The slight increases in mIoU from 72.60 in (a.2) to 72.78 in (a.3), and from 79.91 in (b.1) to 80.56 in (b.2), suggest that using a combination of datasets for AE training offers a marginal benefit over a single dataset AE training approach. This improvement may be attributed to the model's exposure to a broader and more diverse set of scenes during the training, which may not be adequately represented within a single dataset. It is important to note that, to ensure a fair comparison with other methodologies, a single dataset is utilized for training in all other experiments, with the exception of this particular evaluation.

**Table 5.1: Quantitative results** of RAPiD-Seg and SOTA segmentation methods on SemanticKITTI [3] *test* set; **Best**/2nd best highlighted.

| Method | mIoU | car | bicy | moto | truc | o.veh | ped | b.list | m.list | road | park | walk | o.gro | build | fenc | veg | trun | terr | pole | sign |
|---|---|---|---|---|---|---|---|---|---|---|---|---|---|---|---|---|---|---|---|---|
| AMVNet [153] | 65.3 | 96.2 | 59.9 | 54.2 | 48.8 | 45.7 | 71.0 | 65.7 | 11.0 | 90.1 | 71.1 | 75.8 | 32.4 | 92.4 | 69.1 | 85.6 | 71.7 | 69.6 | 62.7 | 67.2 |
| JS3C-Net [156] | 66.0 | 95.8 | 59.3 | 52.9 | 54.3 | 46.0 | 69.5 | 65.4 | 39.9 | 88.9 | 61.9 | 72.1 | 31.9 | 92.5 | 70.8 | 84.5 | 67.9 | 67.9 | 60.7 | 68.7 |
| SPVNAS [154] | 66.4 | 97.3 | 51.5 | 50.8 | 59.8 | 58.8 | 65.7 | 65.2 | 43.7 | 90.2 | 67.6 | 75.2 | 16.9 | 91.3 | 65.9 | 86.1 | 73.4 | 71.0 | 64.2 | 66.9 |
| Cylinder3D [20] | 68.9 | 97.1 | 67.6 | 63.8 | 50.8 | 58.5 | 73.7 | 69.2 | 48.0 | 92.2 | 65.0 | 77.0 | 32.3 | 90.7 | 66.5 | 85.6 | 72.5 | 69.8 | 62.4 | 66.2 |
| AF2S3Net [158] | 69.7 | 94.5 | 65.4 | 86.8 | 39.2 | 41.1 | 80.7 | 80.4 | 74.3 | 91.3 | 68.8 | 72.5 | 53.5 | 87.9 | 63.2 | 70.2 | 68.5 | 53.7 | 61.5 | 71.0 |
| RPVNet [148] | 70.3 | 97.6 | 68.4 | 68.7 | 44.2 | 61.1 | 75.9 | 74.4 | 73.4 | 93.4 | 70.3 | 80.7 | 33.3 | 93.5 | 72.1 | 86.5 | 75.1 | 71.7 | 64.8 | 61.4 |
| SDSeg3D [161] | 70.4 | 97.4 | 58.7 | 54.2 | 54.9 | 65.2 | 70.2 | 74.4 | 52.2 | 90.9 | 69.4 | 76.7 | 41.9 | 93.2 | 71.1 | 86.1 | 74.3 | 71.1 | 65.4 | 70.6 |
| GASN [165] | 70.7 | 96.9 | 65.8 | 58.0 | 59.3 | 61.0 | 80.4 | 82.7 | 46.3 | 89.8 | 66.2 | 74.6 | 30.1 | 92.3 | 69.6 | 87.3 | 73.0 | 72.5 | 66.1 | 71.6 |
| PVKD [38] | 71.2 | 97.0 | 67.9 | 69.3 | 53.5 | 60.2 | 75.1 | 73.5 | 50.5 | 91.8 | 70.9 | 77.5 | 41.0 | 92.4 | 69.4 | 86.5 | 73.8 | 71.9 | 64.9 | 65.8 |
| 2DPASS [164] | 72.9 | 97.0 | 63.6 | 63.4 | 61.1 | 61.5 | 77.9 | 81.3 | 74.1 | 89.7 | 67.4 | 74.7 | 40.0 | 93.5 | 72.9 | 86.2 | 73.9 | 71.0 | 65.0 | 70.4 |
| PCSeg [6] | 72.9 | 97.5 | 51.2 | 67.6 | 58.6 | 68.6 | 78.3 | 80.9 | 75.6 | 92.5 | 71.5 | 78.3 | 36.9 | 93.1 | 71.4 | 85.4 | 73.6 | 69.9 | 66.1 | 68.7 |
| RangeFormer [160] | 73.3 | 96.7 | 69.4 | 73.7 | 59.9 | 66.2 | 78.1 | 75.9 | 58.1 | 92.4 | 73.0 | 83.9 | 42.4 | 93.4 | 73.3 | 86.6 | 73.3 | 72.8 | 66.4 | 66.6 |
| UniSeg [163] | 75.2 | 97.9 | 71.9 | 75.2 | 63.6 | 74.1 | 78.9 | 74.8 | 60.6 | 92.6 | 74.0 | 79.5 | 46.1 | 93.4 | 72.7 | 87.5 | 76.3 | 73.1 | 68.3 | 68.5 |
| **RAPiD-Seg (Ours)** | 76.1 | 97.7 | 71.1 | 76.2 | 72.5 | 80.7 | 79.9 | 79.1 | 59.8 | 91.8 | 78.2 | 78.6 | 46.0 | 93.6 | 72.1 | 86.9 | 74.6 | 72.3 | 65.9 | 68.5 |

**Table 5.2: Quantitative results** of RAPiD-Seg and top-10 SOTA segmentation methods on nuScenes [5] *test* set; **Best**/2nd best highlighted.

| Method | mIoU | barr | bicy | bus | car | const | motor | ped | cone | trail | truck | driv | other | walk | terr | made | veg |
|---|---|---|---|---|---|---|---|---|---|---|---|---|---|---|---|---|---|
| PMF [167] | 77.0 | 82.0 | 40.0 | 81.0 | 88.0 | 64.0 | 79.0 | 80.0 | 76.0 | 81.0 | 67.0 | 97.0 | 68.0 | 78.0 | 74.0 | 90.0 | 88.0 |
| Cylinder3D [20] | 77.2 | 82.8 | 29.8 | 84.3 | 89.4 | 63.0 | 79.3 | 77.2 | 73.4 | 84.6 | 69.1 | 97.7 | 70.2 | 80.3 | 75.5 | 90.4 | 87.6 |
| AMVNet [153] | 77.3 | 80.6 | 32.0 | 81.7 | 88.9 | 67.1 | 84.3 | 76.1 | 73.5 | 84.9 | 67.3 | 97.5 | 67.4 | 79.4 | 75.5 | 91.5 | 88.7 |
| SPVCNN [154] | 77.4 | 80.0 | 30.0 | 91.9 | 90.8 | 64.7 | 79.0 | 75.6 | 70.9 | 81.0 | 74.6 | 97.4 | 69.2 | 80.0 | 76.1 | 89.3 | 87.1 |
| AF2S3Net [158] | 78.3 | 78.9 | 52.2 | 89.9 | 84.2 | 77.4 | 74.3 | 77.3 | 72.0 | 83.9 | 73.8 | 97.1 | 66.5 | 77.5 | 74.0 | 87.7 | 86.8 |
| 2D3DNet [159] | 80.0 | 83.0 | 59.4 | 88.0 | 85.1 | 63.7 | 84.4 | 82.0 | 76.0 | 84.8 | 71.9 | 96.9 | 67.4 | 79.8 | 76.0 | 89.2 | 89.2 |
| GASN [165] | 80.4 | 85.5 | 43.2 | 90.5 | 92.1 | 64.7 | 86.0 | 83.0 | 73.3 | 83.9 | 75.8 | 97.0 | 71.0 | 81.0 | 77.7 | 91.6 | 90.2 |
| 2DPASS [164] | 80.8 | 81.7 | 55.3 | 92.0 | 91.8 | 73.3 | 86.5 | 78.5 | 72.5 | 84.7 | 75.5 | 97.6 | 69.1 | 79.9 | 75.5 | 90.2 | 88.0 |
| LidarMultiNet [166] | 81.4 | 80.4 | 48.4 | 94.3 | 90.0 | 71.5 | 87.2 | 85.2 | 80.4 | 86.9 | 74.8 | 97.8 | 67.3 | 80.7 | 76.5 | 92.1 | 89.6 |
| UniSeg [163] | 83.5 | 85.9 | 71.2 | 92.1 | 91.6 | 80.5 | 88.0 | 80.9 | 76.0 | 86.3 | 76.7 | 97.7 | 71.8 | 80.7 | 76.7 | 91.3 | 88.8 |
| **RAPiD-Seg (Ours)** | 83.6 | 84.8 | 64.3 | 95.0 | 92.2 | 84.6 | 87.9 | 81.8 | 76.8 | 88.5 | 79.0 | 97.8 | 66.6 | 81.2 | 76.7 | 92.5 | 88.4 |

**Table 5.3: Cross-Dataset evaluation** on SemanticKITTI *validation* set. "SemK" and "nuS" refer to the SemanticKITTI and nuScenes datasets, which are used during the AE training stages of the configurations (Conf.); Beams $a \rightarrow b$ represents downsample to $a$ beams for AE training stage, and $b$ beams (without sampling) for fine-tuning and validation.

| Conf. | AE Train | Finetune | Beams | Test | mIoU |
|---|---|---|---|---|---|
| (a.1) | SemK | SemK | $64 \rightarrow 64$ | SemK | **73.02** |
| (a.2) | SemK | SemK | $32 \rightarrow 64$ | SemK | 72.60 |
| (a.3) | SemK + nuS | SemK | $32 \rightarrow 64$ | SemK | 72.78 |
| (b.1) | nuS | nuS | $32 \rightarrow 32$ | nuS | 79.91 |
| (b.2) | SemK + nuS | nuS | $32 \rightarrow 32$ | nuS | **80.56** |

**Qualitative Results**

We present supporting multi-frame qualitative visualizations in Figure 5.8. Whereas the baseline method struggles with accurate vehicle type differentiation, ours achieves consistent segmentation. We further provide more qualitative results, with the segmentation error map (Figures 5.9 and 5.10) and magnification of regional details (Figure 5.11) as follows.

**Segmentation Error Map:** We present qualitative comparisons with PCSeg [163] and ground truth through error maps in Figure 5.9 (SemanticKITTI) and Figure 5.10 (nuScenes). The visualization underscores the superior performance of our method, marked by significantly reduced segmentation errors in each analyzed frame.

**Magnification of Regional Details:** We present a visualization of magnified regional details in Figure 5.11 to showcase segmentation details and performance at a long range. The magnified details indicate that our method performs well in segmenting various classes such as other grounds, parking areas, sidewalks, vegetation, *etc.*

**RAPiD Embedding Visualizations:** We present the visualization of the RAPiD embeddings in Figure 5.12. Specifically, we initially train the RAE, following which the pointwise RAPiD features are processed through the RAE to yield the RAPiD embeddings.

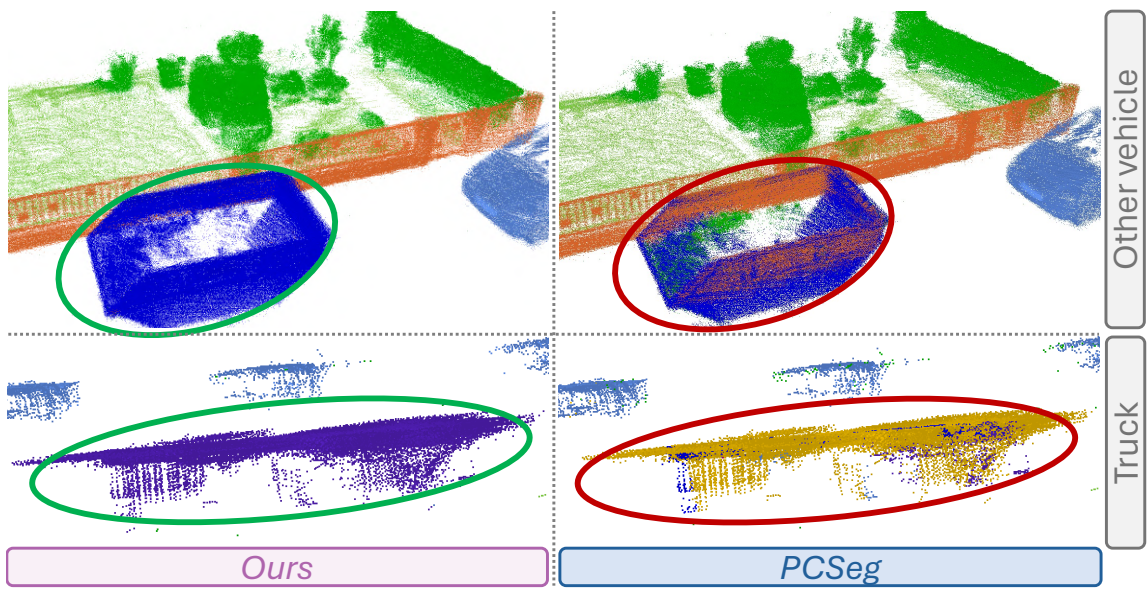

**Figure 5.8: Comparing our results and PCSeg (baseline)** under multi-scan visualization, showing improved segmentation results.

Given that these RAPiD embeddings belong to a high-dimensional space (exceeding three dimensions), we employ Principal Component Analysis (PCA, [7]) to reduce the dimensionality of the RAPiD embeddings to a 3D representation, thereby facilitating visualization. The results show that our RAPiD features are stable and distinctive among semantic categories. Moreover, embeddings of the same surface material/semantic category/object exhibit consistency across different viewpoints or ranges. This further enhances the performance of our semantic segmentation network RAPiD-Seg, which integrates RAPiD features, achieving commendable segmentation results.

### 5.5.3 Ablation Studies

We conduct ablation studies to validate the effectiveness of proposed components, including RAPiD and reflectivity features, RAPiD AE and architecture variants, feature fusion with channel-wise attention, and different backbone networks.

#### Effectiveness of Components

In Table 5.4, we ablate each component of RAPiD-Seg step by step and report the performance on the SemanticKITTI *validation* set. We start from a baseline model, achieving an mIoU of 70.04 on *validation* set. Incorporating the RAPiD features leads to a notable increase in mIoU (+1.17), underscoring the efficacy of RAPiD in enhancing segmentation performance. Building upon the RAPiD framework, the integration of reflectivity further elevates our mIoU to 71.93, with an overall improvement of +1.89 from the baseline. This significant gain illustrates the critical role of reflectivity in capturing different object materials. Utilizing RAPiD AE to get the voxel-wise representation, we observe an additional improvement, taking the mIoU to 72.15 (+2.11 compared to the baseline), which demonstrates its capability in processing feature representations. We subsequently performed a module-by-module reduction to analyze the model performance. Our ablation studies show that while disabling individual components leads to some improvement over the baseline, the peak performance is only attained when all components function together. This highlights the indispensable contribution of each component in maximizing segmentation accuracy.

**Table 5.4: Component-wise ablation** of RAPiD-Seg on the SemanticKITTI *validation* set.

| RAPiD Features | | | Attention | mIoU | Δ |
|---|---|---|---|---|---|
| Geometric | Reflectivity | Embedding | | | |
| | | | | 70.04 | (Baseline) |
| | | | ✓ | 70.46 | (+0.42) |
| ✓ | | | | 71.21 | (+1.17) |
| ✓ | ✓ | | | 71.93 | (+1.89) |
| ✓ | ✓ | ✓ | | 72.15 | (+2.11) |
| | ✓ | ✓ | ✓ | 71.80 | (+1.76) |
| ✓ | | ✓ | ✓ | 72.32 | (+2.28) |
| ✓ | ✓ | | ✓ | 72.78 | (+2.74) |
| ✓ | ✓ | ✓ | ✓ | **73.02** | (+2.98) |

**Table 5.5: Effects of RAPiD and Reflectivity features** compared to other configurations on SemanticKITTI *validation* set. PDD: the compressed PDD embeddings; RAPiD-R: only RAPiD, without reflectivity feature fusion; RAPiD+R: both RAPiD and reflectivity.

| Conf. | mIoU | truck | o.veh | park | walk | o.gro | build | fence |
|---|---|---|---|---|---|---|---|---|
| BaseL [6] | 70.0 | 59.8 | <u>70.3</u> | 69.2 | 76.9 | <u>36.2</u> | <u>93.7</u> | 69.6 |
| PDD [35] | 66.2 | 40.3 | 65.8 | 67.5 | 74.6 | 33.1 | 92.8 | 68.2 |
| RAPiD-R | <u>71.8</u> | <u>62.5</u> | 69.0 | <u>70.3</u> | <u>77.4</u> | 35.8 | 93.6 | <u>70.8</u> |
| RAPiD+R | **73.0** | **70.4** | **78.5** | **75.8** | **78.9** | **44.2** | **94.2** | **73.1** |

**Table 5.6: Effects of various $k$ at different ranges.**

| | | PDD [34] | | | | RAPiD (ours) | | |
|---|---|---|---|---|---|---|---|---|
| | $k_{\text{near}}$ | $k_{\text{mid}}$ | $k_{\text{far}}$ | mIoU | $k_{\text{near}}$ | $k_{\text{mid}}$ | $k_{\text{far}}$ | mIoU |
| **SemK** | 7 | 7 | 7 | 64.74 (-7.1) | 7 | 7 | 7 | 72.04 (+0.2) |
| | 5 | 5 | 5 | 65.18 (-6.6) | 5 | 5 | 5 | 72.28 (+0.5) |
| | 10 | 7 | 5 | 66.23 (-5.6) | **10** | **7** | **5** | **73.02** (+1.2) |
| **nuScene** | 6 | 6 | 6 | 72.19 (-6.5) | 6 | 6 | 6 | 78.76 (+0.1) |
| | 3 | 3 | 3 | 73.68 (-5.0) | 3 | 3 | 3 | 79.43 (+0.8) |
| | 8 | 6 | 3 | 72.24 (-6.4) | **8** | **6** | **3** | **79.91** (+1.3) |

**Table 5.7: 3D segmentation results** of different variants of RAPiD-Seg (ours) on SemanticKITTI *validation* set.

| Method | mIoU % | car | ped | o.gro | pole |
|---|---|---|---|---|---|
| Baseline | 70.0 | 97.2 | 78.1 | 35.4 | 63.5 |
| R-RAPiD-Seg | 72.3 (+2.3) | 97.4 | 77.4 | **45.0** | 62.4 |
| C-RAPiD-Seg | **73.0** (+3.0) | **97.7** | **79.3** | 44.6 | **66.4** |

**Effectiveness of RAPiD and Reflectivity Features**

In Table 5.5, we validate the efficacy of the proposed RAPiD features. The direct application of PDD results in a substantial mIoU decrease of 3.8, with most category IoUs falling below those of the baseline method. This indicates vanilla PDD features are not well-suited for LiDAR segmentation. While our proposed RAPiD (without reflectivity disparities in Equation (5.2)) demonstrates an improvement of 1.4 in mIoU over the baseline. Moreover, the integration of reflectivity with RAPiD features significantly enhances performance, yielding 2.6 mIoU increase over the baseline. Notably, our approach exhibits superior performance in the segmentation of most rigid object categories compared to other configurations.

Table 5.6 further shows the RAPiD outperforms PDD in varying $k$ values at different ranges on SemanticKITTI *validation* set, with non-uniform $k$ configurations yielding the most significant improvements in mIoU for both datasets, highlighting the effectiveness of the range-aware design of RAPiD.

**Table 5.8: Effects of using different backbones** on SemanticKITTI *validation* set, where P and V for Point- and Voxel-wsie methods.

| Repr. | Backbone | mIoU % | car | ped | o.gro | pole |
|-------|----------|--------|-----|-----|-------|------|
| P | PTv2 [73] | 72.6 (+2.8) | **97.4** | 77.4 | **45.0** | 62.4 |
| V | Cylinder3D [20] | 69.8 | 96.9 | 74.2 | 37.9 | 63.0 |
| V | Minkowski-UNet [48] | **73.0** (+3.2) | 97.2 | **78.1** | 43.3 | **65.9** |

**Effectiveness of RAPiD AE and Architecture Variants**

In Table 5.4, we replace our RAPiD AE with a commonly-used ConvFNN [19, 20, 38] to generate voxel-wise embeddings. Our AE demonstrated a 0.24 mIoU improvement over the FNN. In Table 5.7, we investigate the impact of the variant architectures, *i.e.*, R-RAPiD-Seg and C-RAPiD-Seg. Utilizing R-RAPiD features, R-RAPiD-Seg yields an mIoU of 72.3, which outperforms the baseline method by +2.3 mIoU. Concurrent fusing of both R- and C-RAPiD-Seg features in R-RAPiD-Seg improves performance by +0.7 mIoU, which shows a +3.0 overall mIoU enhancement.

**Effectiveness of Feature Fusion with Channel-wise Attention**

In Table 5.4, we assess the prevalent approach of direct feature concatenation. Our method, employing feature fusion with channel-wise attention, enhances mIoU by 0.87, thereby conclusively demonstrating its effectiveness.

**Effectiveness of Backbone Networks**

In Table 5.8, we evaluate the performance across multiple backbone networks to demonstrate the versatility and ease of integration of our modules. The results highlight the adaptability and effectiveness of our approach on both Pointwise (P) and Voxel-wise (V) backbone architectures. Specifically, the point-based PTv2 backbone achieves an mIoU of 72.6, showing strong performance, particularly in the `car` and `other ground` categories. Among the voxel-based methods, Minkowski-UNet stands out with the highest overall mIoU of 73.0, excelling in the `pedestrian` and `pole` categories.

# 5.6 Summary

We present a novel **R**ange-**A**ware **P**o**i**ntwise **D**istance Distribution (RAPiD) feature and the RAPiD-Seg network for LiDAR segmentation, adeptly overcoming the constraints of single-modal LiDAR methods. The rigid transformation invariance and enhanced focus on local details of RAPiD significantly boost segmentation accuracy. RAPiD-Seg integrates a two-stage training approach with reflectivity-guided 4D distance metrics and a class-aware nested AE, achieving SOTA results on the SemanticKITTI and nuScenes datasets. Notably, our single-modal method surpasses the performance of multi-modal methods, indicating the superior efficacy of RAPiD features even compared to other modalities, including RGB and range images.

Our RAPiD features hold significant potential for application in various LiDAR-driven tasks such as object detection and point cloud registration, and extend to promising applications in multi-modal research in the future.

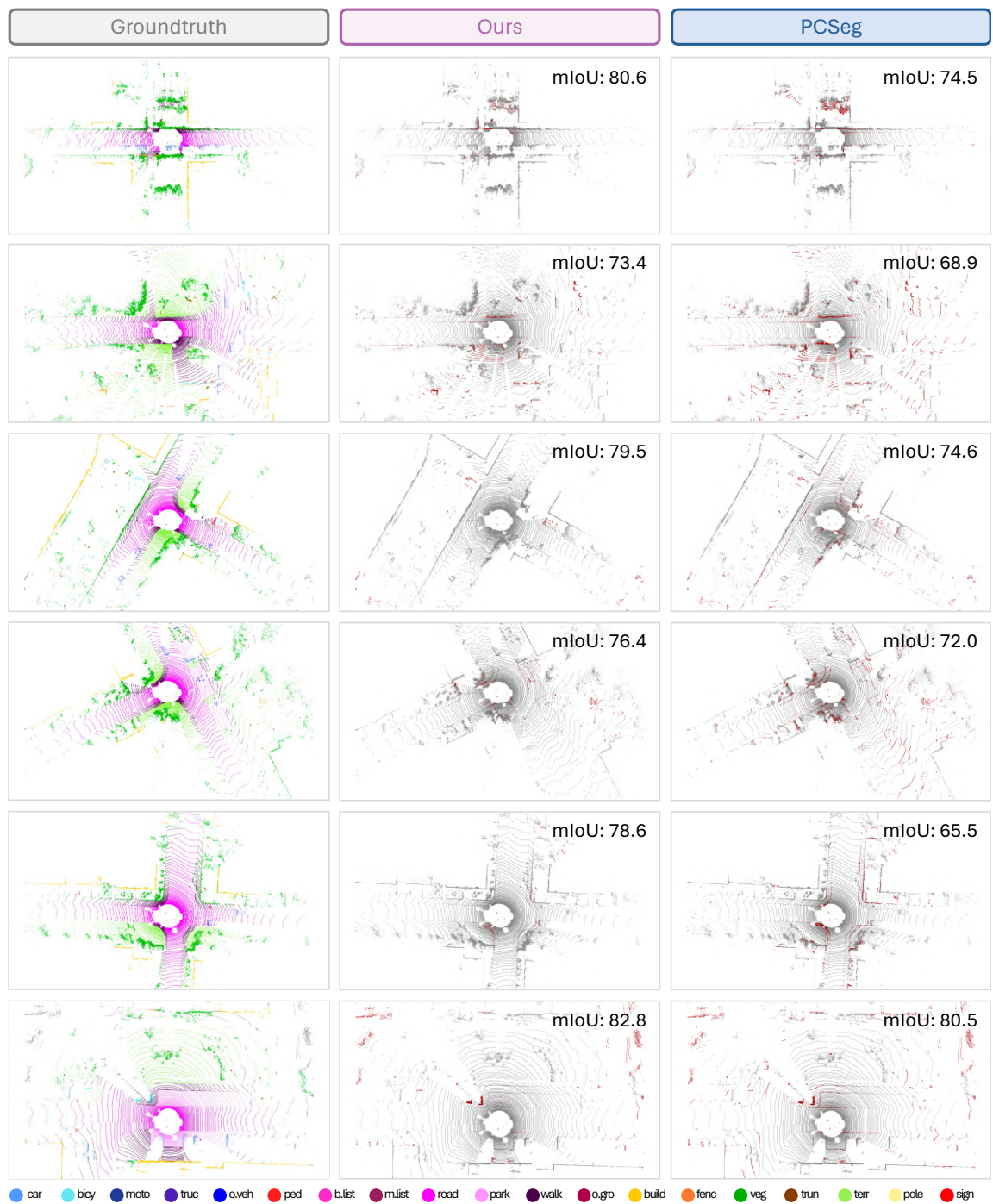

**Figure 5.9: Qualitative comparisons with PCSeg** [6] and groundtruth through error maps on SemanticKITTI [3] *validation* set. To highlight the differences, the **correct** / **incorrect** predictions are painted in **gray** / **dark red**, respectively. Each scene is visualized from the ego-vehicle LiDAR bird's eye view (BEV) and covers a region of 50m by 30m. Best viewed in color.

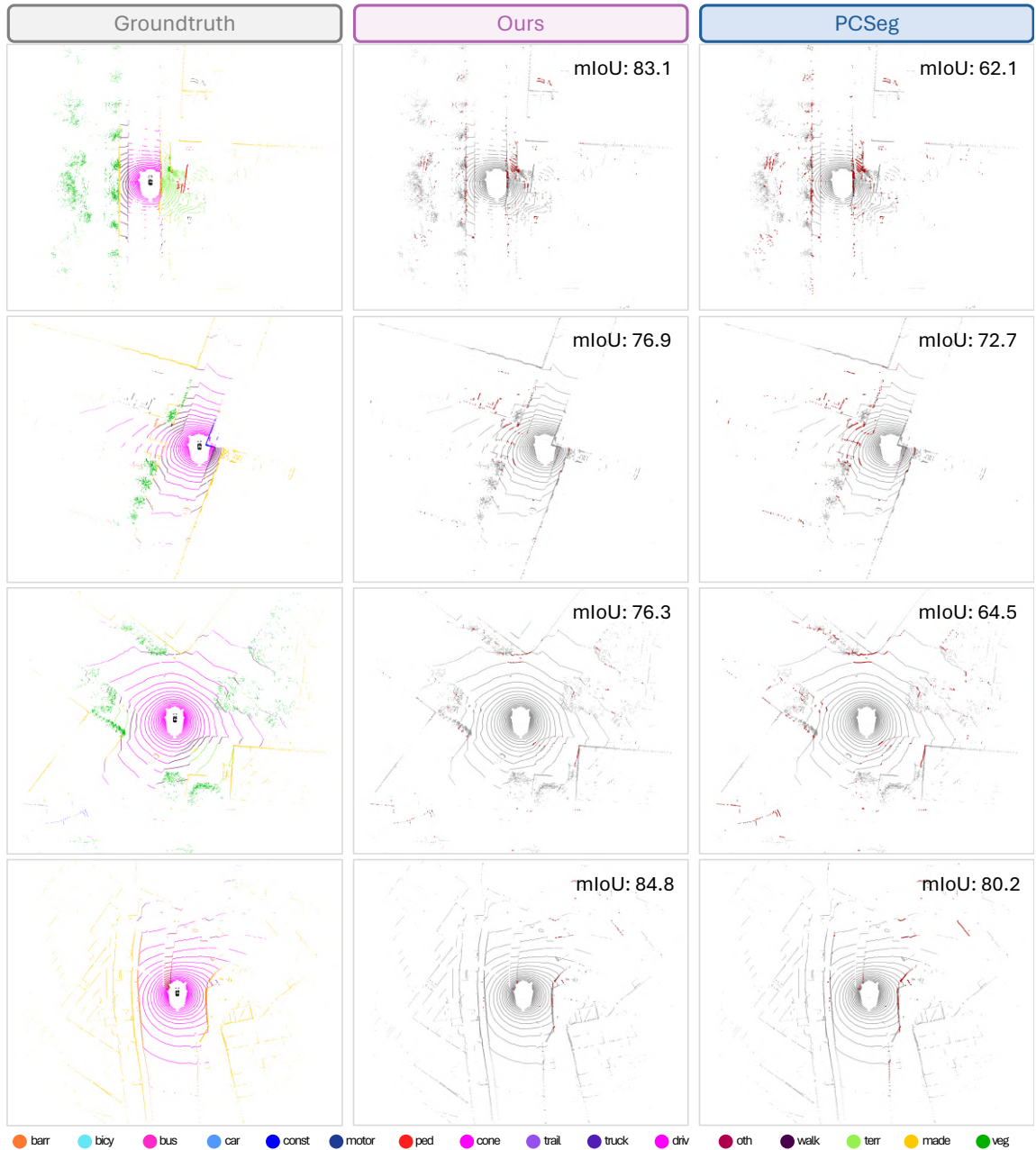

**Figure 5.10: Qualitative comparisons with PCSeg** [6] and groundtruth through error maps on nuScenes [5] *validation* set. To highlight the differences, the **correct** / **incorrect** predictions are painted in **gray** / **dark red**, respectively. Each scene is visualized from the ego-vehicle LiDAR bird's eye view (BEV) and covers a region of 50m by 40m. Best viewed in color.

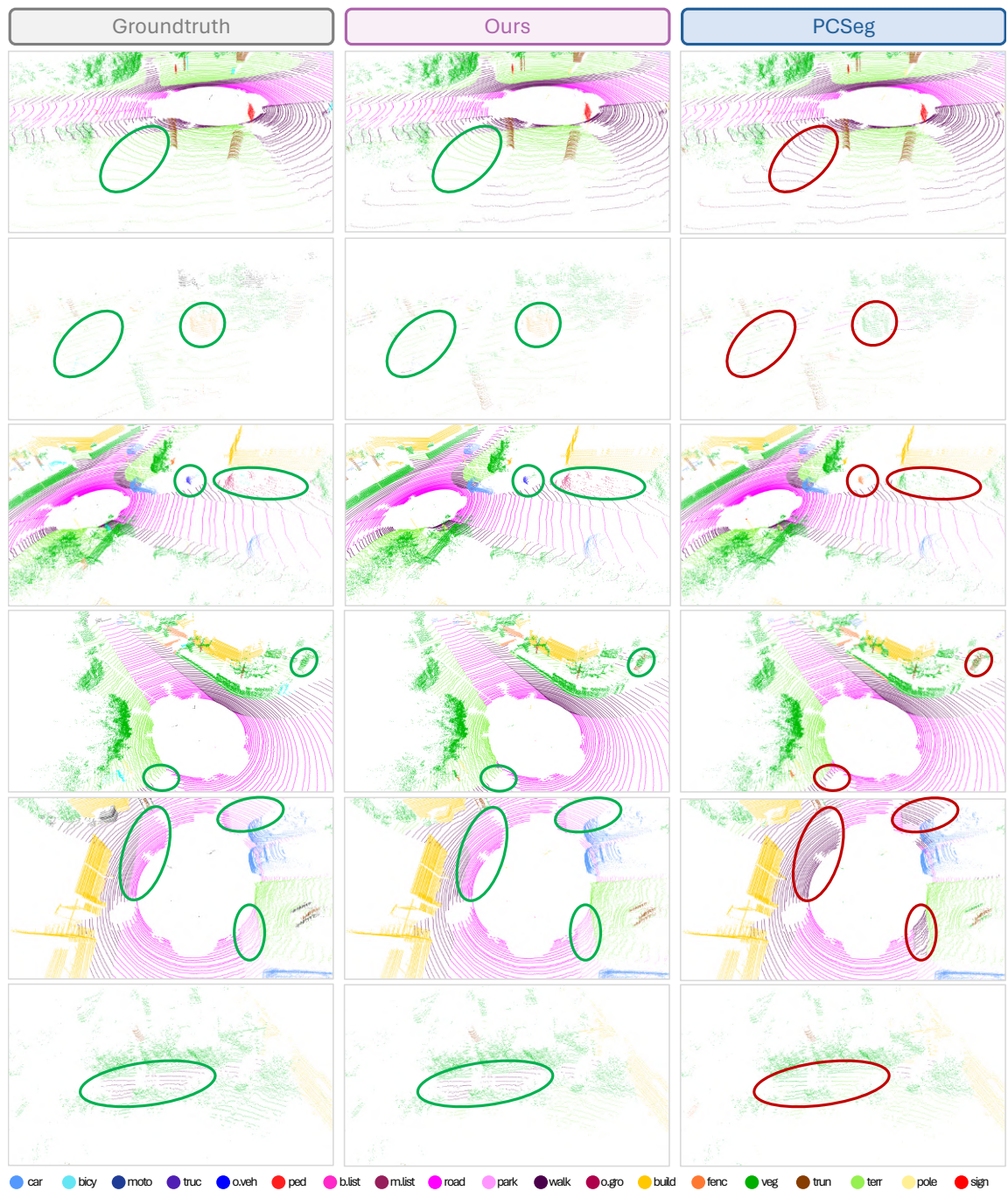

**Figure 5.11: Magnification of regional details**: comparing with PCSeg [6] and groundtruth on SemanticKITTI [3] *validation* set. To highlight the differences, areas of improvement are highlighted in green, and areas of underperformance in red. Best viewed in color.

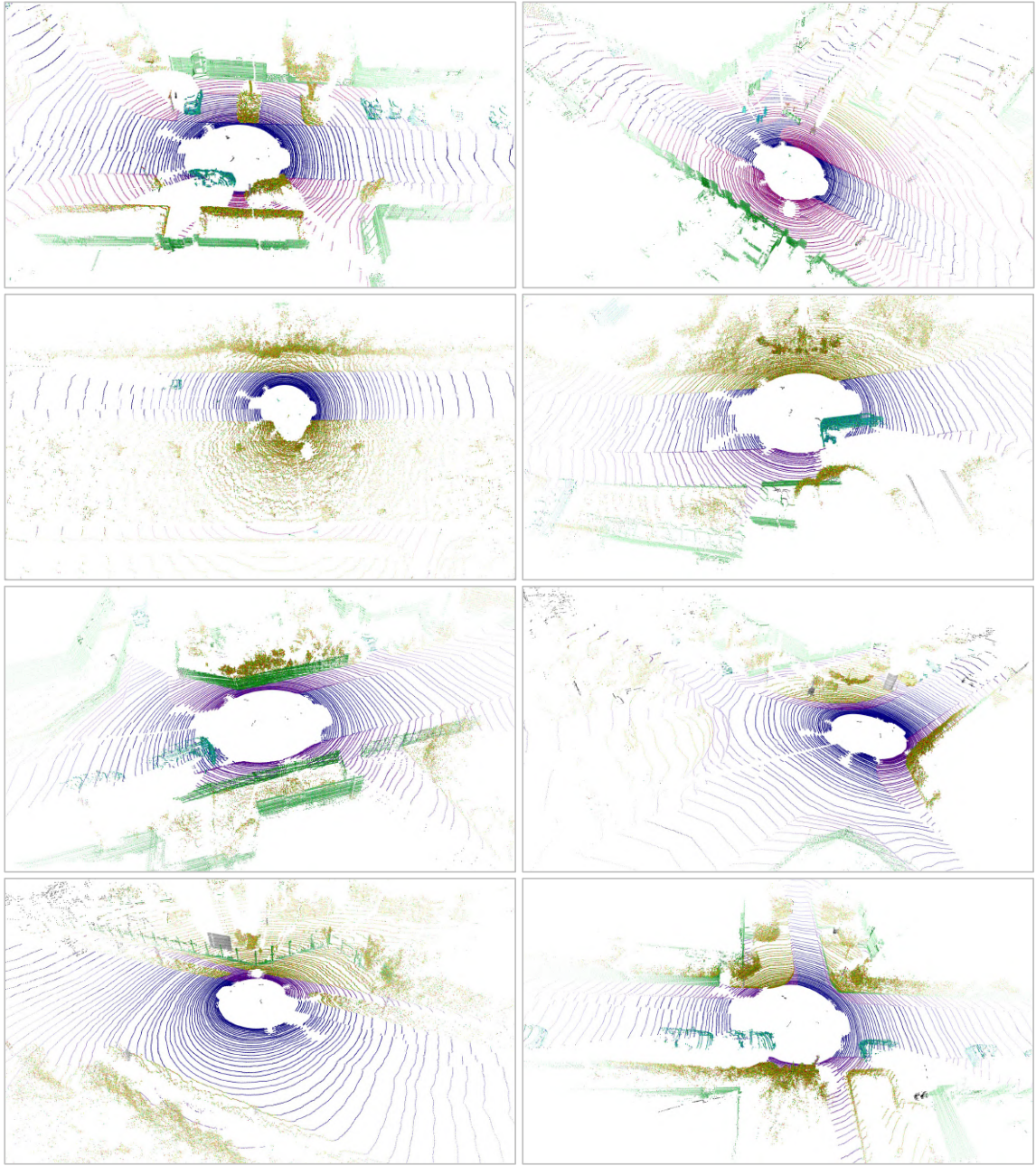

**Figure 5.12: Our method learns a high-dimensional RAPiD latent representation** for capturing the localized geometric structure of neighboring points. We apply PCA [7] to reduce the latent dimension to 3 and plot as RGB. Different colors represent various RAPiD 3D representations. Best viewed in color.

CHAPTER 6

---

Conclusion

---

In the domain of autonomous driving, the integration of advanced sensor technologies and computational methodologies stands as a pivotal factor for the enhancement of vehicle automation and safety. The contributions made during this doctoral research encompass the evaluation and analysis of recent SOTA monocular depth estimation methods (Chapter 3 and Section 3.6), innovative deep learning architecture (Chapter 4 and Chapter 5), and high-fidelity datasets (Chapter 3), substantially advancing the capability of autonomous systems to perceive and interpret their surroundings with greater depth and accuracy.

In Chapter 3, we introduce a high-fidelity 128-channel 3D LiDAR dataset equipped with panoramic ambient (near infrared) and reflectivity imagery, tailored for autonomous driving applications. This dataset marks a significant advancement in the field due to its unprecedented LiDAR vertical resolution and the inclusion of ambient and reflectivity imagery, offering a more detailed representation of the environment for depth estimation and other autonomous driving tasks. The monocular depth estimation task on this benchmark dataset further demonstrates an enhancement in performance across various state-of-the-art approaches, thereby emphasizing the importance of high-resolution LiDAR data in augmenting supervised learning models within the autonomous driving domain.

In Chapter 4, an efficient semi-supervised architecture for 3D point cloud semantic segmentation is proposed, encapsulating the principle of "less is more". Through innovative computational modules and strategies, this architecture delivers superior performance while reducing computational costs and the necessity for extensive annotations and model parameters. This work addresses the challenges associated with model complexity and computational efficiency, effectively bridging the performance gap between semi-supervised and fully-supervised learning approaches in the realm of 3D semantic segmentation tasks.

In Chapter 5, we present a novel LiDAR segmentation approach through Range-Aware Pointwise Distance Distribution (RAPiD) features and the associated RAPiD-Seg network. This method notably improves segmentation accuracy by concentrating on local details and surpassing the limitations of single-modal LiDAR techniques. Demonstrating superior performance over multi-modal approaches in benchmark datasets underscores the potential of RAPiD features not only in LiDAR segmentation but also across a broader spectrum of LiDAR-driven and multi-modal tasks.

These contributions underscore the efforts made in pushing the frontiers of autonomous driving research. By introducing LiDAR datasets of higher fidelity (resolution), deep learning architectures that optimize efficiency and accuracy, and innovative features for better and robust LiDAR segmentation, this research lays the groundwork for future advancements in vehicle autonomy.

## 6.1 Review of Contributions

In this thesis, we address the critical need to enhance both the **accuracy** and **efficiency** of 3D LiDAR-based applications for autonomous vehicles, focusing on geometric and semantic scene understanding. Our research makes multiple significant contributions toward these dual objectives:

In terms of the **accuracy**, we propose DurLAR [17], a novel high-fidelity 128-channel 3D LiDAR dataset that includes panoramic ambient and reflectivity imagery. This dataset also sets a new benchmark for monocular depth estimation, demonstrating that higher resolution and enhanced data availability significantly improve depth estimation accuracy.

We also propose the RAPiD-Seg architecture [229], leveraging Range-Aware Pointwise Distance Distribution (RAPiD) features to improve 3D LiDAR segmentation accuracy. This architecture achieves state-of-the-art results by providing isometry-invariant properties in LiDAR point cloud that enhance local representation and overall segmentation accuracy.

In terms of the **efficiency**, we introduce a semi-supervised methodology [16] for 3D LiDAR semantic segmentation that achieves superior accuracy while significantly reducing computational costs. This is achieved through the Sparse Depthwise Separable Convolution (SDSC) module and Spatio-Temporal Redundant Frame Downsampling (ST-RFD), which reduces model complexity and the need for extensive labeled data, thus enhancing computational efficiency.

Collectively, these contributions advance the state-of-the-art in both accuracy and efficiency of on-vehicle 3D LiDAR systems, supporting the development of safer and more reliable autonomous driving technologies.

## 6.2 Future Research Directions

The methodologies proposed in Chapters 3 to 5, while demonstrating strong performance, possess certain limitations. In this section, we critically analyze these limitations and propose potential directions for future research to address these challenges.

### 6.2.1 Dataset Ground Truth Availability

One limitation is the absence of ground truth in our DurLAR dataset (Chapter 3) for various common autonomous driving tasks, such as semantic segmentation [3, 16, 19, 230] and object detection [21, 22]. This limitation is primarily due to the substantial annotation costs associated with manual annotation [15, 16].

To address this limitation, future research directions should focus on leveraging pretrained models to perform initial automated annotations, thereby generating *pseudo-labels* [16, 229]. These pseudo-labels can subsequently be refined through manual correction to produce accurate ground truth. Providing ground truth for common autonomous driving tasks within the DurLAR [17] dataset has several significant advantages.

Firstly, annotating ground truth for the DurLAR dataset will significantly enhance its utility for a broader range of autonomous driving tasks [21, 230], facilitating more comprehensive research and development and evament. With high-quality ground truth and high-fidelity LiDAR point cloud, deep learning models trained on the DurLAR dataset can achieve improved performance in tasks such as semantic segmentation and object detection, leading to more reliable and robust autonomous driving systems. Moreover, the combination of automated pseudo-labeling and manual correction strikes a balance between annotation accuracy and resource expenditure, making the process more efficient and scalable. Finally, providing ground truth annotations will enable the creation of benchmarks for various tasks, fostering comparative studies and driving advancements in the field.

By addressing these limitations through strategic future research, we can enhance the value of the DurLAR dataset and contribute to the development and evament of more advanced and reliable autonomous driving technologies.

### 6.2.2 Adaptive ST-RFD Strategies

Spatio-Temporal Redundant Frame Downsampling (ST-RFD) strategy (Section 4.5), while effective in minimizing training set size and redundancy, depends heavily on the accurate estimation of temporal correlation between frames [16]. This dependency might limit its applicability in scenarios where the motion patterns of the LiDAR sensor are highly irregular or unpredictable. Furthermore, the ST-RFD strategy requires an empirical supervisor function to determine the sampling rate, introducing an additional layer of complexity that may not be optimal across all use cases.

Developing more adaptive and intelligent sampling strategies [231] that can dynamically adjust to different motion patterns and environmental contexts without relying on pre-defined empirical functions. This could involve machine learning models that predict optimal sampling rates based on real-time analysis of the LiDAR data stream.

### 6.2.3 Extension of RAPiD Features

Despite the significant advancements made by RAPiD-Seg (Chapter 5) in the domain of LiDAR-based 3D semantic segmentation, several limitations remain. Firstly, the computational complexity associated with RAPiD features poses a main challenge. The high-dimensional nature of these features [34, 35], while essential for capturing detailed local geometries, results in computational and memory requirements [171]. This can render the approach impractical for huge-scale point clouds [36, 232] due to the significant storage needs and computational inefficiency involved [171].

Moreover, although RAPiD features demonstrate robustness [229], they are still susceptible to noise prevalent in outdoor environments, particularly in scenarios with sparse or irregular point clouds. This susceptibility can lead to segmentation inaccuracies, especially in long-range observations where data sparsity is a common issue. Furthermore, while RAPiD-Seg has achieved state-of-the-art performance on SemanticKITTI [3] and nuScenes [5] datasets, its ability to generalize across other datasets [37] and varied environmental conditions remains to be thoroughly evaluated. Ensuring consistent performance in diverse contexts with different sensor setups and environmental conditions is an open challenge that needs further investigation.

To address these limitations, multiple future research directions can be explored. One promising avenue is the extension of RAPiD-Seg [229] to semi-supervised [16] and weakly-supervised [19] learning paradigms. Investigating how RAPiD-Seg performs under these conditions could leverage partially labeled data to enhance the model generalization capabilities, which is crucial for practical applications where annotated data is limited [16, 18, 19]. Additionally, a comprehensive comparison of RAPiD features with other LiDAR representations [42, 170, 233] in tasks such as domain adaptation and out-of-distribution scenarios could provide further validation of the robustness of RAPiD features learning capabilities. Such comparisons would elucidate the strengths and weaknesses of RAPiD features relative to other approaches, thereby enhancing the adaptability and resilience of the model to varying data distributions.

Furthermore, integrating RAPiD features into multi-modal systems [163] that utilize data from different sensors, such as cameras and radar, could offer a more holistic scene understanding. This integration would leverage the complementary strengths of various sensors [17, 37], thereby improving the overall perception performance in autonomous systems. Another critical area for future research is optimizing RAPiD features for real-time applications. Reducing the computational burden [16, 171] of RAPiD features through the development of more efficient algorithms and leveraging hardware acceleration techniques is essential to make RAPiD-Seg viable for real-time scenarios such as autonomous driving and robotic navigation.

Lastly, enhancing the robustness of RAPiD features to environmental variability, including changes in lighting, weather conditions, and dynamic obstacles, is paramount. Advanced data augmentation techniques [19, 234] and adaptive learning mechanisms could be employed to maintain high performance under diverse conditions. Addressing these research directions will further unlock the potential of RAPiD-Seg in various LiDAR-driven tasks, contributing to the development of more robust and versatile autonomous systems.

# Content Acknowledgements

In addition to those individuals outlined in the Acknowledgments of this thesis, specific assistance in terms of hardware, software and suggestions have been received as acknowledged below.

We acknowledge Durham NVIDIA CUDA Center (NCC) GPU system [1]. The research and experiments in this thesis has used Durham University's NCC cluster. NCC has been purchased through Durham University's strategic investment funds, and is installed and maintained by the Department of Computer Science.

We acknowledge Bede [2], a supercomputer (otherwise known as an HPC system) run by the N8 group of research intensive universities in the north of England, on behalf of EPSRC, and hosted at Durham University. The research and experiments in this thesis make use of the facilities of the N8 Centre of Excellence in Computationally Intensive Research (N8 CIR) provided and funded by the N8 research partnership and EPSRC (Grant No. EP/T022167/1).

---

[1] https://nccadmin.webspace.durham.ac.uk
[2] https://n8cir.org.uk/bede

## A.1  DurLAR: A High-Fidelity LiDAR Dataset

We acknowledge the Electronics Technician & Electronics Workshop, Department of Engineering for the installation and debugging of part of the hardware and power supply for the Twizy Vehicle. We thank Mr. Richard Marcus for the recent calibration method [214].

In addition, we also acknowledge the use of the following public resources, during the course of Chapter 3:

- ManyDepth[3] ................................. Copyright © Niantic, Inc. 2021.

- BerHu Loss[4] ........................................... CC BY-NC-SA 4.0

- ouster_example[5] ........................................... BSD-3-Clause

- multisense_ros[6] ................... Copyright © Carnegie Robotics, LLC. 2014.

- oxford_gps_eth[7] ........................................... BSD-2-Clause

- yoctolib_python[8] .......................... Copyright © Yoctopuce Sarl. 2011.

## A.2  Efficient 3D LiDAR Semantic Segmentation

We acknowledge Prof. Tobias Weinzierl in Durham University for his suggestions and discussions on various methods for implementing efficient hardware acceleration.

In addition, we also acknowledge the use of the following public resources, during the course of Chapter 4:

- ScribbleKITTI[9] ......................................... CC BY-NC-SA 4.0

- U2PL[10] ............................................... Apache License 2.0

---

[3] https://github.com/nianticlabs/manydepth.
[4] http://ljk.imag.fr/membres/Laurent.Zwald/paper/CodesOnLineBerhuPaper.tar.gz.
[5] https://github.com/ouster-lidar/ouster_example.
[6] https://github.com/carnegierobotics/multisense_ros.
[7] https://github.com/mit-drl/oxford_gps_eth.
[8] https://github.com/yoctopuce/yoctolib_python.
[9] https://github.com/ouenal/scribblekitti.
[10] https://github.com/Haochen-Wang409/U2PL/.

- SemanticKITTI[11] ............................................. CC BY-NC-SA 4.0

- SemanticKITTI-API[12] ............................................. MIT License

- Cylinder3D[13] ............................................. Apache License 2.0

- SpConv[14] ............................................. Apache License 2.0

- PyTorch-Lightning[15] ..................................... Apache License 2.0

## A.3   Accurate 3D LiDAR Semantic Segmentation

We ackowledge Prof. Vitaliy Kurlin in University of Liverpool for his discussion in Pointwise Distance Distribution (PDD).

In addition, we also acknowledge the use of the following public resources, during the course of Chapter 5:

- nuScenes[16] ............................................. CC BY-NC-SA 4.0

- nuScenes-devkit[17] ....................................... Apache License 2.0

- SemanticKITTI[18] ............................................. CC BY-NC-SA 4.0

- SemanticKITTI-API[19] ............................................. MIT License

- PCSeg[20] ............................................. Apache License 2.0

- Pointcept[21] ............................................. MIT License

- MinkowskiEngine[22] ............................................. MIT License

---

[11] http://semantic-kitti.org.
[12] https://github.com/PRBonn/semantic-kitti-api.
[13] https://github.com/xinge008/Cylinder3D.
[14] https://github.com/traveller59/spconv.
[15] https://github.com/Lightning-AI/lightning.
[16] https://www.nuscenes.org/nuscenes.
[17] https://github.com/nutonomy/nuscenes-devkit.
[18] http://semantic-kitti.org.
[19] https://github.com/PRBonn/semantic-kitti-api.
[20] https://github.com/PJLab-ADG/PCSeg.
[21] https://github.com/Pointcept/Pointcept.
[22] https://github.com/NVIDIA/MinkowskiEngine.

- Cylinder3D[23] .............................................. Apache License 2.0

- VoxSeT[24] ...................................................... MIT License

- LiM3D[25] ................................................... Apache License 2.0

- SpConv[26] .................................................. Apache License 2.0

- Average-Minimum-Distance[27] ............................. CC BY-NC-SA 4.0

- PyTorch-Lightning[28] ...................................... Apache License 2.0

APPENDIX B

---

# Public Access for DurLAR Dataset

---

Our DurLAR dataset is open-accessed to the public, which is hosted on Durham Collections. In this chapter, we provide details for accessing the DurLAR dataset [17], as well as descriptions of the data, related tools, and scripts.

## B.1 Data Structure

In DurLAR dataset, each drive folder contains 8 topic folders for every frame,

- `ambient/`: panoramic ambient imagery

- `reflec/`: panoramic reflectivity imagery

- `image_01/`: right camera (grayscale+synced+rectified)

- `image_02/`: left RGB camera (synced+rectified)

- `ouster_points/`: ouster LiDAR point cloud (KITTI-compatible binary format)

- `gps, imu, lux`: csv file format

The folder structure of the DurLAR dataset is shown in Figure B.1. The folder structure of the DurLAR calibration information (both internal and external calibration) is shown in Figure B.2.

```
DurLAR_<date>/
├── ambient/
│   ├── data/
│   │   └── <frame_number.png>    [ ..... ]
│   └── timestamp.txt
├── gps/
│   └── data.csv
├── image_01/
│   ├── data/
│   │   └── <frame_number.png>    [ ..... ]
│   └── timestamp.txt
├── image_02/
│   ├── data/
│   │   └── <frame_number.png>    [ ..... ]
│   └── timestamp.txt
├── imu/
│   └── data.csv
├── lux/
│   └── data.csv
├── ouster_points/
│   ├── data/
│   │   └── <frame_number.bin>    [ ..... ]
│   └── timestamp.txt
├── reflec/
│   ├── data/
│   │   └── <frame_number.png>    [ ..... ]
│   └── timestamp.txt
└── readme.md                     [ README file ]
```

**Figure B.1: The folder structure** of the DurLAR dataset.

```
DurLAR_calibs/
├── calib_cam_to_cam.txt          [ Camera to camera calibration results ]
├── calib_imu_to_lidar.txt        [ IMU to LiDAR calibration results ]
└── calib_lidar_to_cam.txt        [ LiDAR to camera calibration results ]
```

**Figure B.2: The folder structure** of the DurLAR calibration information.

## B.2   Download the Dataset

Access to the complete DurLAR dataset can be requested through the following link: https://forms.gle/ZjSs3PWeGjjnXmwg9. Upon completion of the form, the download script `durlar_download` and accompanying instructions will be automatically provided. The DurLAR dataset can then be downloaded via the command line using Terminal.

For the first use, it is highly likely that the `durlar_download` file will need to be made executable:

```
chmod +x durlar_download
```

By default, this script downloads the exemplar dataset (600 frames, direct link) for unit testing:

```
./durlar_download
```

It is also possible to select and download various test drives:

```
usage: ./durlar_download [dataset_sample_size] [drive]
dataset_sample_size = [ small | medium | full ]
drive = 1 ... 5
```

Given the substantial size of the DurLAR dataset, please download the complete dataset only when necessary:

```
./durlar_download full 5
```

Throughout the entire download process, it is important that your network remains stable and free from any interruptions. In the event of network issues, please delete all DurLAR dataset folders and rerun the download script. Currently, our script supports only Ubuntu (tested on Ubuntu 18.04 and Ubuntu 20.04, amd64). For downloading the DurLAR dataset on other operating systems, please refer to Durham Collections for instructions.

## B.3   Integrity Verification

For easy verification of folder data and integrity, we provide the number of frames in each drive folder in Table B.1, as well as the MD5 checksums of the zip files.

**Table B.1: The number of frames** in each drive folder.

| Drive ID | 20210716 | 20210901 | 20211012 | 20211208 | 20211209 | **Total** |
|---|---|---|---|---|---|---|
| **# of Frames** | 41993 | 23347 | 28642 | 26850 | 25079 | **145911** |

DurLAR LiDAR–Camera Calibration Details

Following the publication of the proposed DurLAR dataset and the corresponding paper [17], we identify a more advanced targetless calibration method [214] that surpasses the LiDAR-camera calibration technique previously employed in Section 3.5. As shown in Figure C.1, by overlaying the LiDAR intensity features and the camera gray-scale features with a certain level of transparency, we can see that our updated calibration results are ideal and accurate.

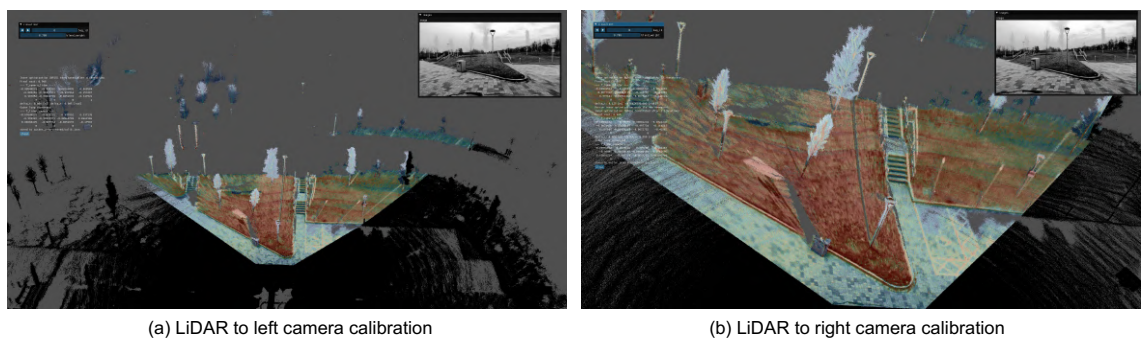

(a) LiDAR to left camera calibration      (b) LiDAR to right camera calibration

**Figure C.1:** LiDAR to stereo camera calibration and visualization.

Given that our Ouster OS1-128 operates as a spinning LiDAR, it faces challenges associated with its sparse and repetitive scan patterns [214], rendering the extraction of meaningful geometrical and texture information from a single scan particularly difficult. To address this, as shown in Figure C.2, we pre-process a continuous series of sparse point cloud frames by accumulating points while compensating for viewpoint changes and distortion [214].

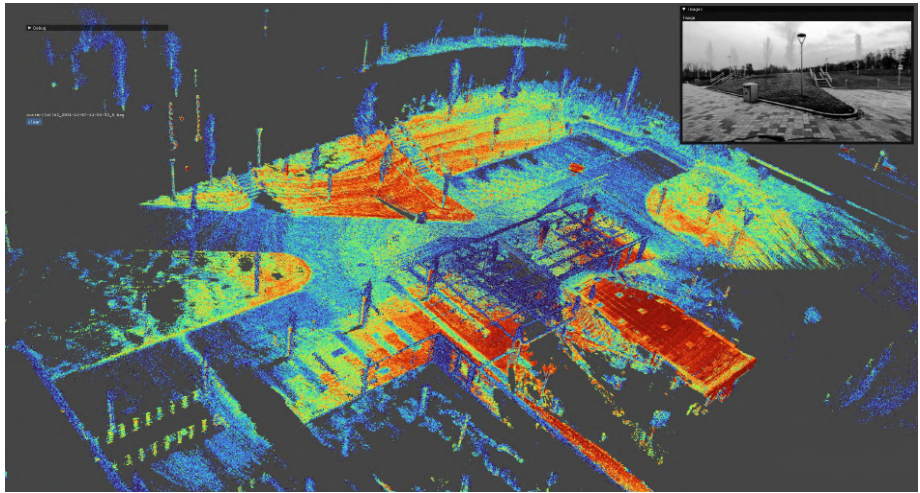

**Figure C.2:** LiDAR frame-wise aggregation allows for the generation of a denser point cloud from continuous dynamic LiDAR frames, resulting in detailed geometrical and surface texture information.

Given the densified point cloud and camera image, we find 2D-3D correspondences using SuperGlue [235]. As shown in Figure C.3, SuperGlue identifies correspondences between LiDAR points and camera images across different modalities, even with a relative low matching threshold. The results include numerous false correspondences that must be filtered out before pose estimation (green: inliers → red: outliers).

Based on the 2D-3D correspondences, an initial estimate of the LiDAR-camera transformation is derived using RANSAC and reprojection error minimization. Finally, the precise LiDAR-camera registration is achieved through NID [236] minimization.

We officially provide both the new and old versions of the calibration results and the original bag files for calibration, allowing users to utilize them as per their requirements.

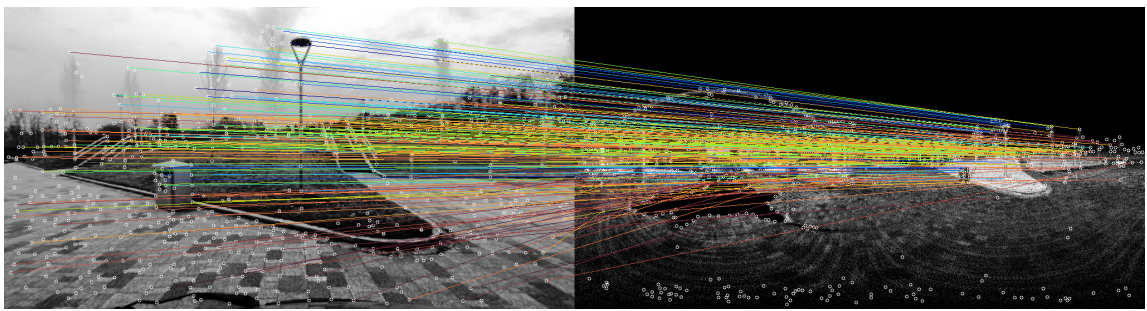

**Figure C.3:** SuperGlue identifies correspondences between LiDAR points and pixels.

---

# The Description on Semantic Classes

---

We supplement the semantic categories in two datasets used in Chapters 4 and 5 with
⬤ visualization colors, **full names** (`abbreviation names`), as well as detailed descriptions [1].

## D.1   SemanticKITTI Dataset

There are a total of 19 classes chosen for training and evaluation by merging classes with
similar motion statuses and discarding sparsely represented ones.

1. ⬤ **car**: This includes cars, jeeps, SUVs, and vans with a continuous body shape (*i.e.*,
   the driver cabin and cargo compartment are one).

2. ⬤ **bicycle** (`bicy`): Includes bicycles without the cyclist or possibly other passengers.
   The cyclist and passengers receive the label cyclist.

3. ⬤ **motorcycle** (`moto`): This includes motorcycles and mopeds without the driver
   or other passengers. Both driver and passengers receive the label motorcyclist.

---

[1]All descriptions are referenced from the official documentations in SemanticKITTI point labeler and
nuScenes devkit.

4. ● **truck** (`truc`): This includes trucks, vans with a body that is separate from the driver cabin, pickup trucks, as well as their attached trailers.

5. ● **other vehicle** (`o.veh`): Caravans, Trailers, and fallback category for vehicles not explicitly defined otherwise in meta category level vehicle.

6. ● **person** (`ped`): Persons moving by their own legs, sitting, or any unusual pose, but not meant to drive a vehicle.

7. ● **bicyclist** (`b.list`): Humans driving a bicycle.

8. ● **motorcyclist** (`m.list`): Persons riding a motorcycle.

9. ● **road**: Paved pathways primarily designed for the movement of vehicles, particularly automobiles, trucks, buses, and motorcycles.

10. ● **parking** (`park`): Areas where vehicles can be parked and left.

11. ● **sidewalk** (`walk`): Areas used mainly by pedestrians, and bicycles, but not meant for driving.

12. ● **other ground** (`o.gro`): Other areas that are not used by pedestrians or meant for driving.

13. ● **building** (`build`): Building walls, doors, *etc.*

14. ● **fence** (`fenc`): fences, small walls, crash barriers, *etc.*

15. ● **vegetation** (`veg`): Trees, and other forms of vertical growing vegetation.

16. ● **trunk** (`trun`): Tree trunks.

17. ● **terrain** (`terr`): Grass and all other types of horizontal spreading vegetation, including soil.

18. ● **pole**: Thin and elongated, typically vertically oriented poles, *e.g.*, sing or traffic signs.

19. ● **traffic sign** (`sign`): Traffic signs without pole.

## D.2 nuScenes Datatset

There are a total of 16 classes for LiDAR semantic segmentation, following the amalgamation of akin classes and the removal of rare ones.

1. ● **barrier** (`barr`): Any metal, concrete, or water barrier temporarily placed in the scene in order to re-direct vehicle or pedestrian traffic. In particular, includes barriers used in construction zones. If there are multiple barriers either connected or just placed next to each other, they should be annotated separately.

2. ● **bicycle** (`bicy`): Human or electric powered 2-wheeled vehicle designed to travel at lower speeds either on road surface, sidewalks, or bicycle paths.

3. ● **bus**: Buses designed to carry more than 10 people.

4. ● **car**: Vehicle designed primarily for personal use, *e.g.* sedans, hatch-backs, wagons, vans, mini-vans, SUVs, and jeeps.

5. ● **construction vehicle** (`const`): Vehicles primarily designed for construction. Typically very slow-moving or stationary. Cranes and extremities of construction vehicles are only included in annotations if they interfere with traffic. Trucks used for hauling rocks or building materials are considered trucks rather than construction vehicles.

6. ● **motorcycle** (`motor`): Gasoline or electric powered 2-wheeled vehicle designed to move rapidly (at the speed of standard cars) on the road surface. This category includes all motorcycles, vespas, and scooters. It also includes light 3-wheel vehicles, often with a light plastic roof and open on the sides, that tend to be common in Asia.

7. ● **pedestrian** (`ped`): All types of pedestrians moving around the cityscape.

8. ● **traffic cone** (`cone`): All types of traffic cones.

9. ● **trailer** (`trail`): Any vehicle trailer, both for trucks, cars, and motorcycles (regardless of whether currently being towed or not). Trailers hauled after a semi-tractor should be labeled as `trail`.

10. ● **truck**: Vehicles primarily designed to haul cargo including pick-ups, lorries, trucks, and semi-tractors.

11. ● **driveable surface** (`driv`): All paved or unpaved surfaces that a car can drive on with no concern of traffic rules.

12. ● **other flat** (`other`): All other forms of horizontal ground-level structures that do not belong to any of driveable surface, curb, sidewalk, and terrain. Includes elevated parts of traffic islands, delimiters, rail tracks, stairs with at most 3 steps, and larger bodies of water (lakes, rivers).

13. ● **sidewalk** (`walk`): Sidewalk, pedestrian walkways, bike paths, *etc*. Part of the ground designated for pedestrians or cyclists. Sidewalks do not have to be next to a road.

14. ● **terrain** (`terr`): Natural horizontal surfaces such as ground level horizontal vegetation ($\leq$ 20 cm tall), grass, rolling hills, soil, sand and gravel.

15. ● **manmade** (`made`): Includes man-made structures but not limited to: buildings, walls, guard rails, fences, poles, drainages, hydrants, flags, banners, street signs, electric circuit boxes, traffic lights, parking meters and stairs with more than 3 steps.

16. ● **vegetation** (`veg`): Any vegetation in the frame that is higher than the ground, including bushes, plants, potted plants, trees, *etc*. Only tall grass ($\geq$ 20 cm) is part of this.

## D.3    Excluded Semantic Classes

Certain categories (● unlabeled, ● outlier, ● other structure, ● other object, *etc*.), despite being annotated in the dataset ground truth, are excluded from the evaluations and tests. This exclusion is attributed to the inherent noise associated with LiDAR data or other related factors.

Related Resources of Publications

We supplement the related resources (*e.g.*, dataset, code, posters, videos, visualizations, homepages, supplementary materials, *etc.*) for all the publications by the author included in this thesis.

- Li Li, Khalid N. Ismail, Hubert P.H. Shum, and Toby P. Breckon. "DurLAR: A High-Fidelity 128-Channel LiDAR Dataset with Panoramic Ambient and Reflectivity Imagery for Multi-Modal Autonomous Driving Applications." In *International Conference on 3D Vision* (3DV). IEEE, 2021.

  **Related links:**  Paper   Dataset   GitHub   Demo Video   Poster

- Li Li, Hubert P.H. Shum, and Toby P. Breckon. "Less is more: Reducing task and model complexity for 3d point cloud semantic segmentation." In *Proceedings of the IEEE/CVF Conference on Computer Vision and Pattern Recognition* (CVPR). 2023.

  **Related links:**  Paper   Code   Demo Video   Poster   Homepage

- Li Li, Hubert P.H. Shum, and Toby P. Breckon. "RAPiD-Seg: Range-Aware Pointwise Distance Distribution Networks for 3D LiDAR Semantic Segmentation." In *European Conference on Computer Vision* (ECCV). Springer, 2024.

  **Related links:**  Paper   Code   Video   Oral   Poster